\documentclass[3p,authoryear,review]{elsarticle}

\let\paragraph\subsubsection

\usepackage[table]{xcolor}
\usepackage{amssymb}
\usepackage{lipsum}
\usepackage{lscape}
\usepackage{graphicx}%
\usepackage{multirow}%
\usepackage{amsmath,amssymb,amsfonts}%

\usepackage{amsfonts}
\usepackage{mathrsfs}%
\usepackage[title]{appendix}%
\usepackage{textcomp}%
\usepackage{manyfoot}%
\usepackage{booktabs}%
\usepackage{algorithm}%
\usepackage{algorithmicx}%
\usepackage{algpseudocode}%
\usepackage{listings}%
\usepackage{url}
\usepackage{physics}
\usepackage{amsmath}
\usepackage{tikz}
\usepackage{mathdots}
\usepackage{yhmath}
\usepackage{cancel}
\usepackage{color}
\usepackage{siunitx}
\usepackage{array}
\usepackage{multirow}
\usepackage{amssymb}
\usepackage{gensymb}
\usepackage{tabularx}
\usepackage{extarrows}
\usepackage{booktabs}
\usetikzlibrary{fadings}
\usetikzlibrary{patterns}
\usetikzlibrary{shadows.blur}
\usetikzlibrary{shapes}
\usepackage{listings}
\usepackage{graphicx} 
\usepackage{fancybox}
\usepackage{adjustbox}
\usepackage{twoopt} 
\usepackage{subcaption}
\usepackage{amsfonts} 
\usepackage{amssymb}
\usepackage{booktabs, multirow} 
\usepackage{soul}
\usepackage{changepage,threeparttable} 
\usepackage{forest}
\usepackage{pdflscape}
\usepackage{adjustbox}
\usetikzlibrary{shadows, trees, positioning}
\usepackage{booktabs}
\usepackage{xltabular}
\usepackage{tabularray}
\usepackage{enumerate}
\usetikzlibrary{shapes,arrows,positioning}
\usepackage[many]{tcolorbox}
\usetikzlibrary{calc}
\tcbuselibrary{skins}
\usepackage{dashrule}
\usetikzlibrary{decorations.pathreplacing}
\usepackage{titlesec}
\usepackage{float}
\usepackage{mathalpha}
\usepackage{dutchcal}
\usepackage{booktabs,multirow}
\usepackage{dblfloatfix}
\usepackage{placeins}

\usepackage{pgfplots}
\pgfplotsset{compat=newest}

\newcommandtwoopt{\insertboxtwo}[3][1.0][4.0cm]{\noindent\fbox{\begin{minipage}[c]{#1\textwidth}\parbox[c][#2]{#1\textwidth}{#3}\hfill\end{minipage}}}

\newenvironment{insertedtext}%
  {}%
  {}

\newcommand{\figcellxai}[1]{%
    \includegraphics[width=0.5\textwidth]{__FIG/XAI/#1}%
}

\renewcommand{\cite}{\citep}
\journal{Neurocomputing}

\begin{document}

\begin{frontmatter}

\title{The Impact of Stochasticity on the Rashomon Effect in Machine Learning}



\author[unisa]{Andrea Apicella}
\affiliation[unisa]{organization={Department of Information Engineering, Electrical Engineering, and Applied Mathematics (DIEM), University of Salerno},
            city={Fisciano (SA)},
            country={Italy}}
\author[unina]{Francesco Isgrò}
\affiliation[unina]{organization={Department of Electrical Engineering and Information Technology (DIETI), University of Naples Federico II},
            city={Naples},
            country={Italy}}

\author[unina]{Andrea Pollastro}

\author[unina]{Roberto Prevete}

\begin{abstract}
{\def\thefootnote{}\footnotetext{This work has been accepted to be published on Neurocomputing. The paper is available in open access at \url{https://www.sciencedirect.com/science/article/pii/S0925231226032200} (DOI: 10.1016/j.neucom.2026.135822). Please refer to the published version of the paper.}}Neural network training is inherently stochastic, with factors such as weight initialization leading to distinct models despite comparable predictive performance. This phenomenon is commonly associated with the \textit{Rashomon} effect, which describes the existence of multiple near-optimal models for the same task. Although the Rashomon effect has received increasing attention, it remains unclear whether different sources of training stochasticity contribute similarly or differently to its manifestations. In this work, we present an empirical study of the Rashomon phenomenon along three complementary dimensions: \textit{solution-space multiplicity}, \textit{predictive multiplicity}, and \textit{decision-basis multiplicity}. These dimensions are quantified through the size of the empirical Rashomon set, predictive ambiguity, and agreement between XAI attribution maps, respectively. By independently controlling three standard sources of stochasticity, namely weight initialization, mini-batch data ordering, and dropout, we isolate their respective contributions to each dimension of the Rashomon phenomenon. Experiments on tabular and image classification benchmarks reveal that these sources affect the three dimensions in different ways. In particular, larger empirical Rashomon sets do not necessarily correspond to greater predictive disagreement or lower explanation agreement, indicating that solution-space, predictive, and decision-basis multiplicity capture complementary rather than interchangeable aspects of the Rashomon effect. Overall, our results show that training stochasticity influences not only predictive performance but also the stability of predictions and explanations, highlighting the importance of identifying the specific sources of stochasticity responsible for different manifestations of the Rashomon phenomenon when assessing the reliability, reproducibility, and interpretability of neural network models.
\end{abstract}


\end{frontmatter}



\section{Introduction}
\label{sec:introduction}
Neural networks have achieved strong empirical performance across a wide range of supervised learning tasks~\cite{bishop2023deep}. 
\begin{insertedtext}
However, the reliability of neural network systems can be affected by uncertainty arising at different levels. For example, in dynamical and control settings, previous studies have addressed uncertainties associated with unknown system dynamics, external disturbances, delayed information, and state estimation~\cite{fan2026adaptive,tan2026adaptive}. By contrast, in supervised learning, a distinct form of uncertainty arises from the learning procedure itself. For instance, 
\end{insertedtext}
common training procedures, such as gradient descent, may not yield a unique solution. Different random factors, including parameter initialization and data ordering, typically induce distinct final neural network models while resulting in nearly identical generalization performance.
This phenomenon is captured by the notion of the \emph{Rashomon effect},  originally introduced in ~\cite{breiman2001statistical} to describe 
situations in which several competing models provide equally accurate descriptions of the same data-generating process. 

The existence of many near-optimal solutions can be formalized through the notion of the \textit{Rashomon set} as introduced in ~\cite{fisher2019all}: the set of models whose empirical loss lies within a tolerance $\epsilon$ of the best-performing  model in a given hypothesis class. 
The Rashomon effect is important not only because it reveals the existence of multiple near-optimal solutions, but also because such solutions may differ in ways that are not captured by aggregate performance metrics. For example, when a model is selected for deployment, one typically expects predictions and inferred input-output relationships to remain stable across equally accurate models. However, two models with equivalent task performance may nonetheless disagree on specific instances~\cite{marx2020predictive}, or assign conflicting importance to input features despite producing identical predictions~\cite{watson2022agree}. 
Such inconsistencies are invisible under aggregate metrics and can undermine reliability and reproducibility in practical applications.

Although these manifestations are often discussed under the common label of the Rashomon effect, they capture different aspects of model multiplicity.

In particular, equally accurate models may differ in at least three complementary ways. First, the learning problem may admit many near-optimal solutions. Second, these solutions may produce different predictions on individual instances. Third, they may rely on different evidence even when their predictions coincide. Distinguishing these aspects is important because they have different implications for reliability, reproducibility, and interpretability, and may also be influenced by different sources of training stochasticity. In this work, we explicitly distinguish these three aspects and analyse how different sources of training stochasticity affect each of them. 

Specifically, we characterise the Rashomon phenomenon along three complementary dimensions:

i) 
\emph{solution-space multiplicity}, describing how many near-optimal models exist for the task;  empirically, this dimension can be quantified through the size of the Empirical Rashomon set ~\cite{semenova2022existence}, i.e. how many models within a given function class achieve near-optimal predictive performance~\cite{semenova2022existence}. 

ii) 
\emph{predictive multiplicity}~\cite{marx2020predictive}, describing the extent to which models in the Rashomon set disagree on individual inputs. Even when aggregate performance is comparable, models may correctly classify disjoint subsets of data, or assign substantially different posterior probabilities to each class while agreeing on the predicted label. When a deployed model is later replaced or retrained, such disagreements may lead to inconsistent decisions for the same individual across time. 
This dimension can be quantified through a proper \emph{ambiguity measure} ~\cite{marx2020predictive} (see Section \ref{sec:setup}), defined as the fraction of samples for which at least two models in the empirical Rashomon set produce different predictions.

iii) 
\begin{insertedtext}
    \emph{decision-basis multiplicity}, describing the extent to which models with comparable predictive performance assign the same predicted class to a given input while associating that prediction with different input features or feature-level evidence.  Notice that this notion overlaps with related concepts in the literature, such as \emph{procedural multiplicity} \cite{black2022model}, \emph{solution diversity} \cite{muller2023empirical}. We adopt a different terminology to emphasize the specific case of identical predictions for the same input arising from different underlying bases. Post-hoc eXplainable Artificial Intelligence (XAI) attribution methods provide a practical means of investigating this dimension by assigning feature-relevance scores to individual predictions~\cite{montavon2019layer,ribeiro2016should,apicella2021general}. In our analysis, we compare attribution profiles across pairs of models that predict the same class for the same input. Higher attribution agreement indicates that the models associate their common prediction with similar feature-level evidence, whereas lower agreement provides stronger attribution-based evidence of decision-basis multiplicity. This interpretation concerns the feature-level evidence associated with model predictions and does not entail a causal claim about the models' internal computational mechanisms~\cite{jacovi2020towards,jacovi2021aligning} .
\end{insertedtext}

\begin{figure}
    \centering
    \scalebox{0.9}{
    \includegraphics[width=1\linewidth]{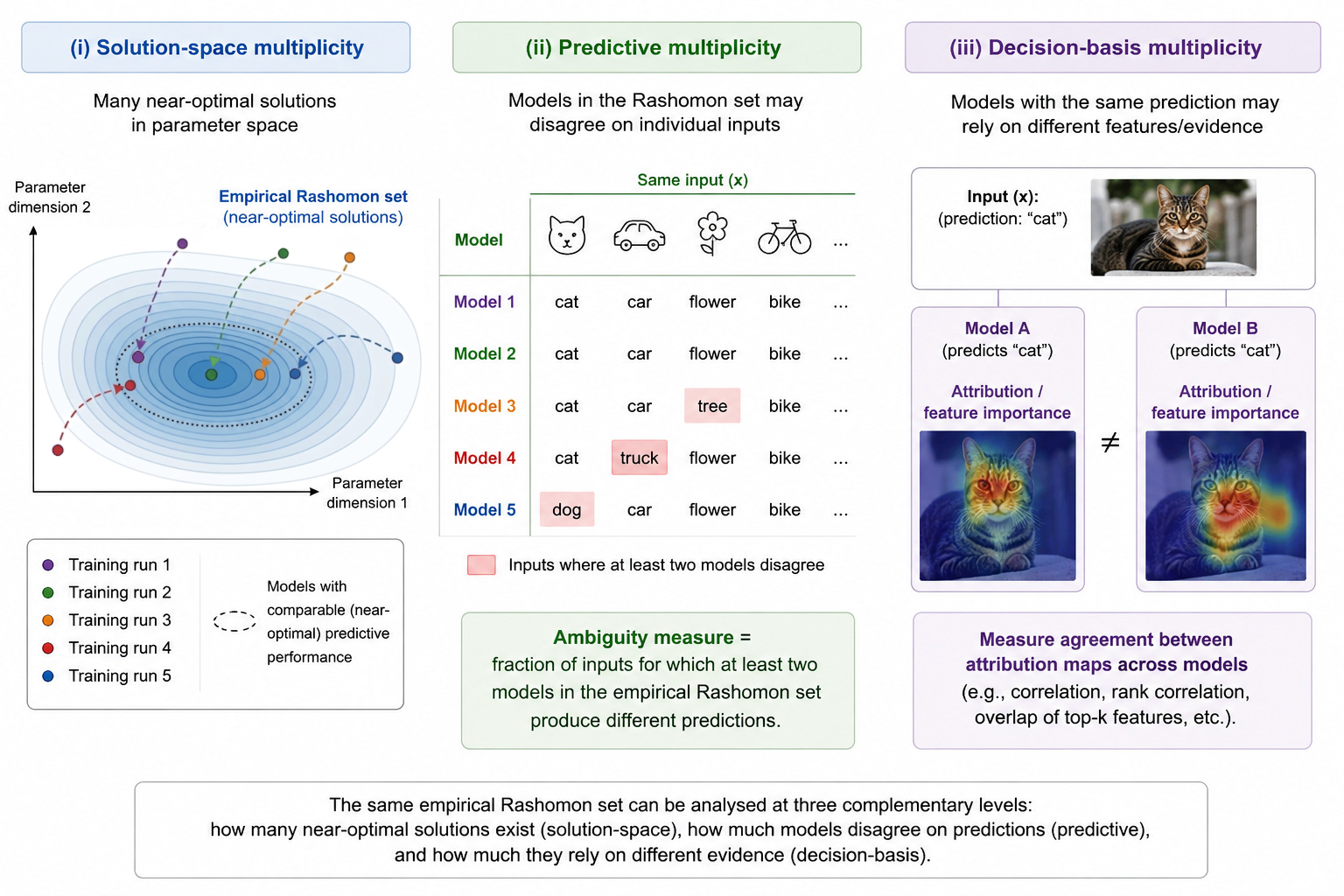}
    }
    \caption{A recap of the investigated Rashomon effect dimensions. See text for further details.}
    \label{fig:main}
\end{figure}

These three dimensions are summarized in Fig. \ref{fig:main}.

\begin{insertedtext}
Previous studies established both the existence of multiple near-optimal models and the possibility that such models disagree on individual predictions \cite{breiman2001statistical,marx2020predictive}. Differently, this work does not propose a new theoretical relationship between Rashomon-set size and predictive disagreement. Instead, it complements these studies through a controlled empirical decomposition of training stochasticity, examining whether parameter initialisation, mini-batch ordering, and dropout affect the size, predictions, and feature-attribution profiles of empirical Rashomon sets in different ways.
\end{insertedtext}

In the current literature, it remains unclear which neural network training factors involved in the Rashomon effect contribute most to each dimension, or in other words: ``which sources of stochasticity are primarily responsible for solution-space, predictive, and decision-basis multiplicity?'' and whether these contributions are consistent across architectures and datasets. 
Notice that understanding which sources of training stochasticity contribute to different manifestations of the Rashomon effect is important because different sources call for different mitigation strategies and have different implications for reliability, reproducibility, and interpretability.
In this work, we provide a systematic empirical analysis of the Rashomon phenomenon along all three dimensions considering three widely used stochastic factors: \textit{weight initialisation}, \textit{data ordering}, or \textit{dropout}. 
This perspective is particularly relevant because the stochastic factors considered in this work are not incidental sources of noise, but standard components of neural network training that are typically introduced with a positive role \cite{bishop2023deep}. However, the same sources of randomness that improve optimisation and generalisation may also induce variability across equally well-performing models. In this work, we therefore study their complementary role as potential drivers of solution-space, predictive, and decision-basis multiplicity.
We conduct experiments on tabular datasets (ISOLET, Optical Recognition of Handwritten Digits, Waveform) using Multilayer Perceptrons (MLPs), and on image classification benchmarks (Fashion-MNIST, CIFAR-10) using a ResNet-20 architecture. 
For each configuration, we train pools of $100$ models by independently varying one source of stochasticity among weight initialisation, mini-batch ordering, and dropout mask sequences, while holding the remaining sources fixed, as well as a baseline in which all sources of randomness vary simultaneously, reproducing standard training conditions.

Our main findings are:
\begin{itemize}
\item The solution-space multiplicity depends strongly on both the source of stochasticity and the dataset. Weight initialisation often produces wider accuracy distributions and consequently smaller Rashomon sets, whereas data shuffling and dropout generally yield larger sets by generating more concentrated distributions of near-optimal models.

\item Predictive multiplicity,  measured through ambiguity~\cite{black2022model}, is not directly related to Rashomon set size. In particular, weight initialisation often produces higher predictive disagreement despite yielding smaller Rashomon sets, while data shuffling and dropout tend to generate larger sets whose members remain comparatively consistent in their predictions.

\item Decision-basis multiplicity, empirically estimated through attribution-map similarity using four different post-hoc XAI methods, that are Vanilla Gradients, Integrated Gradients, Occlusion, and LRP, reveals that models with similar predictive performance may nevertheless rely on different features. On tabular datasets, weight initialisation generally produces the lowest top-$k$ values, whereas the differences among stochasticity sources are less pronounced on image datasets.

\item The relationship between solution-space multiplicity, predictive multiplicity, and decision-basis multiplicity is non-trivial. \begin{insertedtext}
Consistent with \cite{marx2020predictive,breiman2001statistical}, our experimental assessment shows that
\end{insertedtext}
larger Rashomon sets do not necessarily correspond to higher predictive disagreement or higherdecision-basis multiplicity indicating that these dimensions capture complementary aspects of the Rashomon phenomenon and should be analysed jointly.

\end{itemize}

These results indicate that models in the Rashomon set are not interchangeable, even when their aggregate test performance is indistinguishable. 
Controlling the stochastic factors rather than treating them as an irrelevant implementation detail, may therefore be necessary to ensure reproducibility of both predictions and explanations across independent training runs.

The remainder of the paper is organised as follows. Section \ref{sec:background} reviews the background and related work on the Rashomon effect, predictive multiplicity, and explanation stability. Section \ref{sec:setup} describes the experimental methodology, including the formal definitions of the three investigated dimensions, the datasets, models, and training protocol. Section \ref{sec:results} presents the experimental results. Section \ref{sec:discussion} discusses the implications of the findings and their limitations. Section \ref{sec:guidelines} provides practical guidelines for mitigating stochastic multiplicity. Finally, Section \ref{sec:conclusion} concludes the paper and outlines directions for future work.
\section{Background and Related Work}
\label{sec:background}

\begin{insertedtext}
The reliability of machine learning systems can be affected by different forms of uncertainty and variability. One line of research investigates changes in the data encountered at deployment time. In this context, open-set classification methods aim to recognise inputs belonging to classes that were not observed during training \cite{sun2025open}, or domain generalisation studies investigate how models can maintain their performance under changes in operating conditions and data distributions~\cite{sun2025multi,apicella2025domain}. These approaches primarily address uncertainty related to data and deployment by considering changes in the label space or input distribution.

A different source of variability arises even when the learning task, label space, and data-generating setting remain fixed. Because a learning procedure may return multiple models having comparable predictive performance, uncertainty can also originate from the particular model selected or produced during training. This second form of variability is captured by the Rashomon effect, which constitutes the focus of the present work.
\end{insertedtext}

The notion of the Rashomon effect \cite{breiman2001statistical} has become increasingly influential in machine learning (ML) for describing the coexistence of multiple models that achieve similar empirical performance while differing substantially in structure or behavior. 
The term was introduced into the ML literature by Breiman~\cite{breiman2001statistical}, who used it to illustrate how complex learning systems may admit several high-performing models providing distinct explanations or decision rules for the same data-generating process. In Breiman's formulation, the Rashomon effect can be summarized as follows: if several competing models fit the data equally well but imply different explanations of the underlying phenomenon, it becomes unclear which explanation should be preferred. 
\begin{insertedtext}

Recent work tried to systematize the rapidly expanding literature on model multiplicity. \cite{ganesh2025systemizing} propose a general framework in which multiplicity encompasses any relevant behavioural difference among models indistinguishable according to the criteria defining a Rashomon set. Their taxonomy distinguishes predictive and explanation multiplicity from other forms involving fairness, representations, datasets, model complexity, and feature interactions. It also identifies random seeds and other development choices as sources of arbitrariness in model selection. Our work complements this systematisation through a controlled empirical study of how three specific sources of training stochasticity—weight initialization, data shuffling, and dropout—affect Rashomon-set size, predictive multiplicity, and decision-basis multiplicity in deep learning models.
\end{insertedtext}
Since Breiman's original formulation, several works have sought to characterize the Rashomon effect more precisely by formalizing its different manifestations.
In \cite{black2022model}, the authors distinguish between \emph{predictive multiplicity} and \emph{procedural multiplicity}. Predictive multiplicity occurs when models with comparable performance produce different predictions for some inputs \cite{marx2020predictive}, whereas procedural multiplicity refers to models that achieve the same performance while relying on different internal mechanisms. 
Predictive multiplicity can be viewed as a particular case of procedural multiplicity, since different predictions necessarily imply different decision boundaries and  therefore different internal representations.  
In this context, \cite{marx2020predictive} introduced two formal measures to quantify predictive multiplicity, namely \emph{ambiguity}, representing the number of predictions that may change across competing models, and \emph{discrepancy}, representing the maximum number of predictions that may change across competing models, together with methods to compute them in linear classification problems.
\begin{insertedtext}
Recent research has also addressed the computational construction of Rashomon sets for specific model classes.  \cite{arslan2026sorted} introduced a framework  that enumerates sparse decision trees with binary features in increasing order of their objective value within a Rashomon set.
\end{insertedtext}
~\cite{semenova2022existence} investigated when a large Rashomon set is likely to contain simpler models. They showed that if many near-optimal solutions exist, then some of them can be chosen to be structurally simple, while still achieving competitive performance. 
The same authors in~\cite{semenova2023path} showed that increasing noise in a dataset tends to enlarge the Rashomon set, as standard training practices cause many distinct models to achieve similar performance under noisy conditions. 
\cite{coupkova2024rashomon} studied theoretical properties of the Rashomon phenomenon by analyzing the proportion of near-optimal models within a hypothesis class, referred to as the Rashomon ratio, focusing on theoretical guarantees for model selection.
\begin{insertedtext}
Similarly, ~\cite{cavus2026quantifying} considered the Rashomon ratio, defined as the proportion of candidate models satisfying the near-optimality criterion, as a coarse indicator of the potential variability of model explanations.
\end{insertedtext}

The relationship between Rashomon sets and model explanations has also received growing attention. Leventi et al. in \cite{leventi2022rashomon} discuss the phenomenon of \emph{explanative multiplicity}, highlighting how models belonging to the same Rashomon set may rely on different explanatory mechanisms. 
\begin{insertedtext}
\citet{cavus2026quantifying} introduce Rashomon Partial Dependence Profiles, which aggregate feature-effect profiles across near-optimal models generated by an AutoML system.
\end{insertedtext}
In parallel, a large body of work has investigated the stability of explanations produced by explainable artificial intelligence (XAI) methods. 
For instance, Ghorbani et al. in \cite{ghorbani2019interpretation} define an interpretation as \emph{fragile} if it is possible to construct a perceptually indistinguishable input that yields the same prediction as the original one but produces a substantially different explanation.
Further evidence of explanation instability has been provided by \cite{watson2022agree}, who showed that explanations generated by methods such as SHAP and Integrated Gradients can vary significantly due to changes in training hyperparameters, including weight initialization and the order of training samples. 
To quantify this variability, the authors introduced the notion of \emph{explanation separability} between two models. 
Their results suggest that deep learning models can learn different feature representations depending on training conditions, leading to different explanations. 
To disentangle model variability from explanation artifacts, the authors also considered several explanation quality metrics, including \emph{infidelity}, \emph{sensitivity}, and \emph{explanation accuracy}. 
However, these analyses focus primarily on the behavior of a single explanation method rather than comparing explanations generated by different XAI techniques.
\begin{insertedtext}
\cite{parashar2026quantifying} study the disagreement problem between LIME and SHAP explanations generated for the same predictive model and quantify agreement through the overlap of highly attributed features.
\end{insertedtext}

Other studies such as \cite{mehrer2020individual} and \cite{bansal2020sam} study how internal representations of deep neural networks and the explanations generated by XAI methods vary with different hyperparameters. 
Explanation agreement is typically measured using metrics such as the overlap between the \textit{top-$k$} most important features or standard distance measures such as Euclidean distance. 
More generally, explanation disagreement has been widely documented in the XAI literature \cite{krishna2022disagreement}. Different explanation techniques applied to the same model and input often produce substantially different results \cite{neely2021order}. 

The impact of Rashomon sets on explainability has been investigated in specific contexts. 
\cite{poiret2023can} analyze how dataset size influences the consistency of explanations within a Rashomon set, showing that explanations tend to become more stable as the amount of training data increases. 
\cite{muller2023empirical} conducted a comprehensive empirical evaluation of attribution-based XAI methods from several complementary perspectives. Their study examined the effect of model initialization, the hyperparameters of the explanation methods, changes in the underlying predictive model, and the disagreement across different XAI techniques.
Similarly, \cite{bogaert2024explanation} showed that large language models with similar predictive performance but different hyperparameter configurations may provide substantially different explanations for the same inputs.

Despite these efforts, the interaction between different stochastic factors in model training, the structure of Rashomon sets, and the agreement of explanations remains insufficiently understood. 
In particular, it remains unclear how different sources of training stochasticity contribute to the various manifestations of the Rashomon effect. In particular, no previous study has systematically disentangled the impact of individual stochastic training factors on solution-space, predictive, and decision-basis multiplicity within a unified experimental framework. To address this gap, we present a systematic empirical analysis of the effect of different sources of training stochasticity on solution-space, predictive, and decision-basis multiplicity across multiple datasets and neural network architectures.

\section{Experimental Setup}
\label{sec:setup}
This section describes the experimental framework adopted to analyse the impact of different sources of stochasticity on the Rashomon phenomenon.
In particular, we investigate three complementary dimensions, each capturing a different aspect of the phenomenon:
(i) \emph{solution-space multiplicity}, describing the number of near-optimal models;
(ii) \emph{predictive multiplicity}, describing the extent to which near-optimal models produce different predictions; and
(iii) \emph{decision-basis multiplicity}, describing the extent to which near-optimal models rely on different input features when making their predictions.
Each dimension is approximated through a specific experimental procedure, described in the following subsections. The following subsections present the notation adopted throughout the paper, followed by the experimental setup, including the datasets, models, and training configurations considered in the experiments.

\subsection{Adopted notation and definitions}
Let ${D} = \{(\mathbf{x}^{(i)}, y^{(i)})\}_{i=1}^{N}$ denote a labelled  dataset, where $\mathbf{x}^{(i)} \in {X}$ is an input sample and  $y^{(i)} \in {Y}$ its corresponding label.
In general, we consider a parametric model family $\mathcal{F}=\{f_\theta | \theta \in \Theta\}$ where $f_\theta : {X} \rightarrow {Y}$ is a ML model with parameters $\theta \in \Theta$.
Let $L :\mathcal{F} \rightarrow \mathbb{R}_{\geq 0}$ denote a loss function evaluated on a held-out set ${D}_{\mathrm{HO}}$, which is not used during training. Lower values of $L$ correspond to better predictive performance.
\paragraph{Rashomon set and Empirical Rashomon set}
The \emph{Rashomon set} at tolerance $\epsilon \geq 0$ is the subset of $\mathcal{F}$  consisting of all models whose empirical loss is within $\epsilon$ of the best-performing model:
\[ \mathcal{R}_\epsilon = \left\{ f_\theta \in \mathcal{F} \;\middle|\; L(f_\theta) \leq L_\mathcal{F}^{*} + \epsilon\right\},\]
where $L_{\mathcal{F}}^{*} = \min_{f_\theta \in \mathcal{F}} L(f_\theta)$. Note that $\mathcal{R}_0$ contains all models achieving the minimum empirical loss,  while $|\mathcal{R}_\epsilon|$ is a monotonically non-decreasing function of $\epsilon$.

However, computing the exact Rashomon set of a neural network over a continuous hypothesis  space is generally intractable~\cite{semenova2022existence}. 
We therefore approximate it empirically by training multiple instances of the same architecture under controlled stochastic configurations, as detailed in  Sec. ~\ref{sec:setup}. 
We refer to this as the \emph{empirical Rashomon set}, emphasising that it captures  only the subset of near-optimal models reachable through the specific training  procedures considered.
More formally, let \[ {F} = \{f_{\theta_1}\in \mathcal{F}, \dots, f_{\theta_K}\in \mathcal{F}\} \] denote the pool of models obtained by training the same architecture $K$ times under different stochastic configurations.
The \emph{empirical Rashomon set} at tolerance $\epsilon \geq 0$ is the subset of ${F}$  whose loss is within $\epsilon$ of the best-performing model in the pool:
\[ {R}_\epsilon = \left\{ f_\theta \in {F} \;\middle|\; L(f_\theta) \leq L_{F}^{*} + \epsilon\right\},\]
where $L_{F}^{*} = \min_{f_\theta \in {F}} L(f_\theta)$.

In our experiments, $L$ is defined as the classification error $L(f_\theta) = 1 - \mathrm{Acc}(f_\theta)$, where
\[
\mathrm{Acc}(f_\theta) = \frac{1}{|{D}_{\mathrm{HO}}|} 
\sum_{(\mathbf{x}, y) \in {D}_{\mathrm{HO}}} 
\mathbf{1}\!\left[f_\theta(\mathbf{x}) = y\right].
\]
Using this definition, the Empirical Rashomon set can be equivalently expressed as:
\[
{R}_\epsilon =
\left\{f_\theta \in {F}\;\middle|\;
\mathrm{Acc}(f_\theta) \geq \mathrm{Acc}_{F}^{*} - \epsilon
\right\},
\quad \mathrm{Acc}_{F}^* = \max_{f_\theta \in {F}} \mathrm{Acc}(f_\theta).
\]

\paragraph*{Solution-space multiplicity}
The existence of Rashomon sets, i.e., sets of models achieving nearly equivalent predictive performance on a given task, reflects a form of solution-space multiplicity. 
In this context, we consider the \emph{empirical Rashomon set size}, denoted as $|\mathcal{R}_\epsilon|$, as a practical proxy for its extent: larger sets indicate that many distinct parameter configurations achieve comparable performance, whereas smaller sets may suggest that the learning problem constrains the space of viable solutions more tightly.

\paragraph*{Predictive multiplicity}
The size of the Rashomon set alone does not fully characterize model multiplicity, as models within $\mathcal{R}_\epsilon$ may still differ in their predictions and input--output relationships.
With \textit{Predictive multiplicity} we refer to the phenomenon where models with comparable performance differ in their predictions. In other words, two models predict different classes for the same input ~\cite{marx2020predictive,black2022model}.
We quantify predictive multiplicity in an empirical Rashomon Set $R_{\epsilon}$ through the ambiguity of the empirical Rashomon set, defined as the fraction of held-out samples produce different predictions, that is:
\begin{equation}
        \mathrm{Amb}({R}_\epsilon) = 
    \frac{1}{|{D}_{HO}|}
    \sum_{(\mathbf{x}, y) \in {D}_{\mathrm{HO}}}
    \mathbf{1}\!\left[
        \exists\, f_{\theta_i}, f_{\theta_j} \in {R}_\epsilon :
        f_{\theta_i}(\mathbf{x}) \neq f_{\theta_j}(\mathbf{x})
    \right].
\end{equation}


\paragraph*{Decision-basis multiplicity}
\begin{insertedtext}
\emph{Decision-basis multiplicity} describes the possibility that models with comparable predictive performance produce the same decision for a given input while differing in the input features or feature-level evidence associated with that decision. In this work, we empirically investigate decision-basis multiplicity by comparing post-hoc feature-attribution profiles across pairs of models that predict the same class for the same input. However, although lower attribution agreement can be interpreted as stronger attribution-based evidence of decision-basis multiplicity, it should not be regarded as conclusive evidence that the models implement different causal mechanisms~\cite{jacovi2020towards,jacovi2021aligning}.
\end{insertedtext}

Formally, given a model $f_\theta \in \mathcal{R}_\epsilon$ and an input $\mathbf{x}$, an attribution method $E$ produces a relevance map $ R_{\theta}^{(E)}(\mathbf{x}) \in \mathbb{R}^{d}, $ where $d$ is the number of input features and each entry represents the relevance assigned to the corresponding feature for the prediction $f_\theta(\mathbf{x})$.
Given two models $f_{\theta_i}, f_{\theta_j} \in \mathcal{R}_\epsilon$, we compare their attribution maps through a similarity measure $ S\!\left(R_{\theta_i}^{(E)}(\mathbf{x}), R_{\theta_j}^{(E)}(\mathbf{x})\right).$
In this work, we adopt a feature-based similarity measure based on the overlap between the top-$k$ most relevant features. Let $ \mathcal{T}_k\!\left(R_{\theta}^{(E)}(\mathbf{x})\right) $ denote the set of indices corresponding to the $k$ highest relevance scores. The resulting Top-$k$ similarity is defined as 
\begin{equation}
S_k\!\left(R_{\theta_i}^{(E)}(\mathbf{x}), R_{\theta_j}^{(E)}(\mathbf{x})\right) = \frac{\left| \mathcal{T}_k\!\left(R_{\theta_i}^{(E)}(\mathbf{x})\right) \cap \mathcal{T}_k\!\left(R_{\theta_j}^{(E)}(\mathbf{x})\right) \right| }{k}.    
\end{equation}
 
This measure quantifies the fraction of the top-$k$ most relevant features shared by the two attribution maps.

\subsection{Datasets}
Experiments are conducted on five classification benchmarks spanning two data modalities. 
In particular, we use three tabular datasets from the UCI Machine Learning Repository~\cite{asuncion2007uci}:
\begin{itemize}
    \item \textbf{ISOLET}: 617 acoustic features extracted from 
    spoken letter names, 26 classes (letters A--Z), approximately  7797 samples.
    \item \textbf{Optical Recognition of Handwritten Digits}: 
    containing 5,620 instances represented by 64 features corresponding to an \(8\times8\) pixel grid.
    \item \textbf{Waveform Database Generator}: 21 continuous 
    features from a mixture of waveforms, 3 classes, 5000 samples.
\end{itemize}
For all adopted tabular datasets, samples are split into training ($80 \%$) and test ($20 \%$) sets using a fixed random seed, shared across all experimental configurations. The Rashomon sets were computed using as held-out set $D_{HO}$ the respective test set. A fixed $10 \%$ subset of the training set is used as a validation set.

Furthermore, we use two standard image classification benchmarks:
\begin{itemize}
    \item \textbf{Fashion-MNIST}~\cite{xiao2017fashion}: $28 \times 28$ grayscale  images of clothing items, 10 classes, 60000 training and 10000 test samples.
    \item \textbf{CIFAR-10}~\cite{krizhevsky09learning}: $32 \times 32$ colour images across 10 object categories, 50{,}000 training and 10{,}000 test samples.
\end{itemize}
For both image datasets, we adopt the standard train/test splits. Also in these cases, the Rashomon set for each datasets was computed using as held-out set $D_{HO}$ the respective test set. A fixed $10 \%$ subset of the training set is used as a validation set.

\subsection{Model Architectures and Training}
\begin{table}[t]
\centering
\begin{tabular}{lcc}
\toprule
\textbf{Hyperparameter} & \textbf{Tabular datasets (MLP)} & \textbf{Image datasets (ResNet-20)} \\
\midrule
Architecture & MLP (2 hidden layers) & ResNet-20 \\
Hidden units & 256, 256 & -- \\
Activation function & ReLU & ReLU \\
Optimizer & SGD & SGD \\
Momentum $\mu$ & 0.9 & 0.9 \\
Learning rate $\eta$ & 0.001 & 0.1 \\
Weight decay & $10^{-4}$ & $10^{-4}$ \\
Epochs & 100 & 100 \\
Batch sizes & $\{32,64,128,256\}$ & $\{32,64,128,256\}$ \\
Dropout rate & $0.5$ & $0.5$ \\
\bottomrule
\end{tabular}
\caption{Hyperparameter values for the adopted architectures.}
\label{tab:hyperparameters}
\end{table}
For the tabular and image datasets, we adopted a 2-layer MLP and \text{ResNet-20}~\cite{he2016deep}, respectively.


All the hyperparameters are summarised in Table~\ref{tab:hyperparameters}.
We evaluate four batch sizes ($32$, $64$, $128$, and $256$), covering a range from relatively small to relatively large mini-batches, in order to investigate whether the stochasticity induced by mini-batch sampling influences the size and characteristics of the empirical Rashomon set.

\begin{insertedtext}
All the analyses are conducted over ten values of the tollerance parameter $\epsilon$, ranging from $0.001$ to $0.03$. This interval was selected empirically after preliminary experiments and was found to be sufficient to capture the full evolution of the empirical Rashomon set across the considered experimental settings.  We examine how the investigated dimensions evolve across this range, basing our conclusions on trends observed across the complete tolerance range.
\end{insertedtext}

To isolate the contribution of each stochastic component of the training procedure, we consider four configurations. In all of them, dropout is present as a regulariser with $p = 0.5$; 
\begin{insertedtext}
Each configuration produces a collection (hereafter referred to as a \textit{pool}) of models that differ according to the sources of randomness varied across training runs. Each pool contained 100 models, indexed by $ i \in \{1,2,\dots,100\}$. Whenever weight initialization or mini-batch ordering or dropout masks had to remain fixed within a pool, the corresponding pseudorandom seeds were held constant across models with a value of $0$. For each source of stochasticity varied across the model pool, model $i$, with $i\in\{1,2,\ldots,100\}$, was assigned the corresponding seed $i$.
\end{insertedtext}
\begin{insertedtext}
\begin{table}[t]
\centering

\caption{Variance-stabilisation analysis for the empirical Rashomon-set fraction. Results aggregate 80 configurations across five datasets, four batch sizes, four stochasticity conditions, and ten performance tolerances. Each configuration was evaluated using 500 bootstrap resamples. SD denotes the bootstrap standard deviation; absolute deviation is computed relative to the estimate obtained from the complete observed pool.}
\label{tab:pool-size-stability}
\begin{tabular}{rccc}
\toprule
Number of models &
Median SD &
75th percentile SD &
Median absolute deviation \\
\midrule
10  & 0.097 & 0.170 & 0.143 \\
20  & 0.074 & 0.137 & 0.082 \\
30  & 0.062 & 0.118 & 0.054 \\
50  & 0.046 & 0.095 & 0.033 \\
75  & 0.038 & 0.079 & 0.019 \\
100 & 0.032 & 0.066 & 0.012 \\
\bottomrule
\end{tabular}
\label{tab:models}
\end{table}
The number of models per configuration was chosen as $100$ after assessing the stability of the empirical Rashomon-set fraction through bootstrap analysis (reported in Tab. \ref{tab:models}).
\end{insertedtext}

The investigated configurations are:
\begin{itemize}
    \item \textbf{Weight Initialisation} (\texttt{init}): each model is assigned  a different random initialisation seed for the weight initialization. The mini-batch ordering and dropout mask sequence are kept identical across all models in the pool.

    \item \textbf{Data Shuffling} (\texttt{shuffle}): each model sees a different random permutation of training mini-batches across epochs. Weight initialisation and dropout mask sequence are fixed.

    \item \textbf{Dropout} (\texttt{dropout}): each model is assigned a different random seed for the dropout mask sequence, producing a distinct sequence of masks across training steps. Weight initialisation and mini-batch ordering are fixed.

    \item \textbf{All Sources} (\texttt{all}): initialisation, mini-batch ordering,  and dropout mask sequence all vary independently across models, reproducing standard training conditions in which no source of randomness is controlled.
\end{itemize}

Results are compared across stochasticity configurations at fixed batch size and dataset.
This design allows a controlled comparison of the effect of each stochastic  component on all the investigated dimensions. In particular:
\begin{itemize}
    \item for solution-space multiplicity, we measure the \textit{Empirical Rashomon Set size} $|{R}_\epsilon|$ as a function of the tollerance value  $\epsilon$ (see eq.\ref{})  for each configuration, estimating how the number of near-optimal models grows as the tolerance $\epsilon$ increases.
    
    \item for predictive multiplicity, we analyse the output of the Ambiguity $Amb$ function as defined in eq. \ref{} jointly with the best test accuracy achieved within the  Rashomon set  $\mathrm{Acc}^*({R}_\epsilon)$. In other words, for each stochasticity configuration and value of $\epsilon$, we consider $\mathrm{Amb}({R}_\epsilon)$ against $\mathrm{Acc}^*({R}_\epsilon)$, for each value of $\epsilon$.

    \item for decision-basis multiplicity, we assess whether models in ${R}_\epsilon$ rely on similar input features to produce their predictions, using four attribution methods: Vanilla Gradients (VG), Integrated Gradients (IG)~\cite{sundararajan2017axiomatic}, 
    \begin{insertedtext}
        the occlusion-based method proposed in \cite{zeiler2014visualizing} (which we refer to as "Occlusion" throughout the remainder of this paper)
    \end{insertedtext}
    and Layer-wise Relevance Propagation (LRP)~\cite{montavon2019layer}. The similarity is computed using 
    \begin{insertedtext}
    Top-$k$     
    \end{insertedtext}
     feature overlap 
     \begin{insertedtext}
     with $k \in \{5\%,10\%,20\%,25\%\}$
     \end{insertedtext}
     as attribution similarity measure $S$.
\end{itemize}

\section{Results}
\label{sec:results}

This section presents the experimental results obtained using the methodology described in Section~\ref{sec:setup}. We begin by reporting the distribution of model accuracies, as it provides the reference for constructing the empirical Rashomon sets and allows us to verify that the subsequent analyses are performed on models with comparable predictive performance. We then analyze the three complementary dimensions of the Rashomon phenomenon introduced earlier: (i) \emph{solution-space multiplicity}, quantified through the empirical Rashomon set size; (ii) \emph{predictive multiplicity}, measured by Ambiguity; and (iii) \emph{decision-basis multiplicity}, assessed through the similarity of XAI attribution maps.
\subsection{Accuracy Distributions}

\begin{table*}[t]
\centering
\scriptsize
\setlength{\tabcolsep}{5pt}
\renewcommand{\arraystretch}{1.15}

\caption{Mean test accuracy (\%) $\pm$ standard deviation across datasets, batch sizes, and stochasticity sources.}
\label{tab:accuracy_summary}

\begin{tabular}{llccc|cc}
\toprule

&
&
\multicolumn{3}{c|}{\textbf{MLP}}
&
\multicolumn{2}{c}{\textbf{ResNet-20}}
\\

\cmidrule(lr){3-5}
\cmidrule(lr){6-7}

\textbf{Source}
&
\textbf{BS}
&
\textbf{ISOLET}
&
\textbf{Optical}
&
\textbf{Waveform}
&
\textbf{CIFAR-10}
&
\textbf{F-MNIST}
\\

\midrule

\multirow{4}{*}{All}
& 32  & $86.4\pm1.5$ & $95.4\pm0.6$ & $84.3\pm1.2$ & $86.7\pm0.4$ & $93.1\pm0.2$ \\
& 64  & $86.4\pm1.6$ & $95.6\pm0.5$ & $84.2\pm1.1$ & $86.4\pm0.3$ & $93.0\pm0.2$ \\
& 128 & $86.0\pm1.7$ & $95.6\pm0.5$ & $84.2\pm1.1$ & $86.0\pm0.4$ & $92.9\pm0.2$ \\
& 256 & $86.2\pm1.8$ & $95.6\pm0.5$ & $84.2\pm1.1$ & $85.7\pm0.5$ & $92.8\pm0.3$ \\

\midrule

\multirow{4}{*}{Init}
& 32  & $86.2\pm1.6$ & $95.4\pm0.6$ & $84.5\pm1.3$ & $86.7\pm0.3$ & $93.0\pm0.2$ \\
& 64  & $86.3\pm1.7$ & $95.5\pm0.5$ & $84.1\pm1.2$ & $86.4\pm0.3$ & $92.9\pm0.2$ \\
& 128 & $85.9\pm1.8$ & $95.6\pm0.5$ & $84.2\pm1.1$ & $86.1\pm0.3$ & $92.9\pm0.2$ \\
& 256 & $86.0\pm1.7$ & $95.6\pm0.4$ & $84.0\pm1.3$ & $85.8\pm0.5$ & $92.8\pm0.2$ \\

\midrule

\multirow{4}{*}{Dropout}
& 32  & $86.3\pm0.9$ & $95.8\pm0.3$ & $83.2\pm1.0$ & $86.8\pm0.3$ & $92.9\pm0.2$ \\
& 64  & $86.3\pm0.8$ & $95.9\pm0.3$ & $83.5\pm1.0$ & $86.5\pm0.4$ & $92.9\pm0.2$ \\
& 128 & $86.1\pm1.1$ & $95.8\pm0.3$ & $83.2\pm1.0$ & $86.1\pm0.3$ & $92.8\pm0.2$ \\
& 256 & $86.4\pm1.0$ & $95.8\pm0.3$ & $83.4\pm1.0$ & $85.8\pm0.4$ & $92.7\pm0.2$ \\

\midrule

\multirow{4}{*}{Shuffle}
& 32  & $86.4\pm1.2$ & $95.8\pm0.3$ & $83.2\pm1.2$ & $86.6\pm0.3$ & $92.9\pm0.2$ \\
& 64  & $86.2\pm1.1$ & $95.8\pm0.3$ & $83.6\pm0.9$ & $86.3\pm0.4$ & $92.8\pm0.2$ \\
& 128 & $86.3\pm0.9$ & $95.8\pm0.3$ & $83.9\pm0.9$ & $86.0\pm0.4$ & $92.7\pm0.2$ \\
& 256 & $86.3\pm1.0$ & $95.8\pm0.3$ & $83.9\pm1.0$ & $85.7\pm0.4$ & $92.7\pm0.2$ \\

\bottomrule
\end{tabular}

\end{table*}

Tab.~\ref{tab:accuracy_summary} reports the distributions of test accuracy across the 100 models along each investigated dimension, for all datasets and configurations.
On the tabular datasets, the four stochasticity configurations differ in the spread of their accuracy distributions. The \texttt{init} configuration consistently produces the widest distributions, reflecting substantial sensitivity to the choice of initial weights. The \texttt{shuffle} and \texttt{dropout} configurations yield more concentrated distributions, with most models clustering near the best accuracy of each pool.

This difference in spread has direct consequences for the construction of the empirical Rashomon set, since wider distributions may result in fewer models falling within a fixed $\epsilon$ of the best-performing model.

For the Fashion-MNIST and CIFAR-10 datasets trained with ResNet-20, the accuracy distributions remain highly concentrated across all stochasticity configurations. This indicates that the considered sources of randomness primarily influence the particular solution reached during training rather than the final level of predictive performance. As a consequence, different stochastic realizations consistently converge to models with comparable accuracy, suggesting the presence of large empirical Rashomon sets even for relatively small values of~$\epsilon$.

For the Fashion-MNIST and CIFAR10 datasets, the distributions are highly concentrated around similar mean values with only minor differences across stochastic configurations. 
This suggests that the learning problem is relatively stable, and that stochastic factors have a limited impact on both the central tendency and dispersion of performance. Notably, increasing the batch size does not significantly alter either the mean accuracy or its variability, indicating that the optimization landscape contains a broad region of near-optimal solutions that are consistently reachable.

Across both datasets, no single source of stochasticity consistently dominates in terms of accuracy distribution. 
This indicates that, although these factors may drive models toward different regions of the parameter space, they do not substantially affect the final predictive performance. 

However, these results highlight an important aspect of solution-space multiplicity: even when the distribution of accuracy is highly concentrated, the underlying models may still differ significantly in their internal representations and behavior. 
This further motivates the need to analyze multiplicity beyond considering also predictive variability and differences in input--output relationships.

\subsection{solution-space multiplicity}
\label{sec:results:size}

\newcommand{\figcell}[1]{%
    \includegraphics[width=0.24\textwidth]{__FIG/SIZE/#1}%
}

\begin{figure*}[!t]
\centering
\setlength{\tabcolsep}{2pt}
\renewcommand{\arraystretch}{1.1}
\scalebox{0.95}{%
\begin{tabular}{rcccc}
&
\textbf{BS=32} &
\textbf{BS=64} &
\textbf{BS=128} &
\textbf{BS=256}
\\
\textbf{\rotatebox{90}{ISOLET}}
&
\figcell{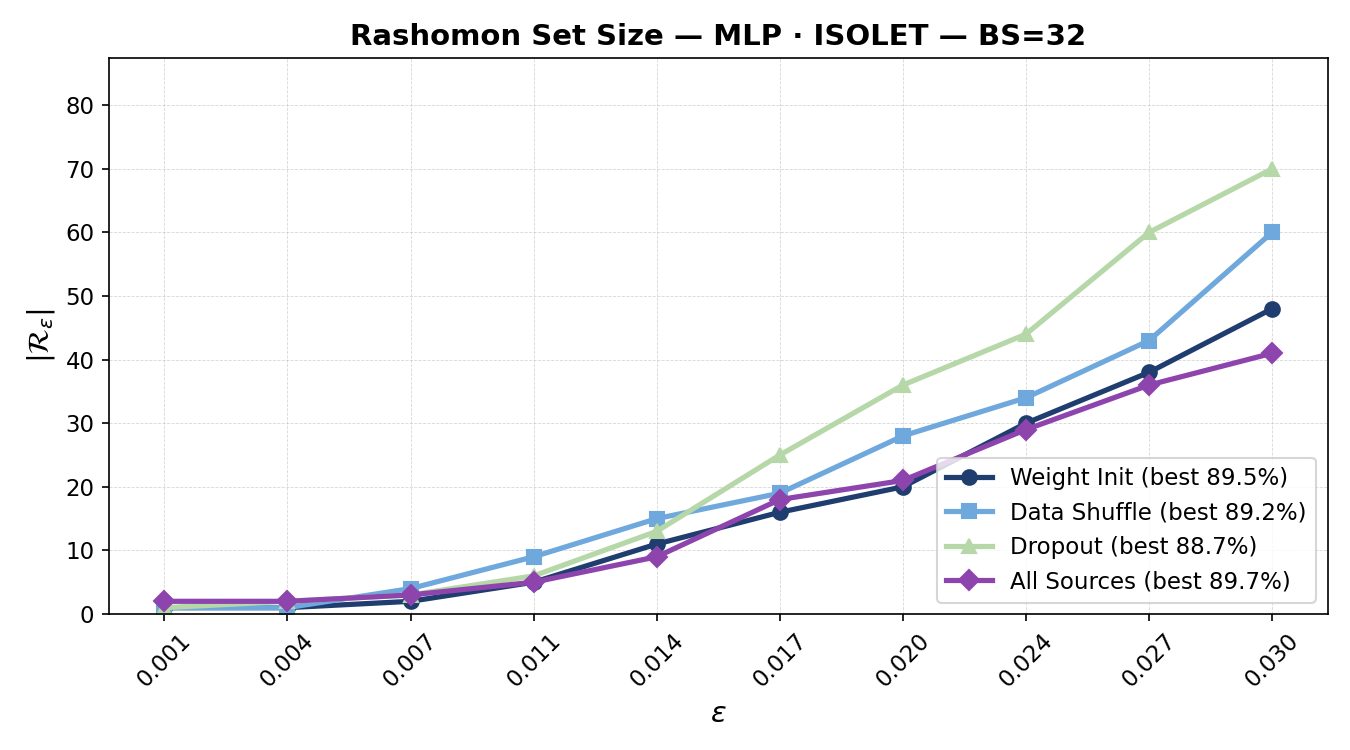}
&
\figcell{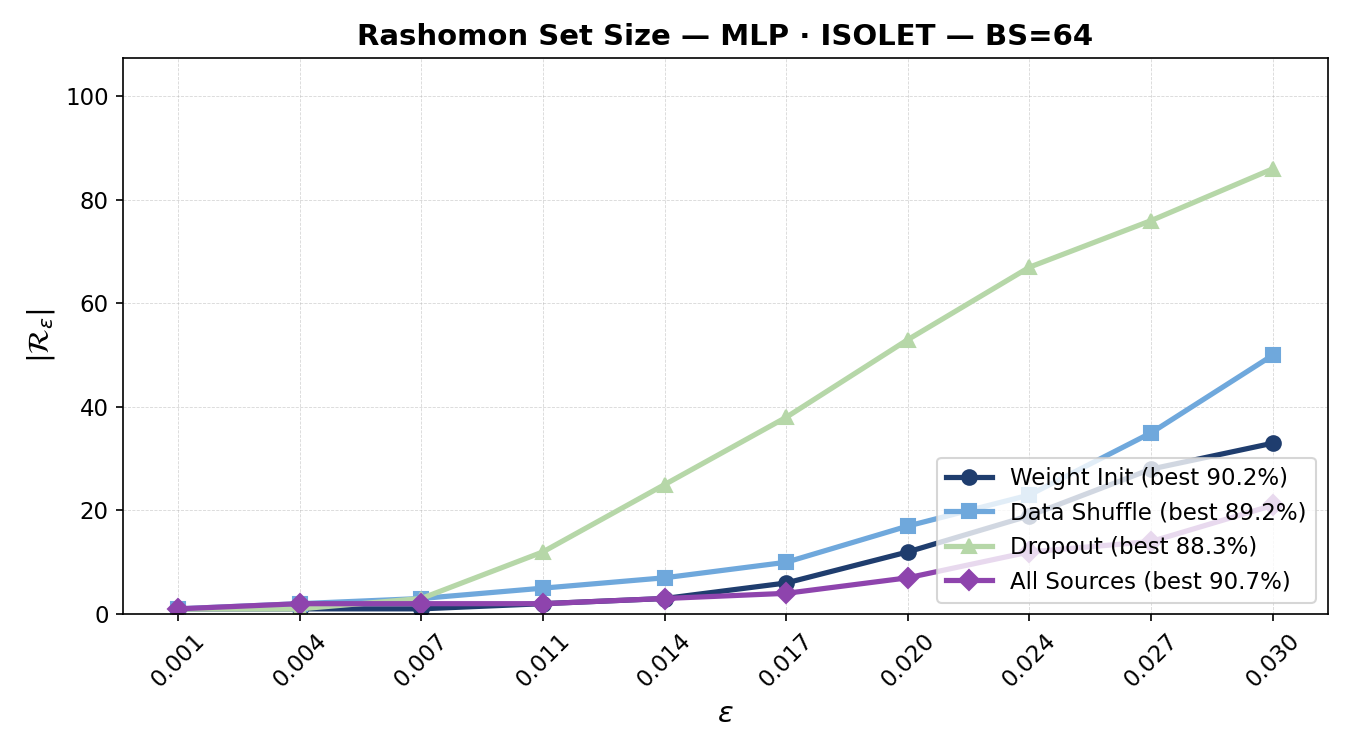}
&
\figcell{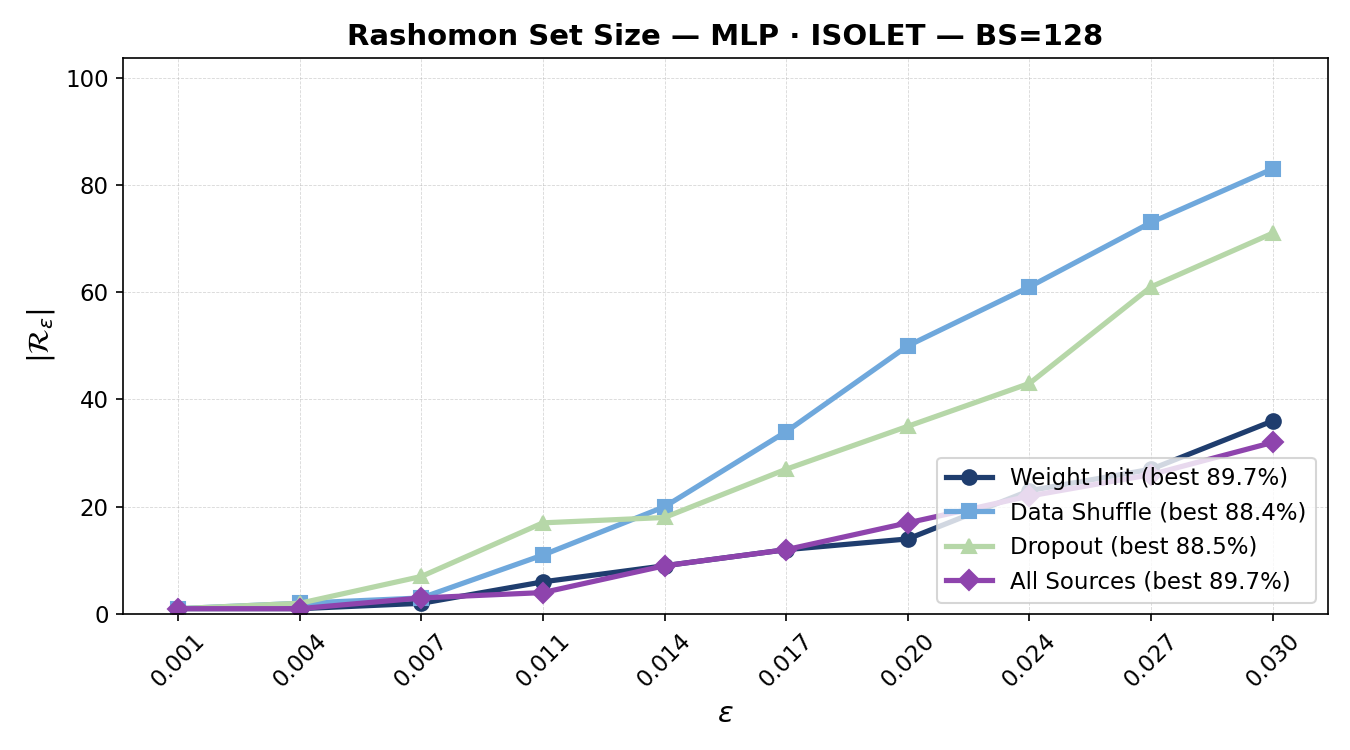}
&
\figcell{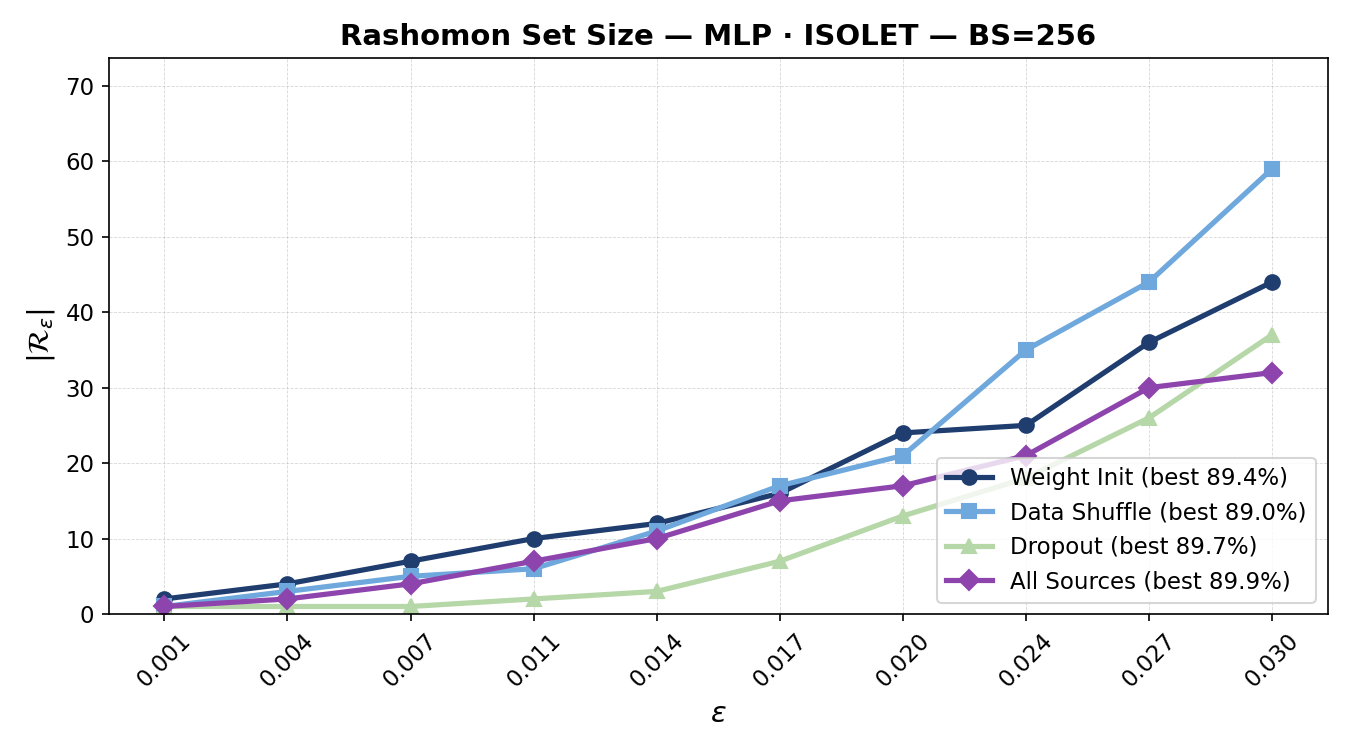}
\\
\textbf{\rotatebox{90}{Optical}}
&
\figcell{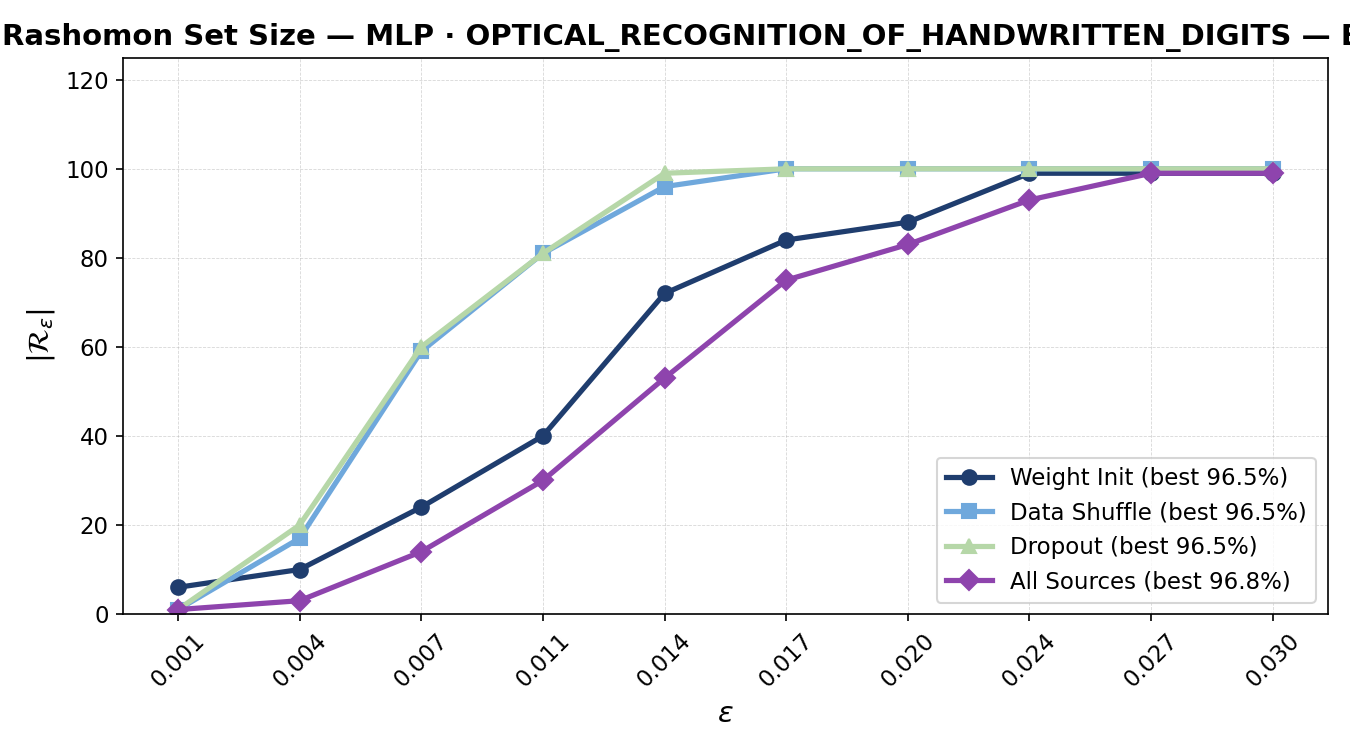}
&
\figcell{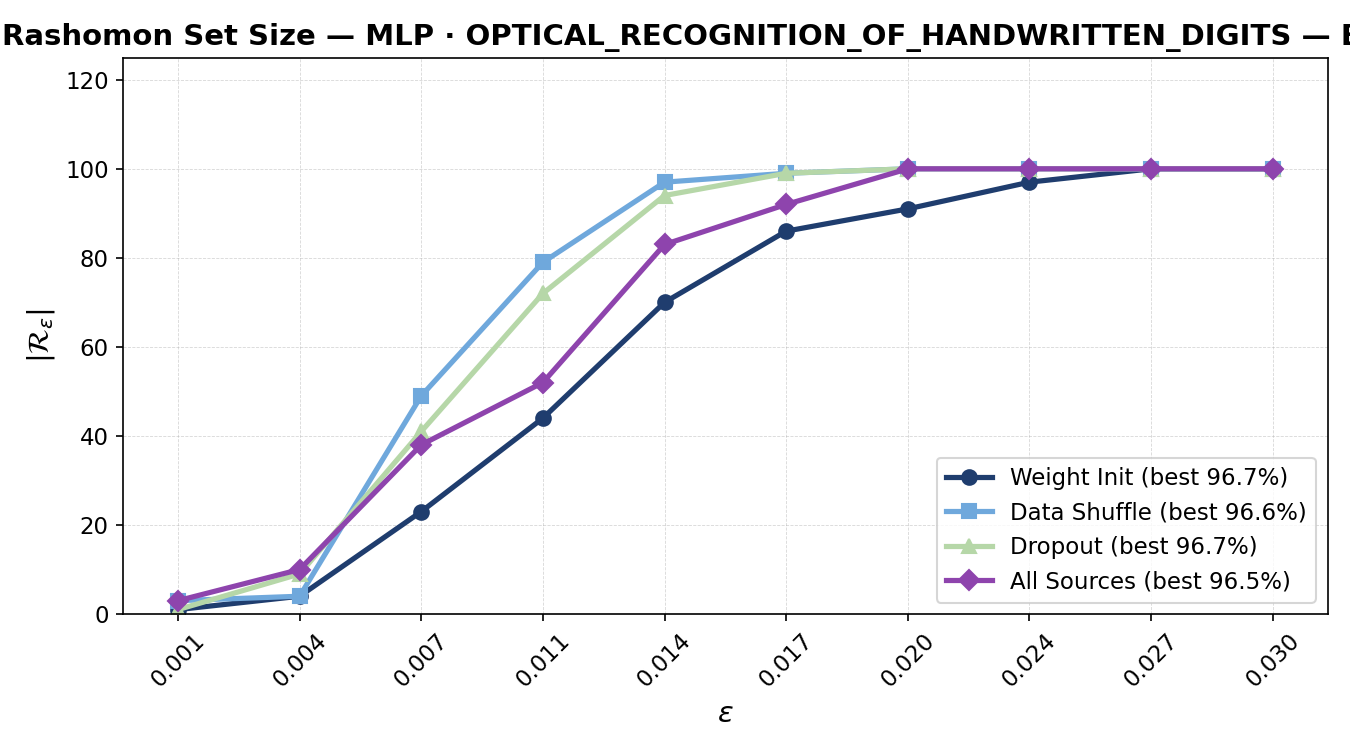}
&
\figcell{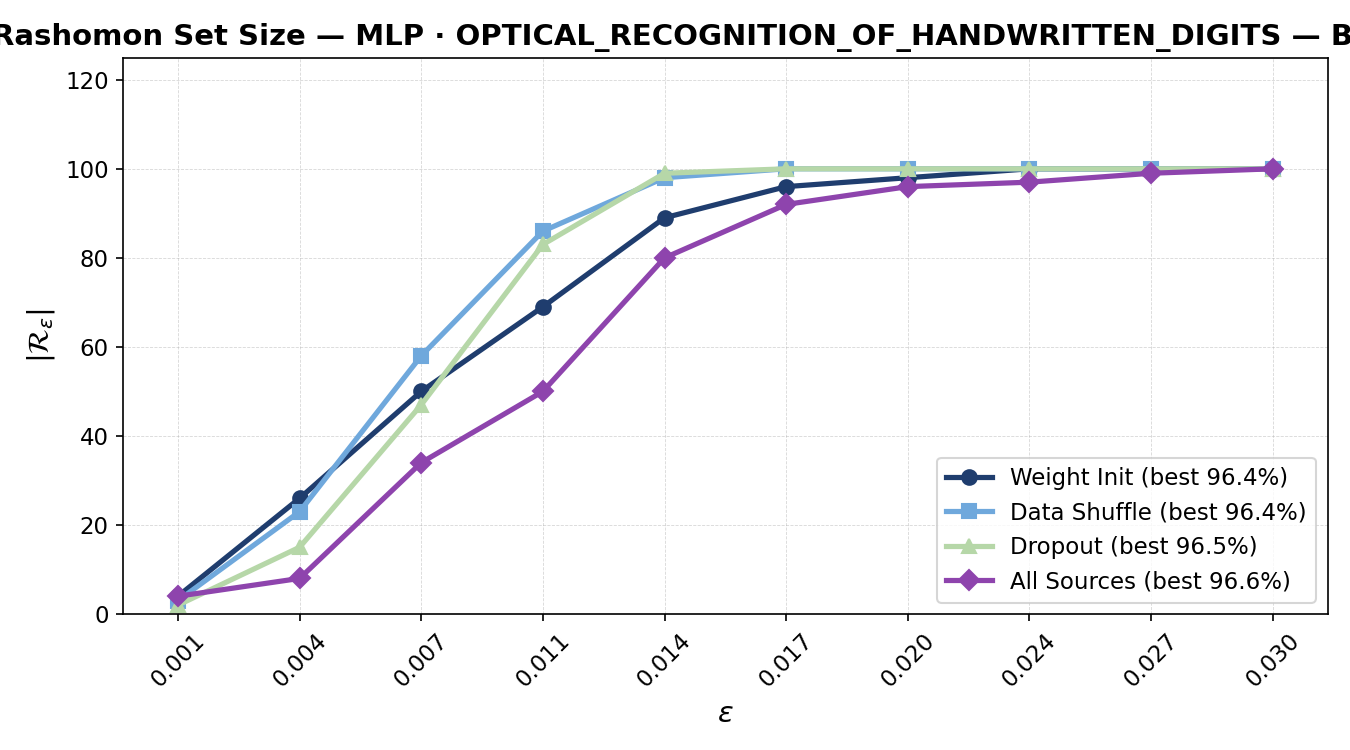}
&
\figcell{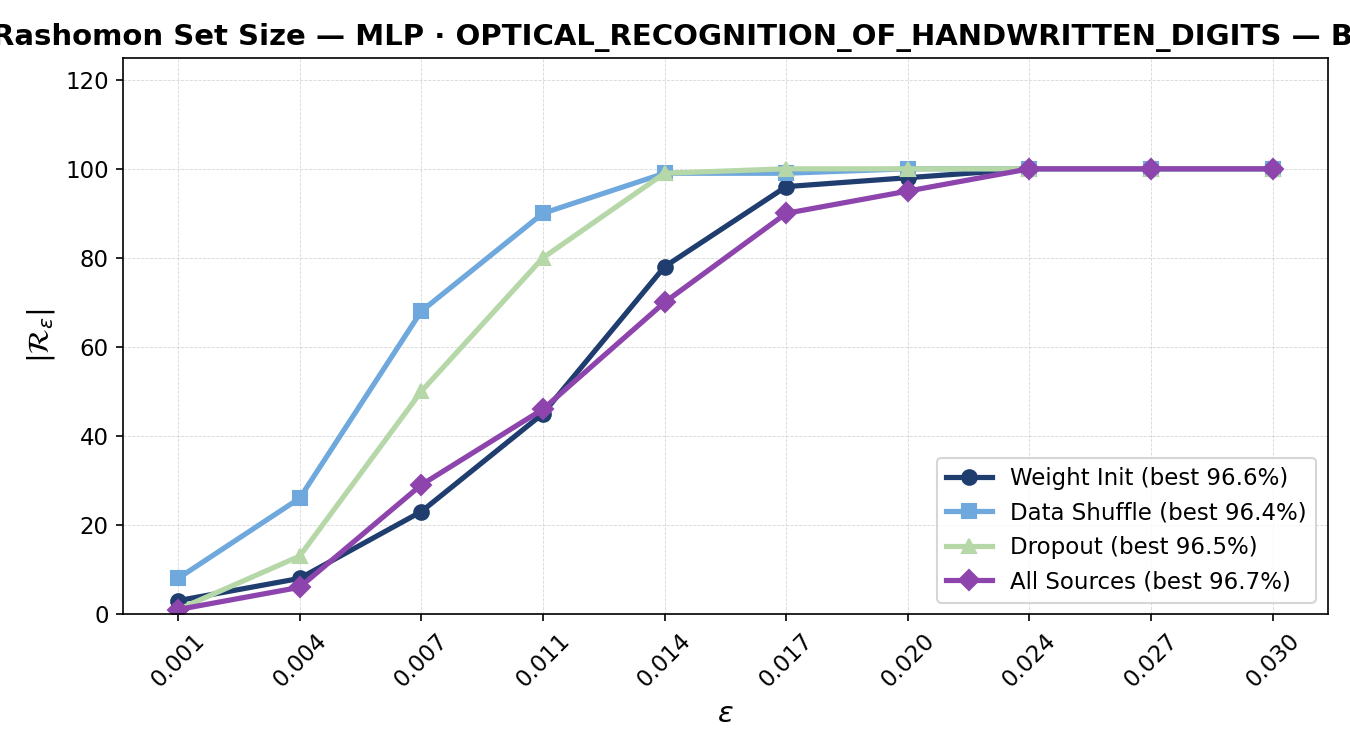}
\\
\textbf{\rotatebox{90}{Waveform}}
&
\figcell{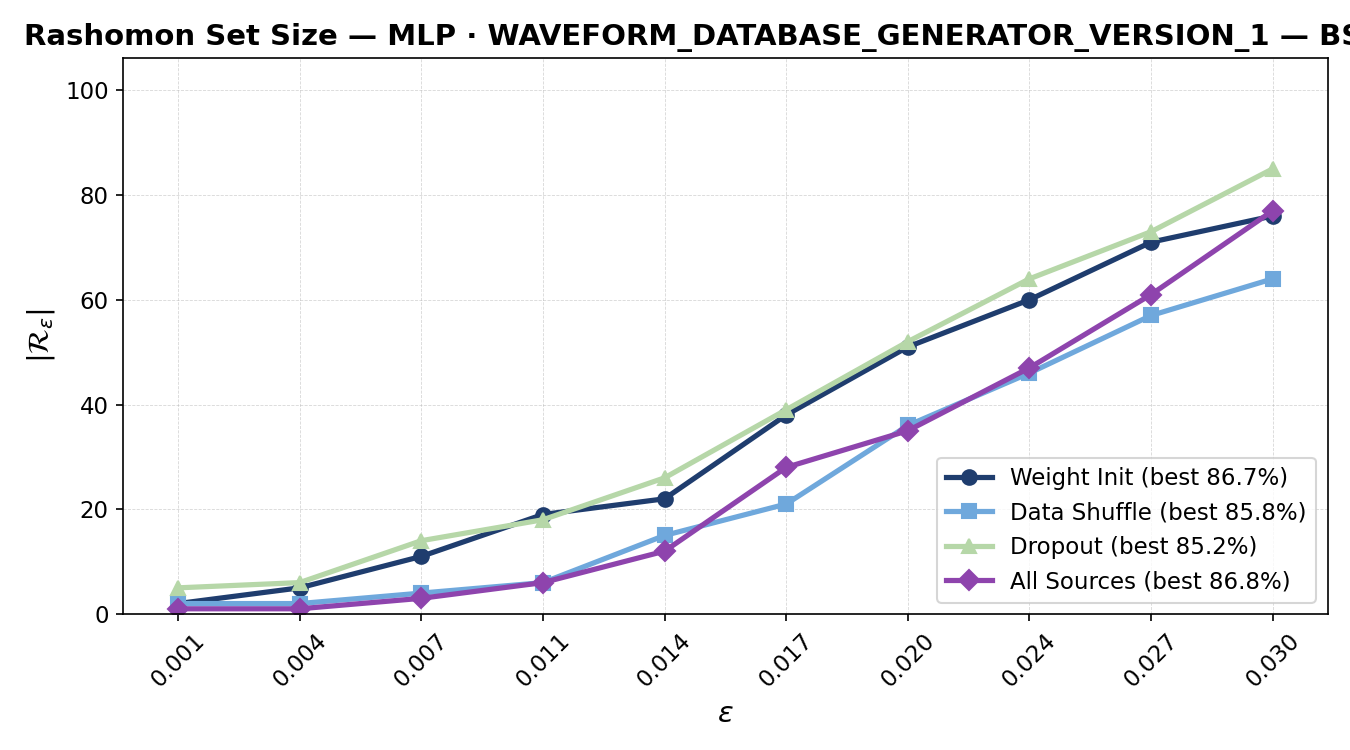}
&
\figcell{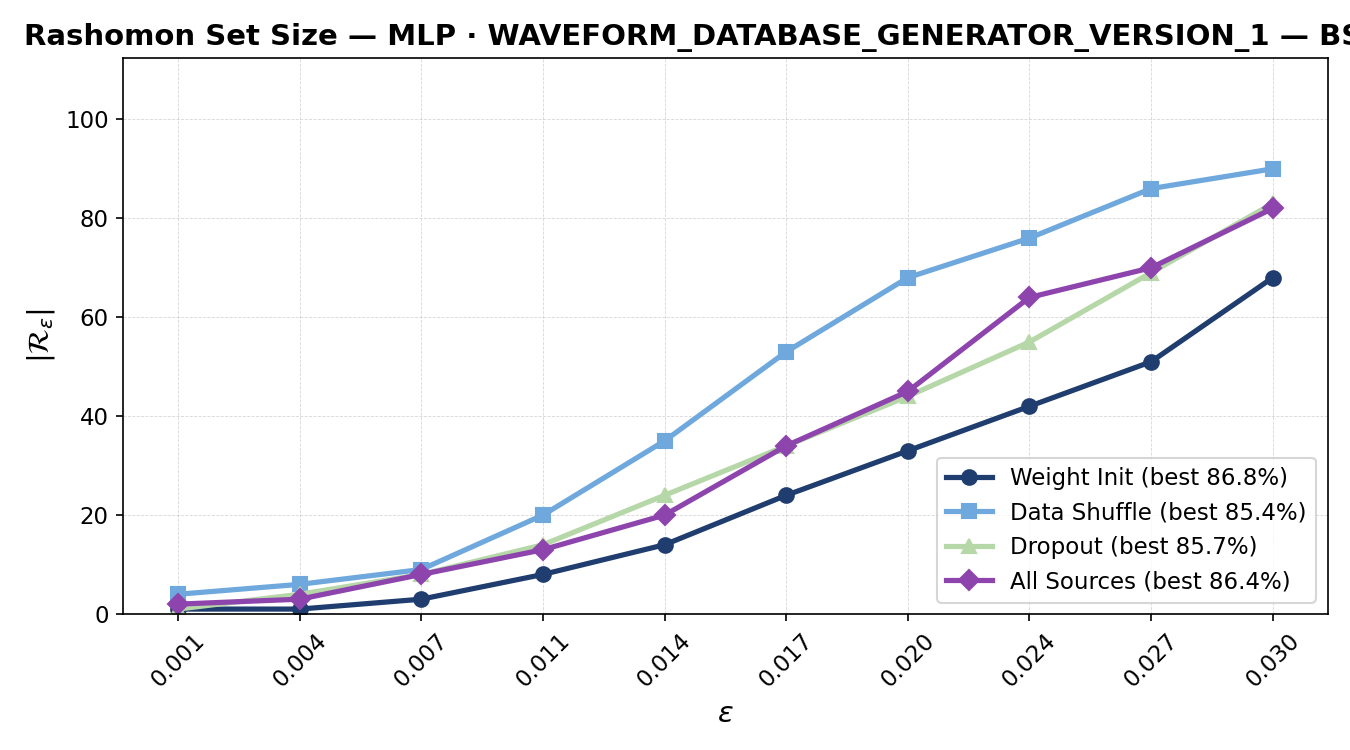}
&
\figcell{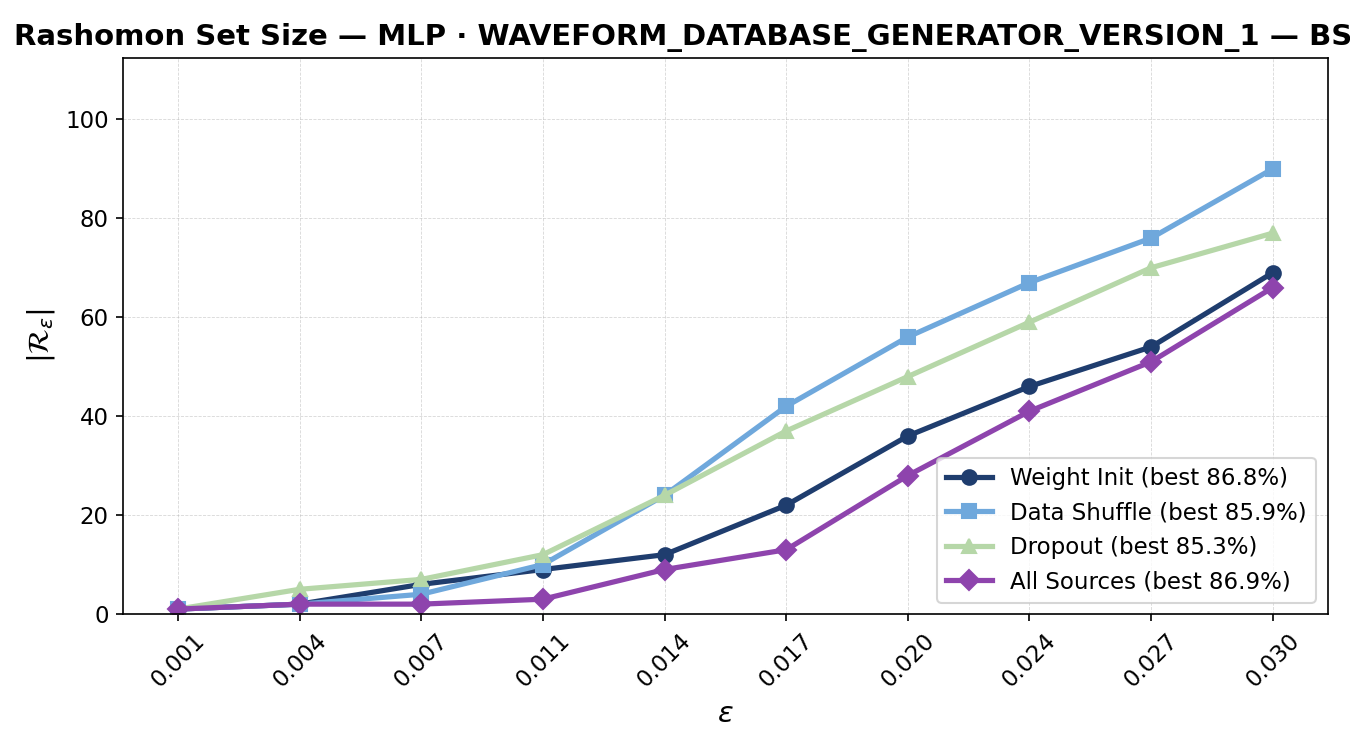}
&
\figcell{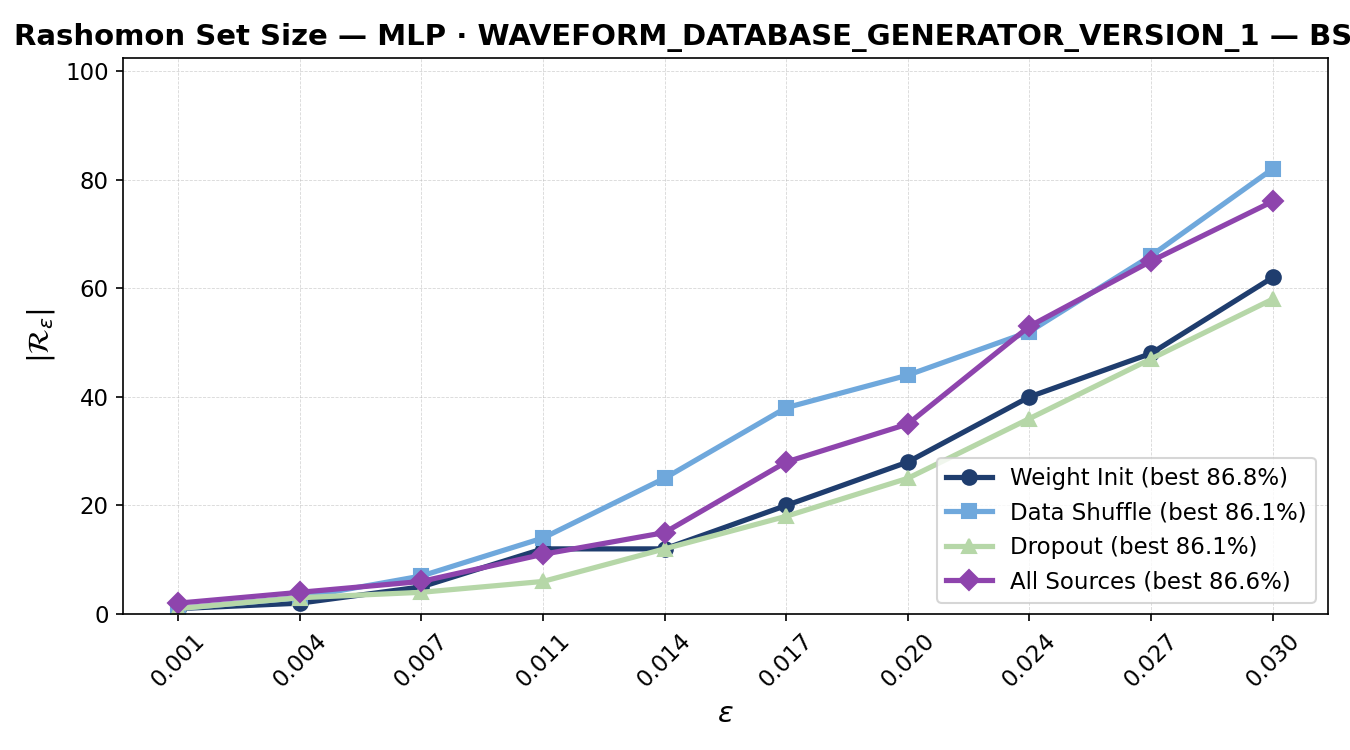}
\\
\end{tabular}
}
\caption{Rashomon set size as a function of $\epsilon$ for tabular datasets (MLP).
Rows correspond to datasets, columns to batch sizes.}
\label{fig:rashomon-size-tabular}
\end{figure*}

\begin{figure*}[t]
\centering
\setlength{\tabcolsep}{2pt}
\renewcommand{\arraystretch}{1.1}
\resizebox{\textwidth}{!}{%
\begin{tabular}{rcccc}
&
\textbf{BS=32} &
\textbf{BS=64} &
\textbf{BS=128} &
\textbf{BS=256}
\\
\textbf{\rotatebox{90}{CIFAR-10}}
&
\figcell{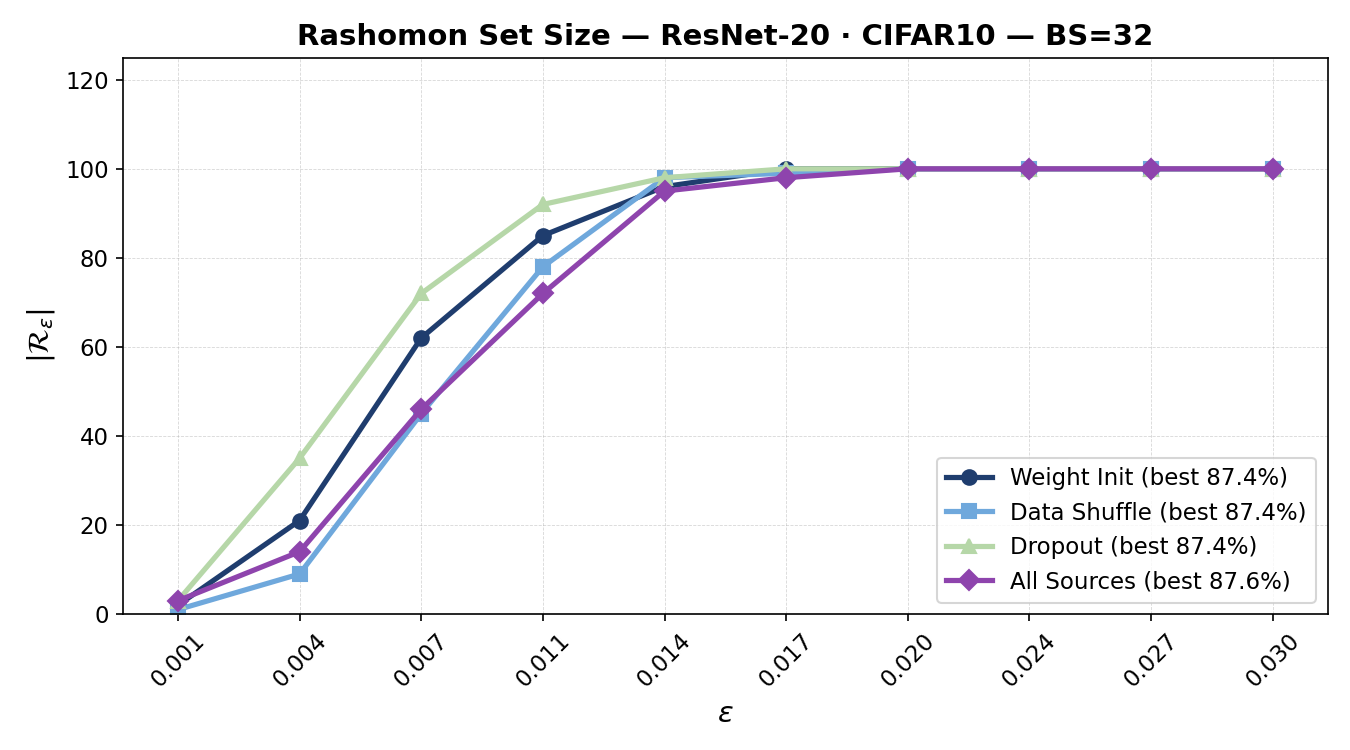}
&
\figcell{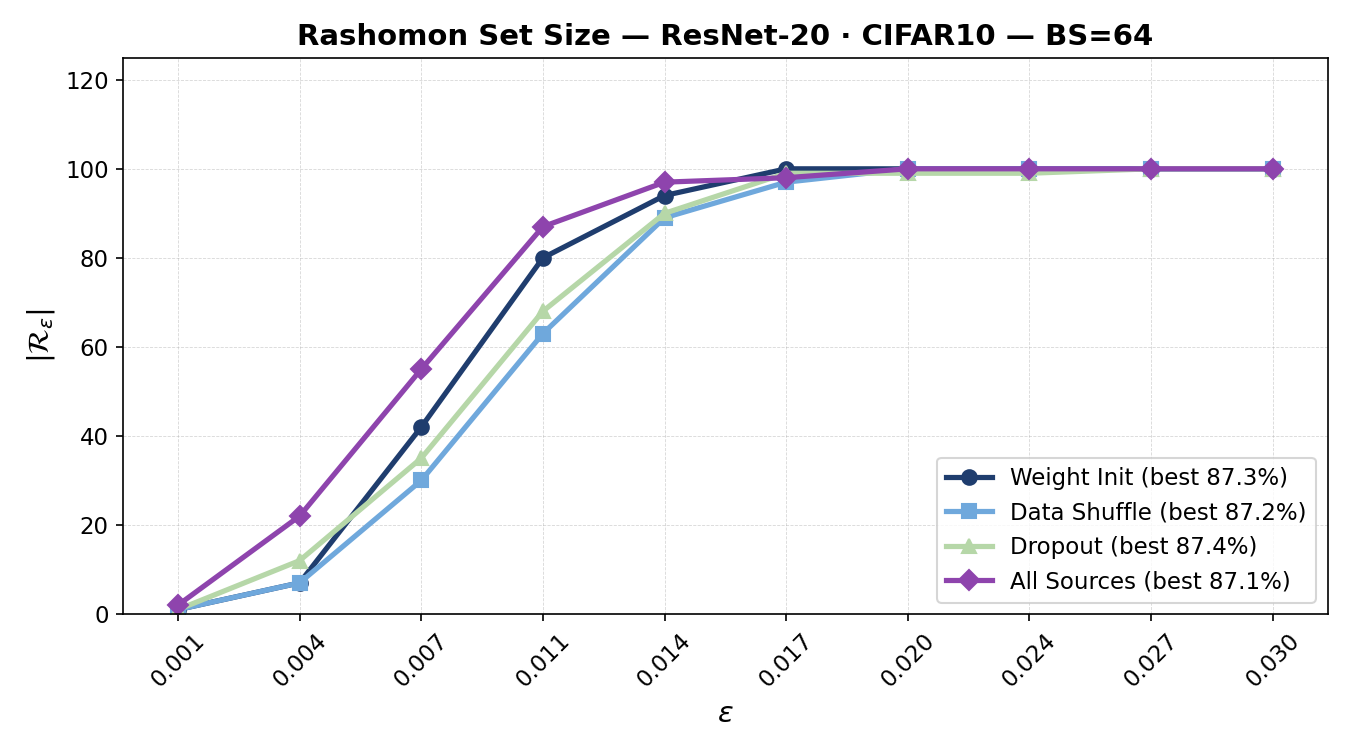}
&
\figcell{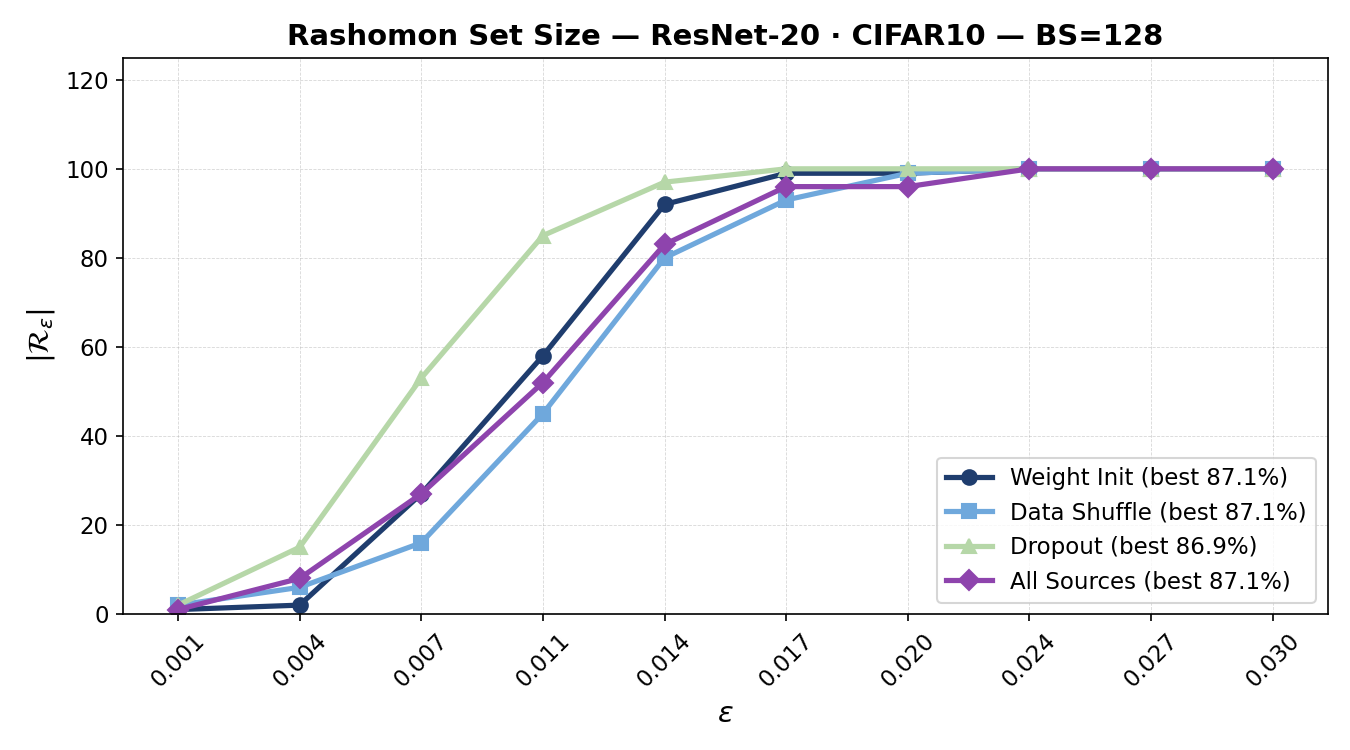}
&
\figcell{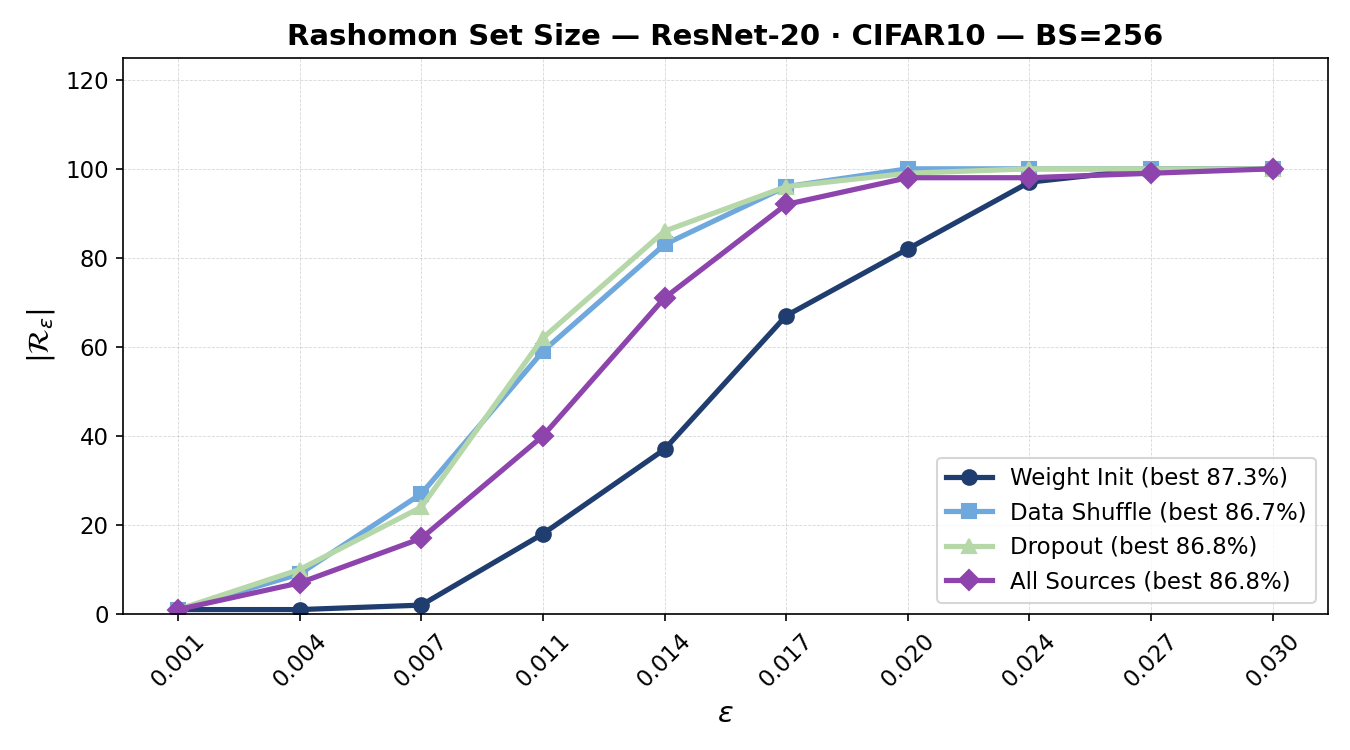}
\\
\textbf{\rotatebox{90}{Fashion-MNIST}}
&
\figcell{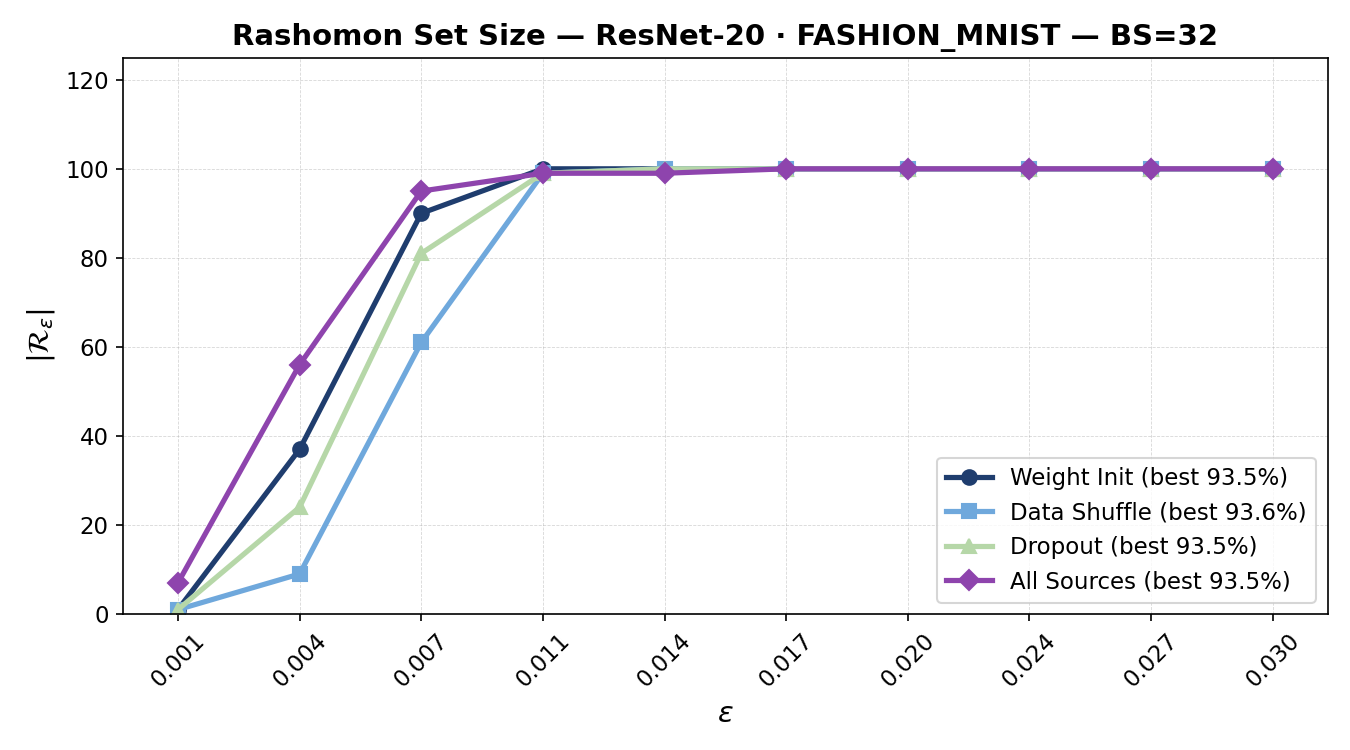}
&
\figcell{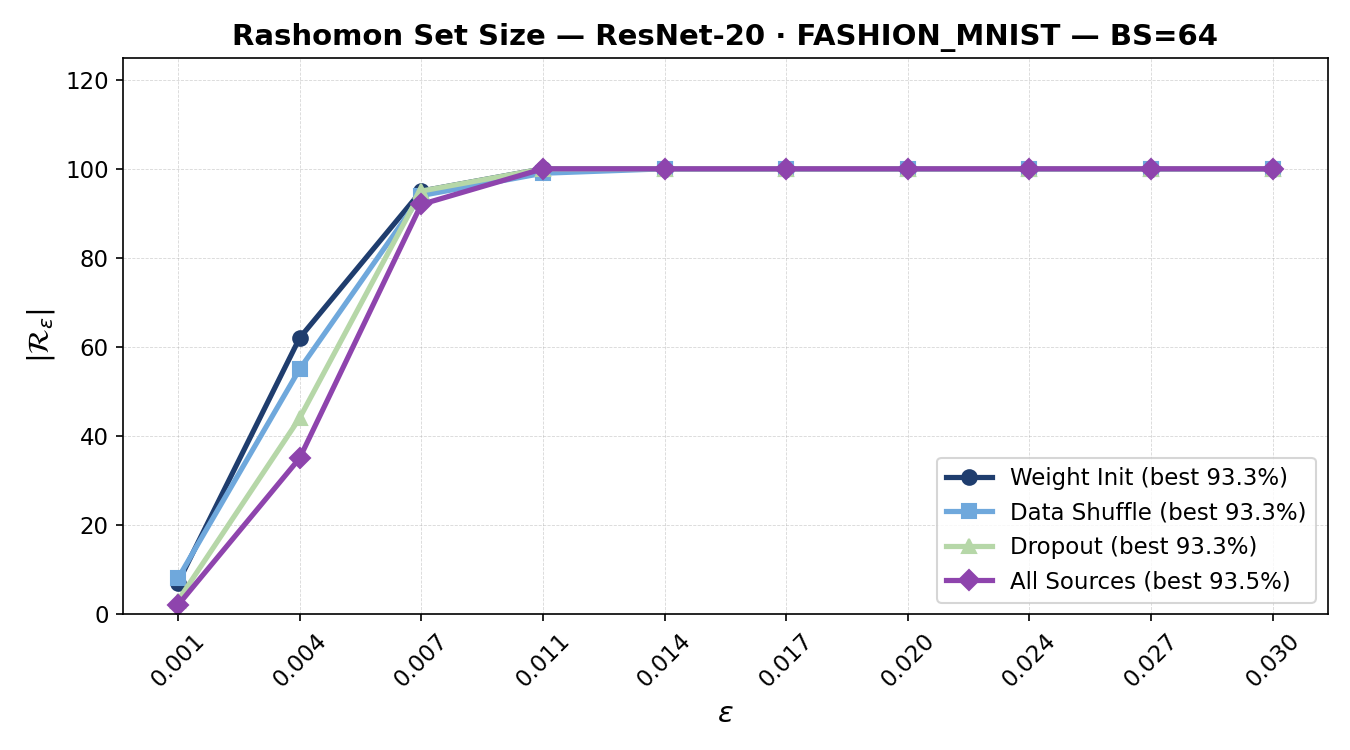}
&
\figcell{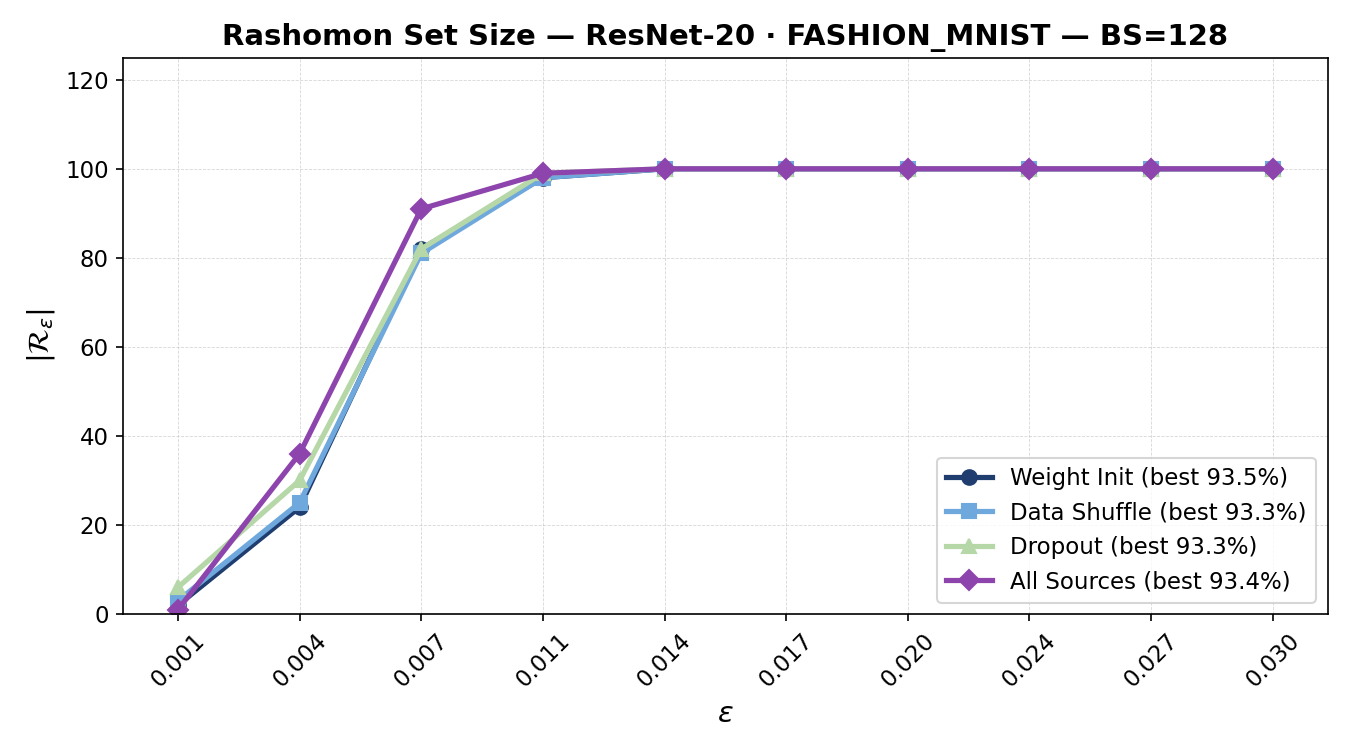}
&
\figcell{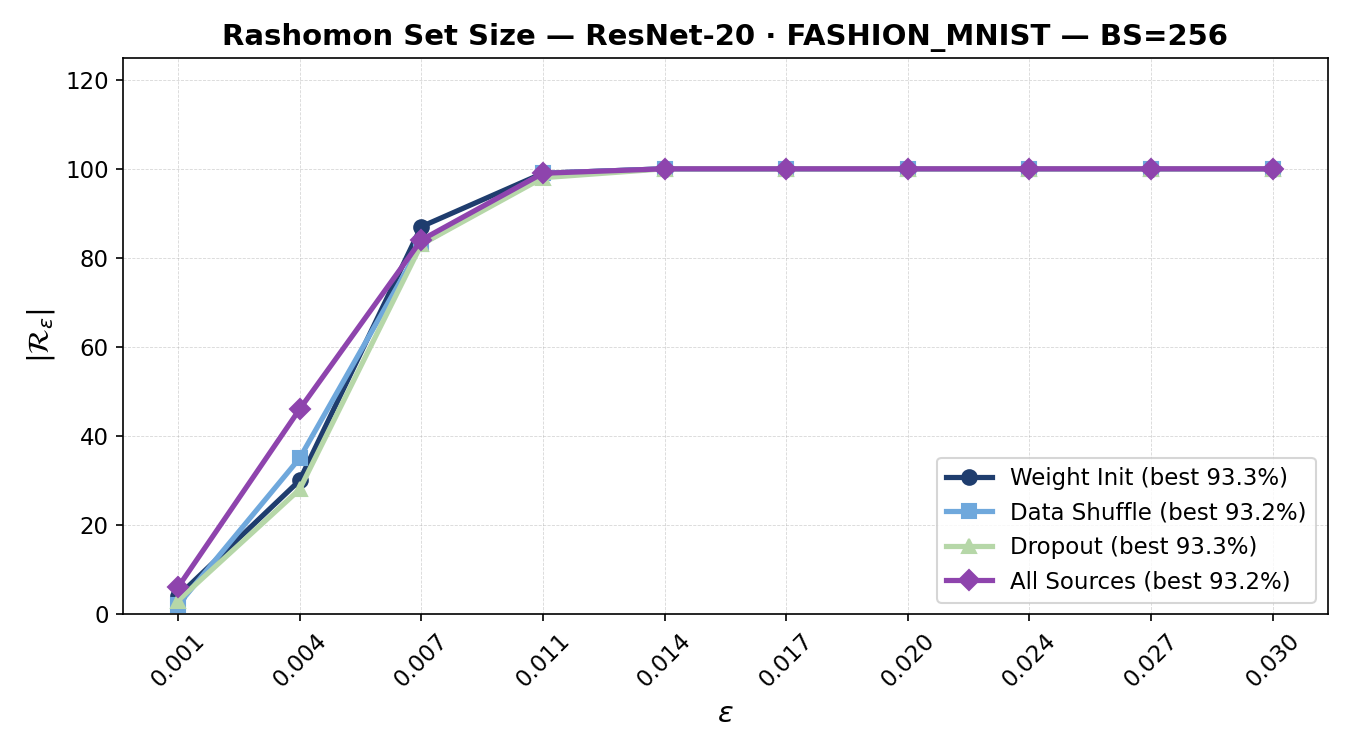}
\\
\end{tabular}
}
\caption{Rashomon set size as a function of $\epsilon$ for image datasets (ResNet-20).
Rows correspond to datasets, columns to batch sizes.}
\label{fig:rashomon-size-image}
\end{figure*}

Figs. \ref{fig:rashomon-size-tabular} and \ref{fig:rashomon-size-image} report $|\mathcal{R}_\epsilon|$ as a function of $\epsilon$ for each dataset, architecture, and stochasticity configuration.

On all three tabular datasets (Fig.  \ref{fig:rashomon-size-tabular}), $|\mathcal{R}_\epsilon|$ grows steadily with $\epsilon$, reflecting the wide spread of test accuracies within each pool (Tab. \ref{tab:accuracy_summary}).
In large part of cases, \texttt{shuffle} produces the largest Rashomon sets, followed by \texttt{dropout}, while \texttt{init} yields the smallest sets quite evenly.
This ordering can be due to the accuracy distributions (Tab. \ref{tab:accuracy_summary}): the wider spread of \texttt{init} implies that fewer models fall within any fixed $\epsilon$ of the best model, whereas the more concentrated distributions of \texttt{shuffle} and \texttt{dropout} result in more models qualifying at the same $\epsilon$.
The \texttt{all} configuration tracks \texttt{init} closely in most settings, indicating that weight initialisation is the dominant factor when all sources of randomness vary simultaneously.

For image datasets (Fig. \ref{fig:rashomon-size-image}) a first observation is that dropout and init configurations exhibit a rapid growth of the Rashomon set size as $\epsilon$ increases. 
However, even for relatively small tolerance values, a large fraction of the trained models already satisfies the Rashomon criterion, confirming that the solution space contains many near-optimal models.

For Fashion-MNIST, the Rashomon set size grows extremely rapidly 
reaching nearly the full model pool even for small values of $\epsilon$. Moreover, the curves corresponding to different stochastic configurations almost completely overlap across all batch sizes indicating that, for this relatively simple task, the learning problem admits a large and dense region of near-optimal solutions that is largely insensitive to the specific source of stochasticity.

In contrast, CIFAR10 exhibits a more gradual growth with the curves corresponding to different stochastic configurations more clearly separated. This suggests that different sources of stochasticity leads toward solutions with different performance.

Notably, dropout tends to produce larger Rashomon sets for smaller values of $\epsilon$, suggesting that stochastic regularization facilitates exploration of multiple near-optimal solutions.
Conversely, weight initialization and data shuffling show more variable behavior, especially at larger batch sizes, where their impact on the reachable solution space becomes more pronounced.

Across both datasets, increasing the batch size tends to slightly reduce the growth rate of the Rashomon set, particularly in the CIFAR10 setting.  This suggests that larger batch sizes are associated with a more concentrated subset of near-optimal solutions, although the effect is relatively mild and dataset-dependent.

\subsection{Predictive Multiplicity}
\newcommand{\ambcell}[1]{%
    \includegraphics[width=0.26\textwidth]{__FIG/AMBIGUITY/#1}%
}

\begin{figure*}[!t]
\centering
\setlength{\tabcolsep}{2pt}
\renewcommand{\arraystretch}{1.1}
\scalebox{0.9}{%
\begin{tabular}{rcccc}
&
\textbf{BS=32} &
\textbf{BS=64} &
\textbf{BS=128} &
\textbf{BS=256}
\\
\textbf{\rotatebox{90}{ISOLET}}
&
\ambcell{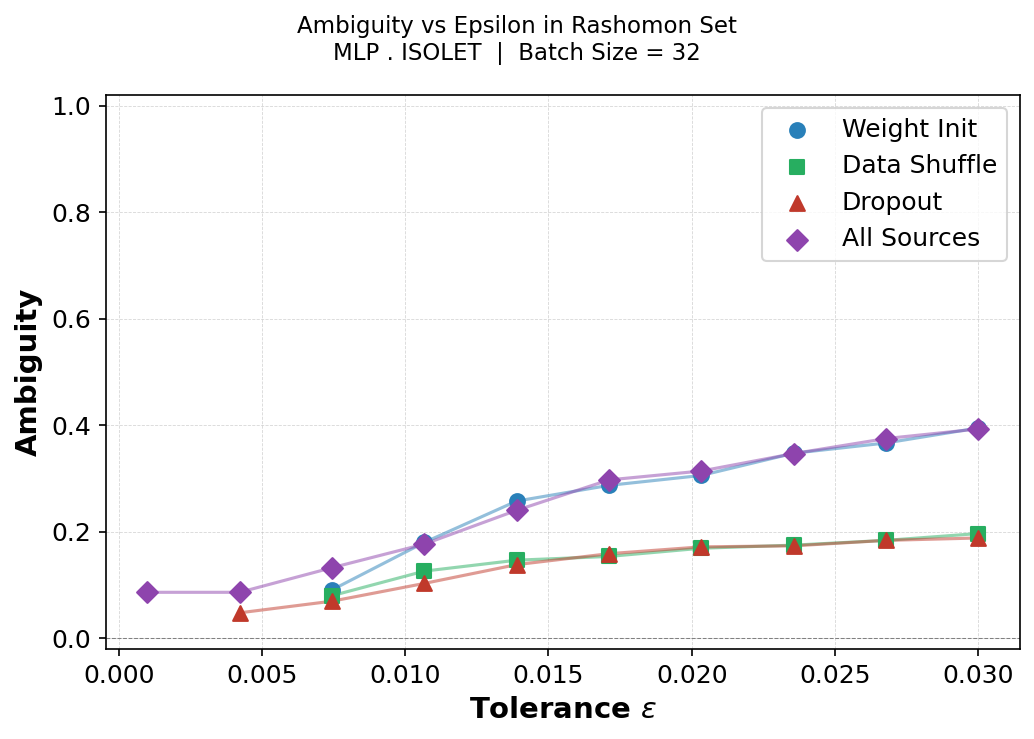}
&
\ambcell{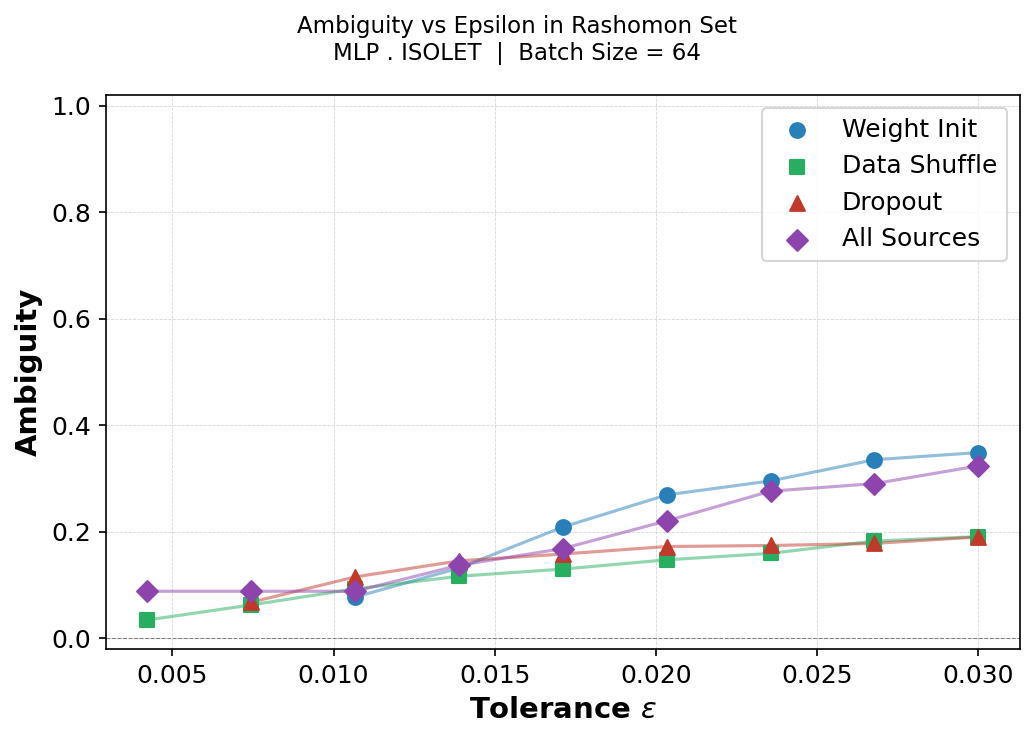}
&
\ambcell{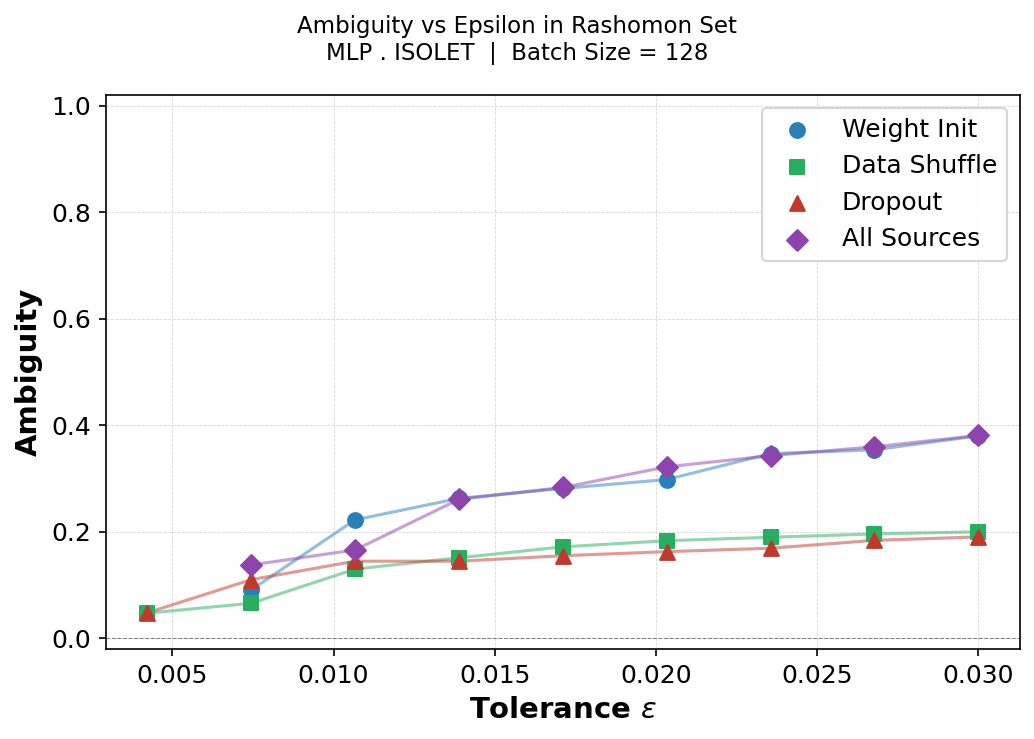}
&
\ambcell{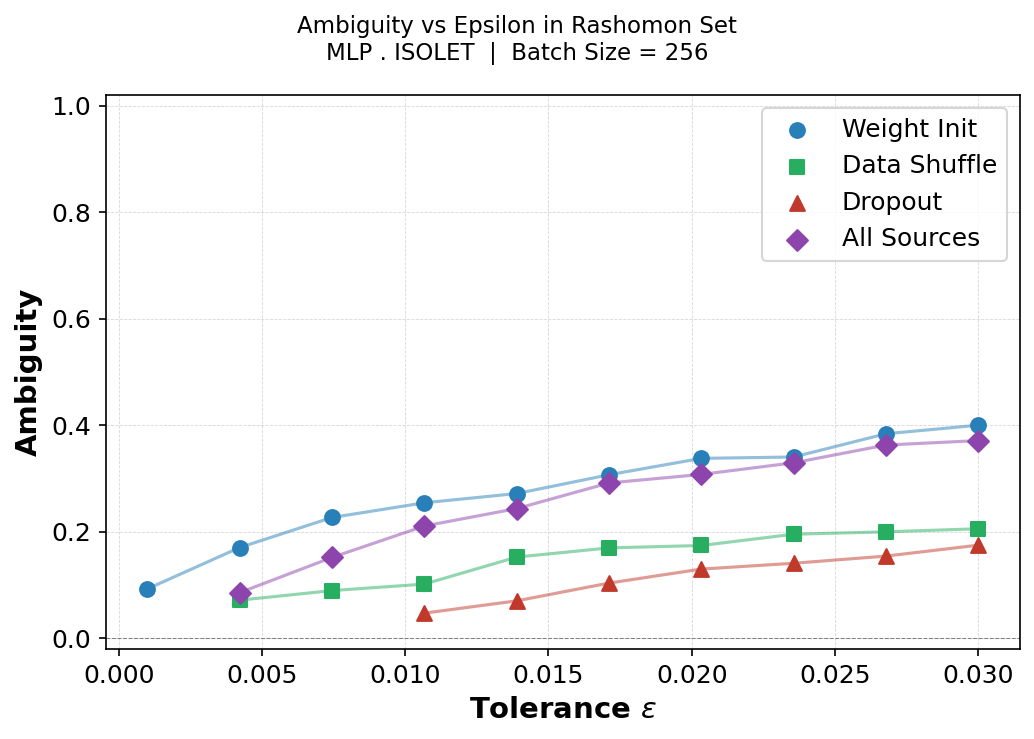}
\\
\textbf{\rotatebox{90}{Optical}}
&
\ambcell{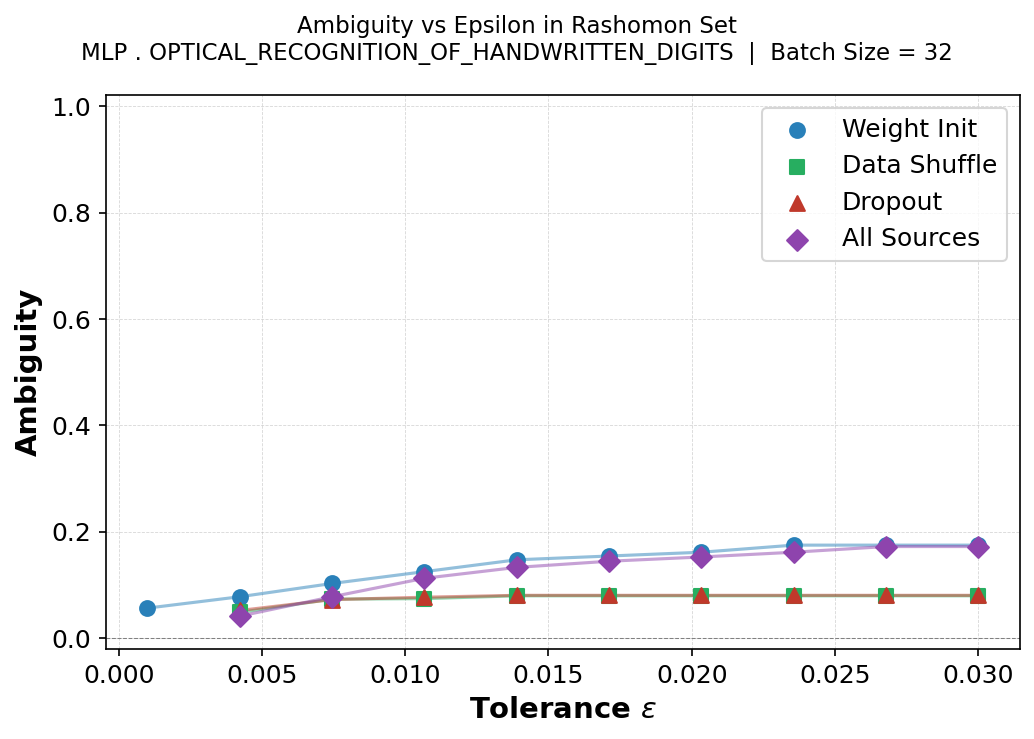}
&
\ambcell{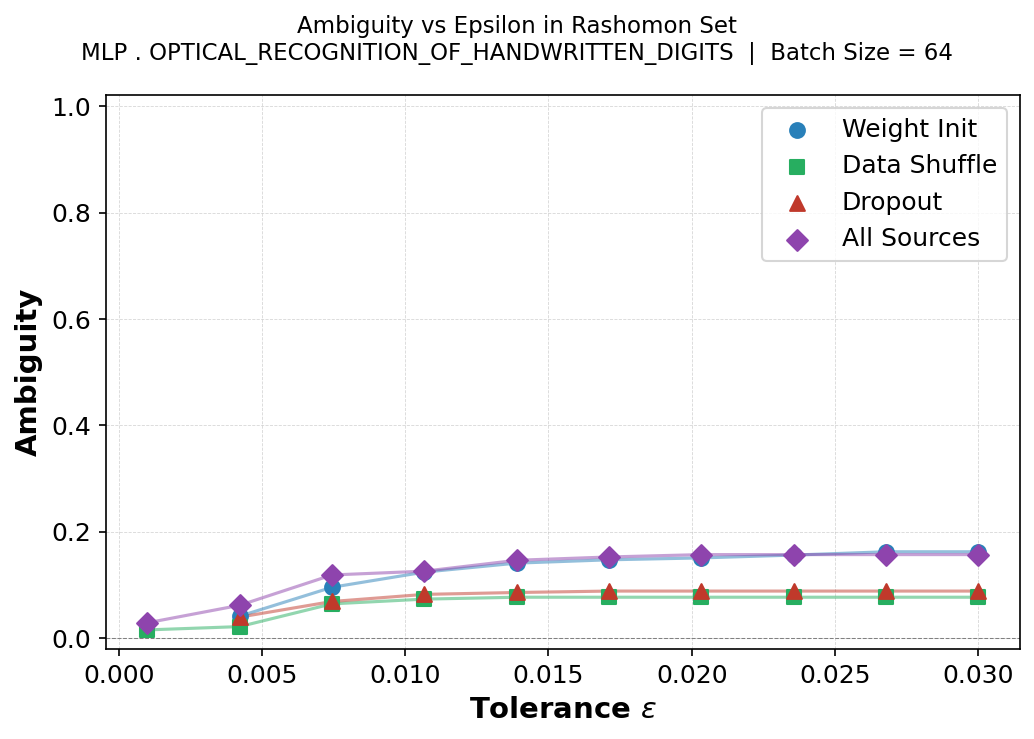}
&
\ambcell{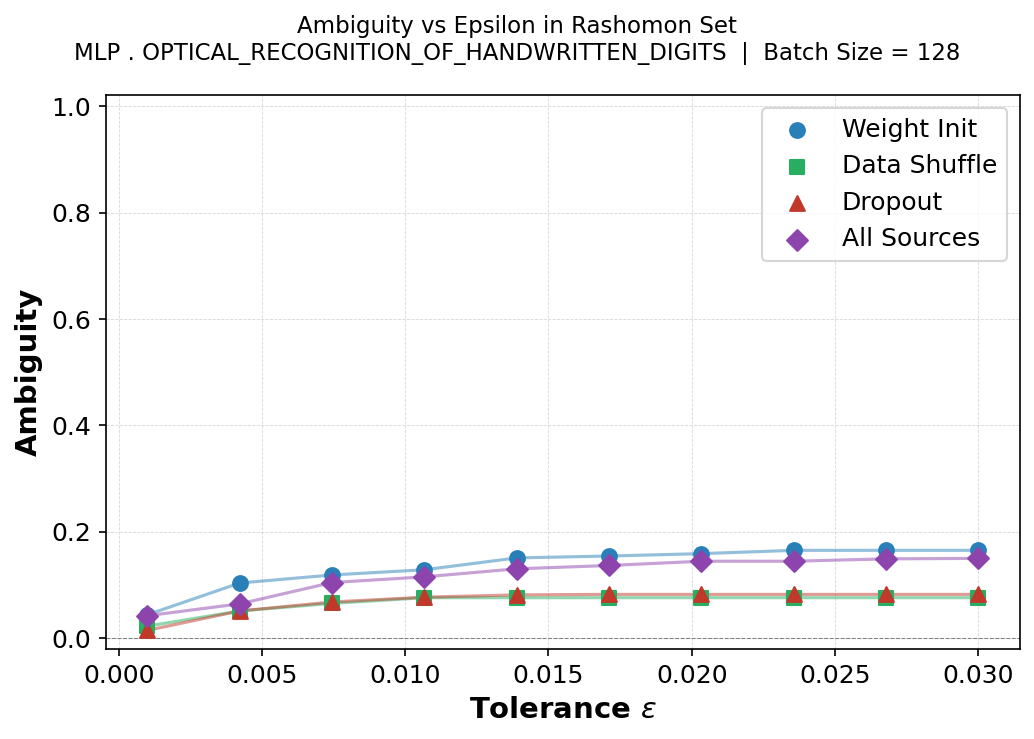}
&
\ambcell{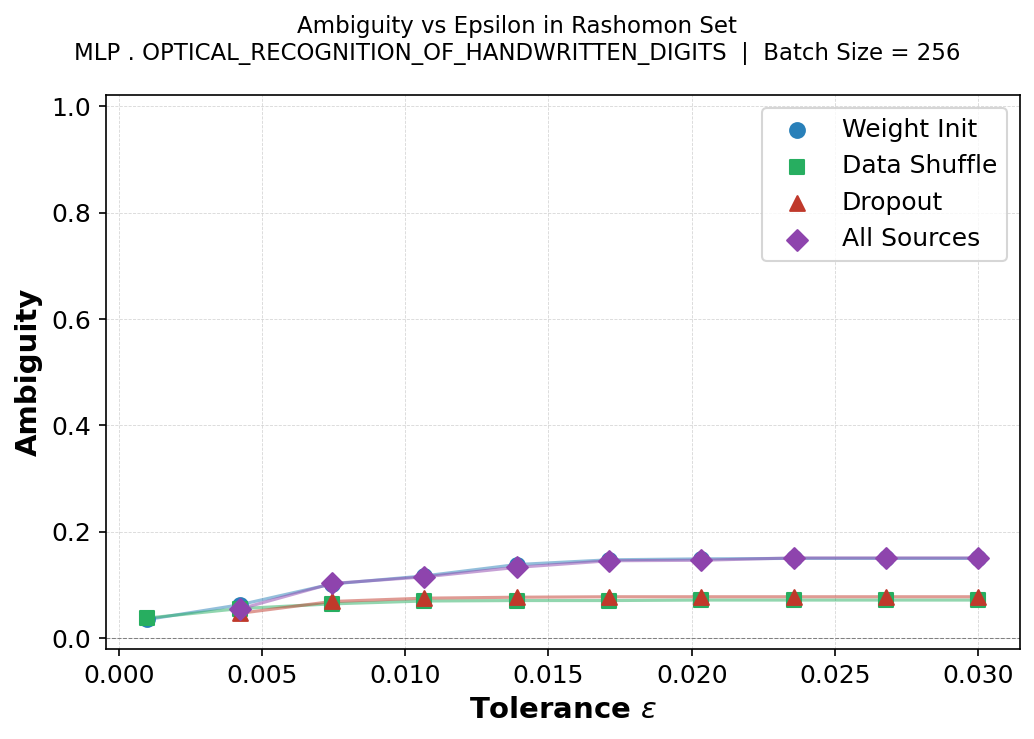}
\\
\textbf{\rotatebox{90}{Waveform}}
&
\ambcell{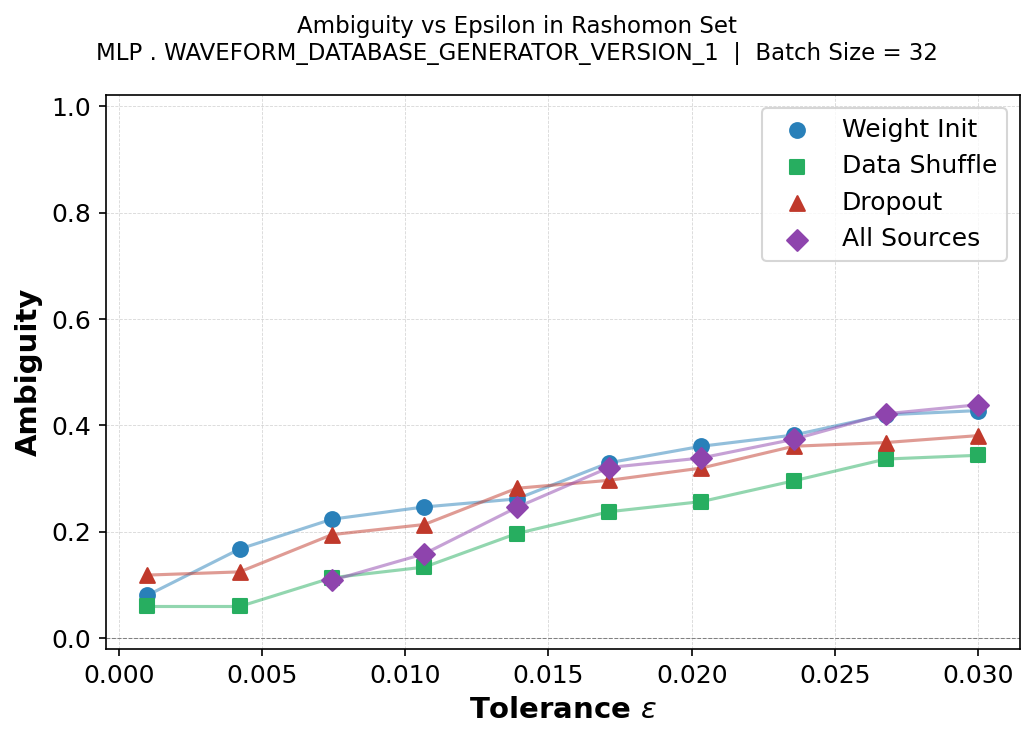}
&
\ambcell{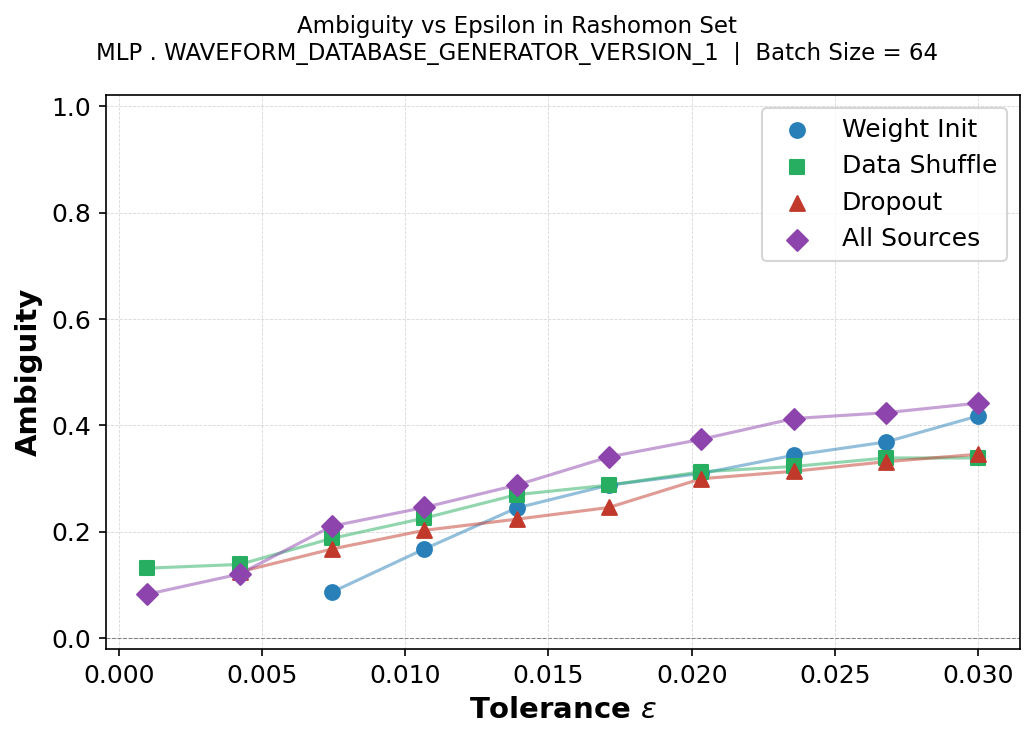}
&
\ambcell{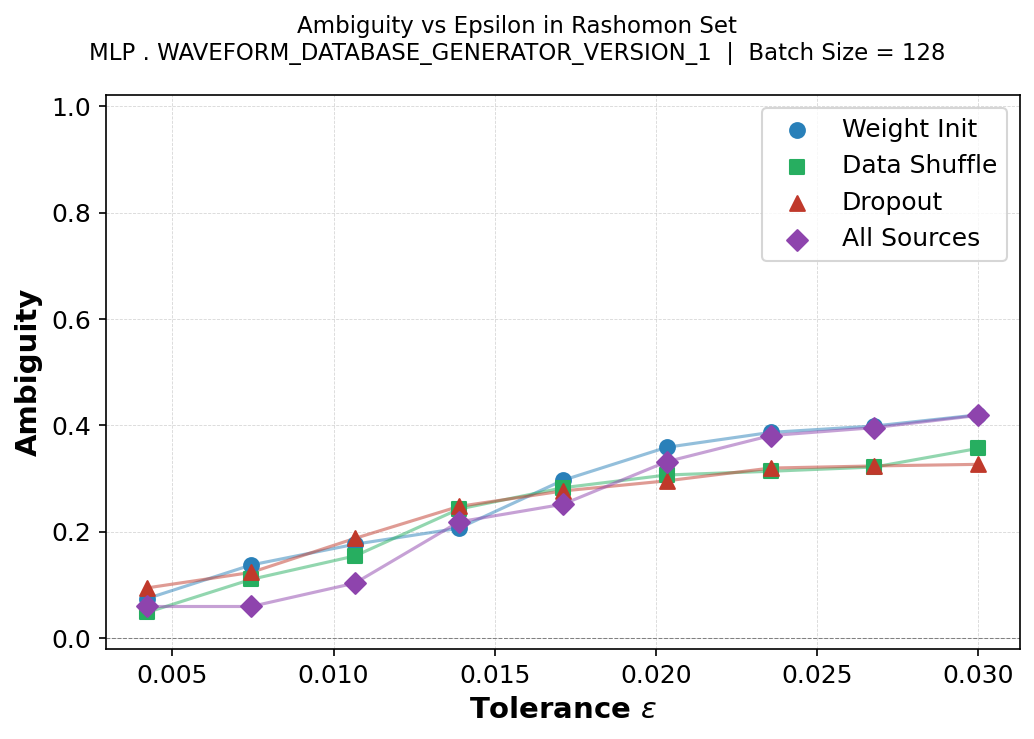}
&
\ambcell{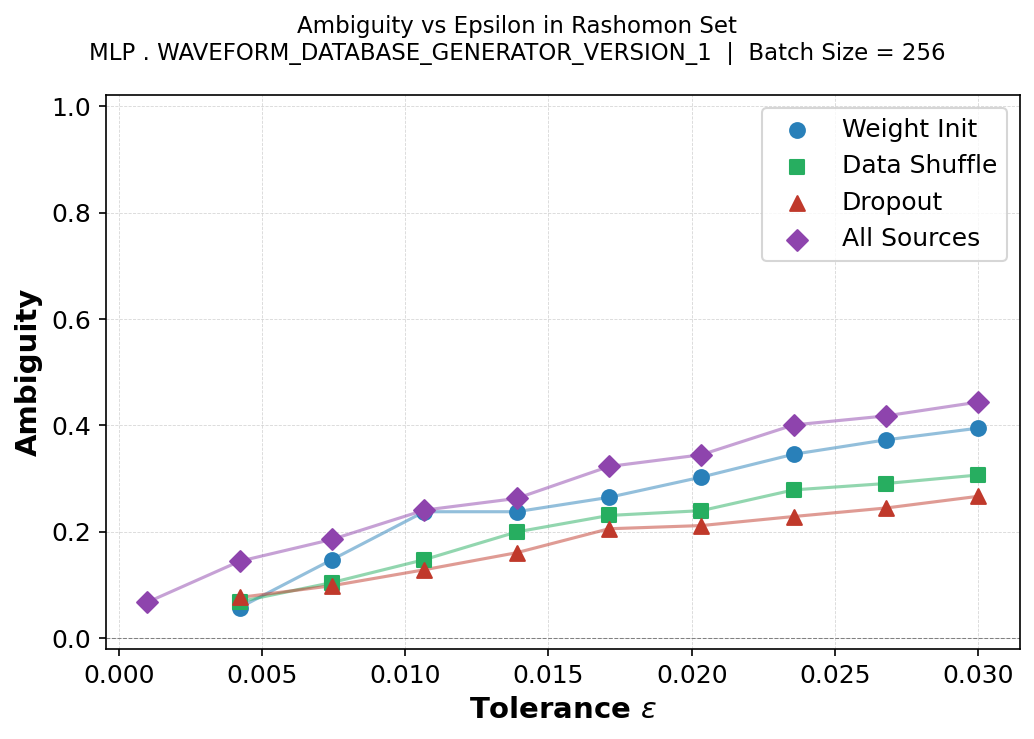}
\\
\end{tabular}
}
\caption{Ambiguity as a function of $\epsilon$ for tabular datasets (MLP), broken down by stochasticity source. Rows correspond to datasets, columns to batch sizes.}
\label{fig:ambiguity-tabular}
\end{figure*}

\begin{figure*}[!t]
\centering
\setlength{\tabcolsep}{2pt}
\renewcommand{\arraystretch}{1.1}
\scalebox{0.85}{%
\begin{tabular}{rcccc}
&
\textbf{BS=32} &
\textbf{BS=64} &
\textbf{BS=128} &
\textbf{BS=256}
\\
\textbf{\rotatebox{90}{CIFAR-10}}
&
\ambcell{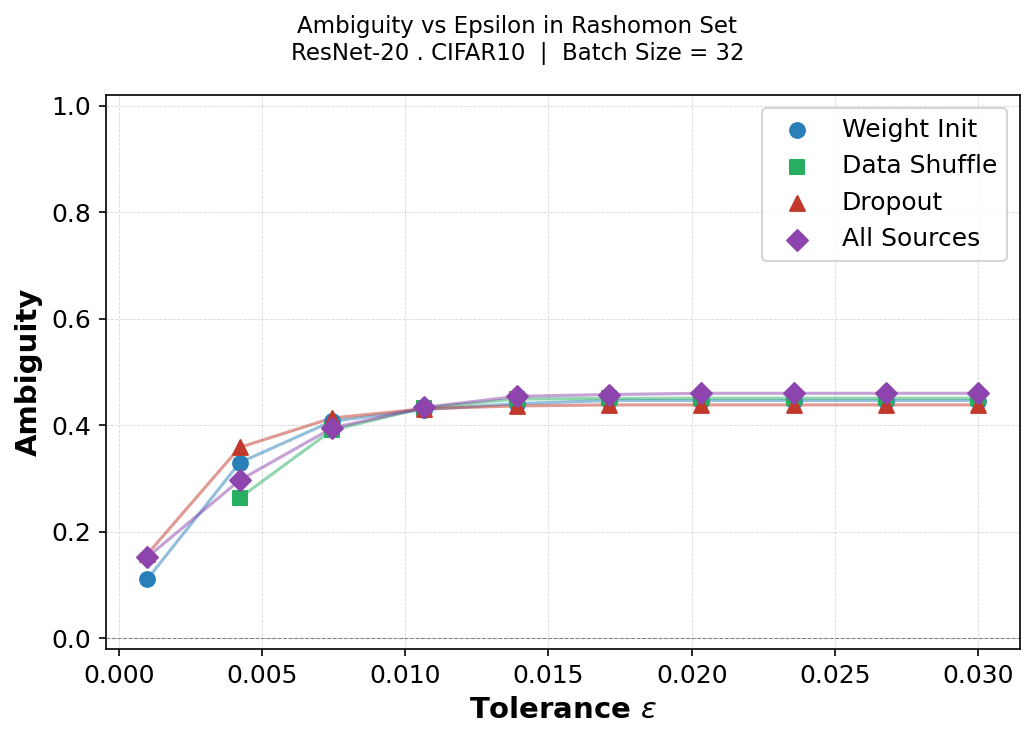}
&
\ambcell{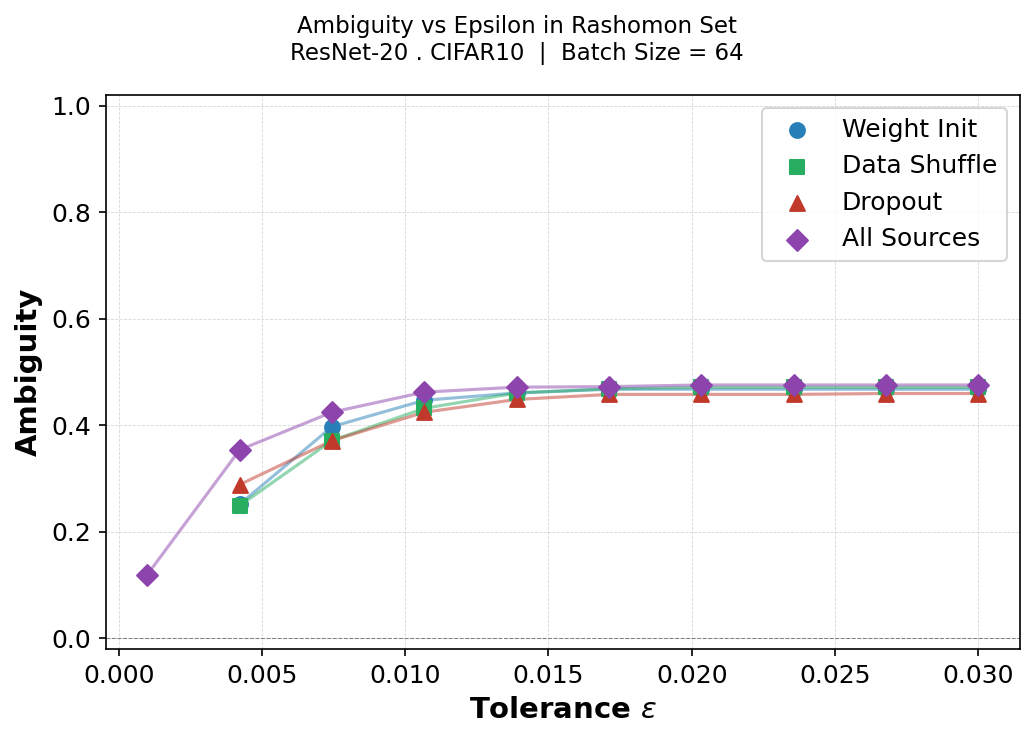}
&
\ambcell{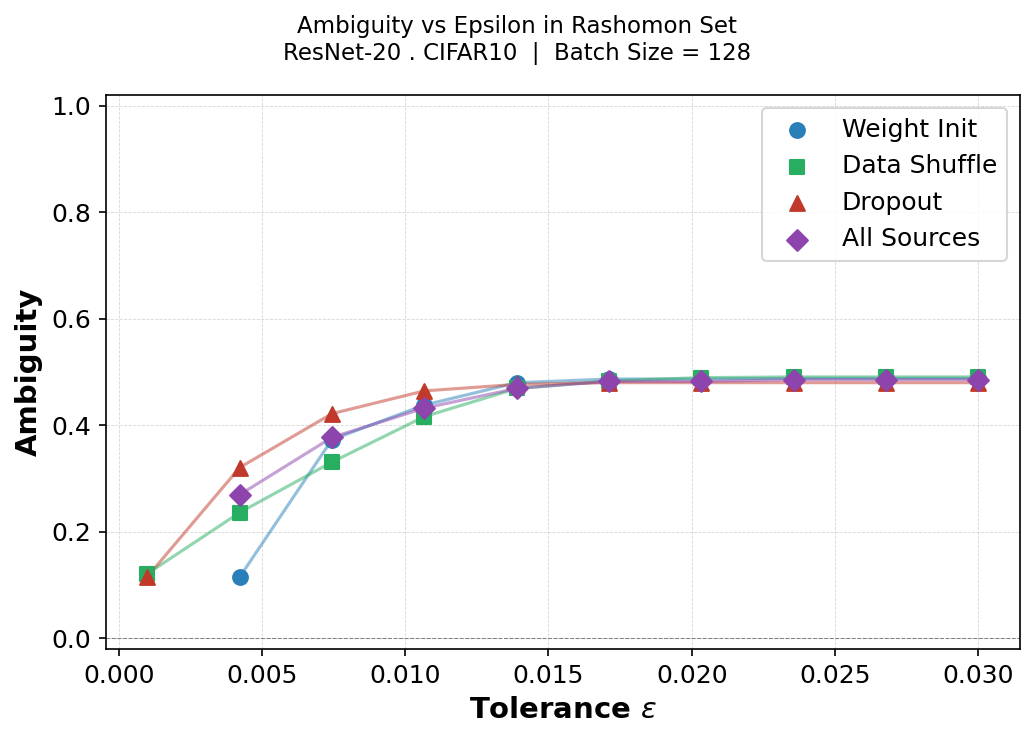}
&
\ambcell{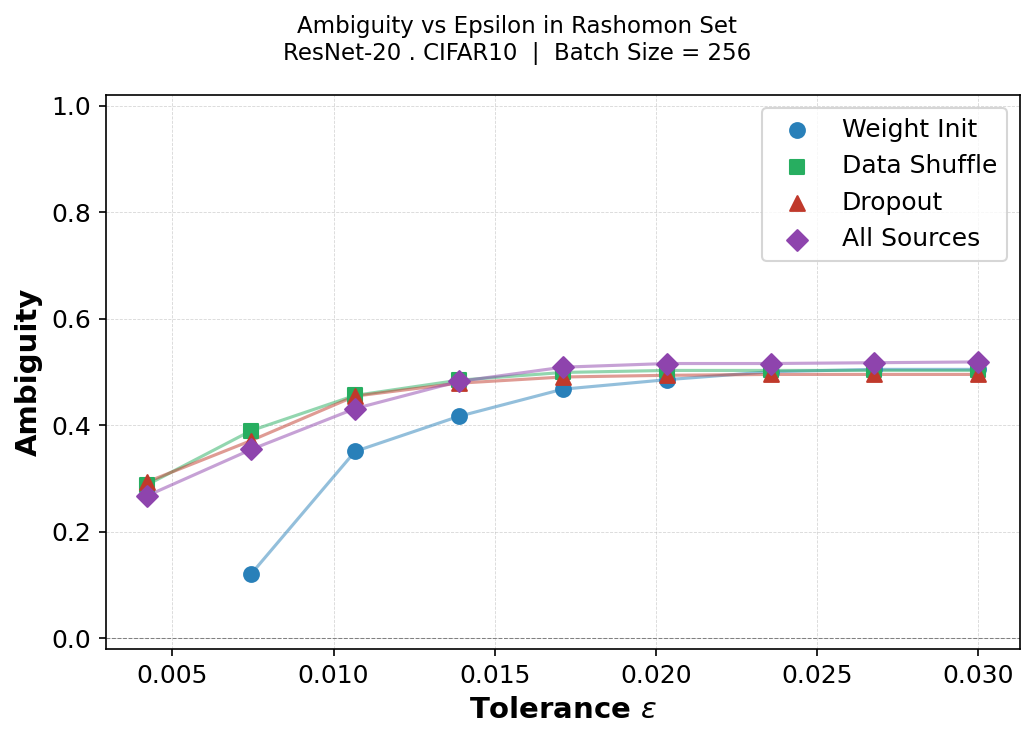}
\\
\textbf{\rotatebox{90}{Fashion-MNIST}}
&
\ambcell{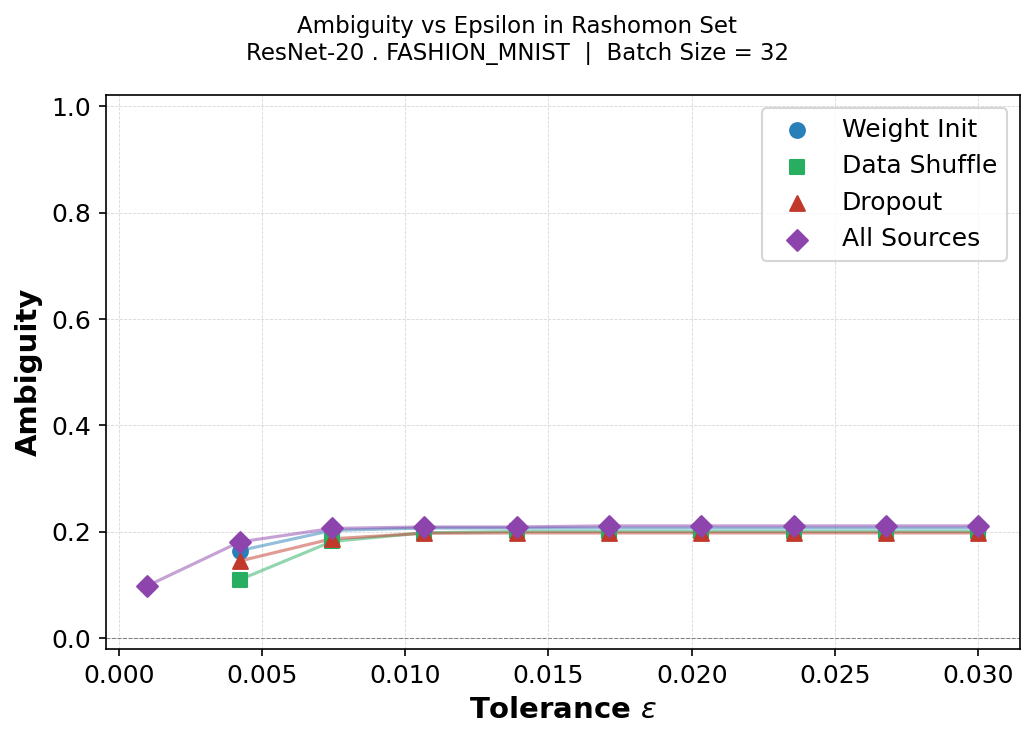}
&
\ambcell{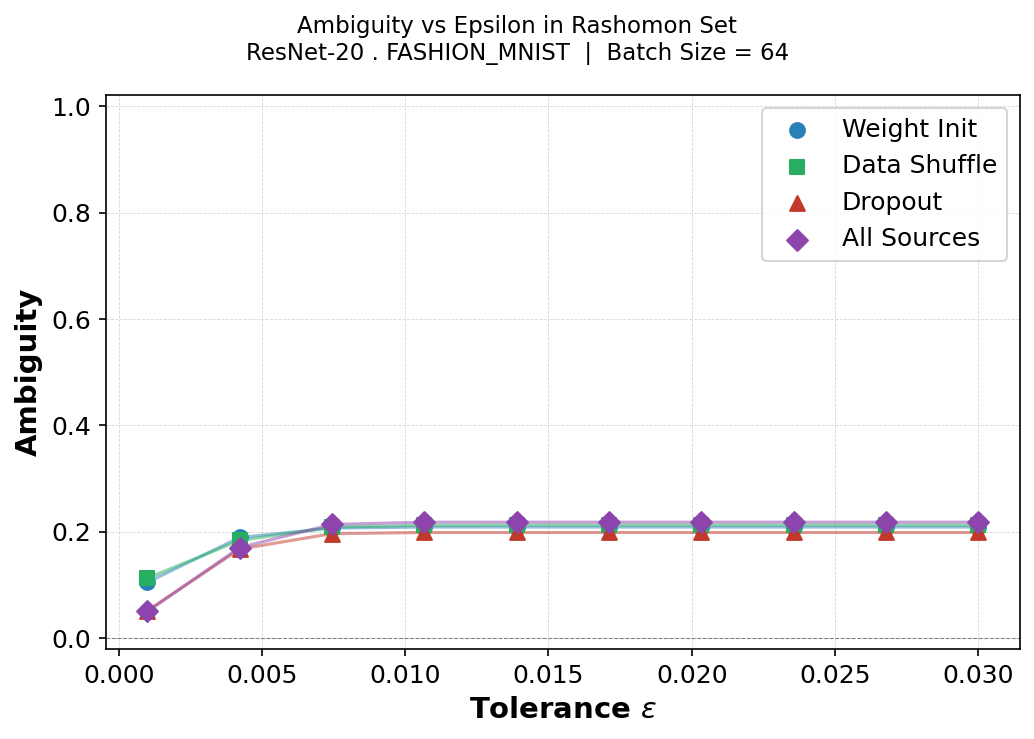}
&
\ambcell{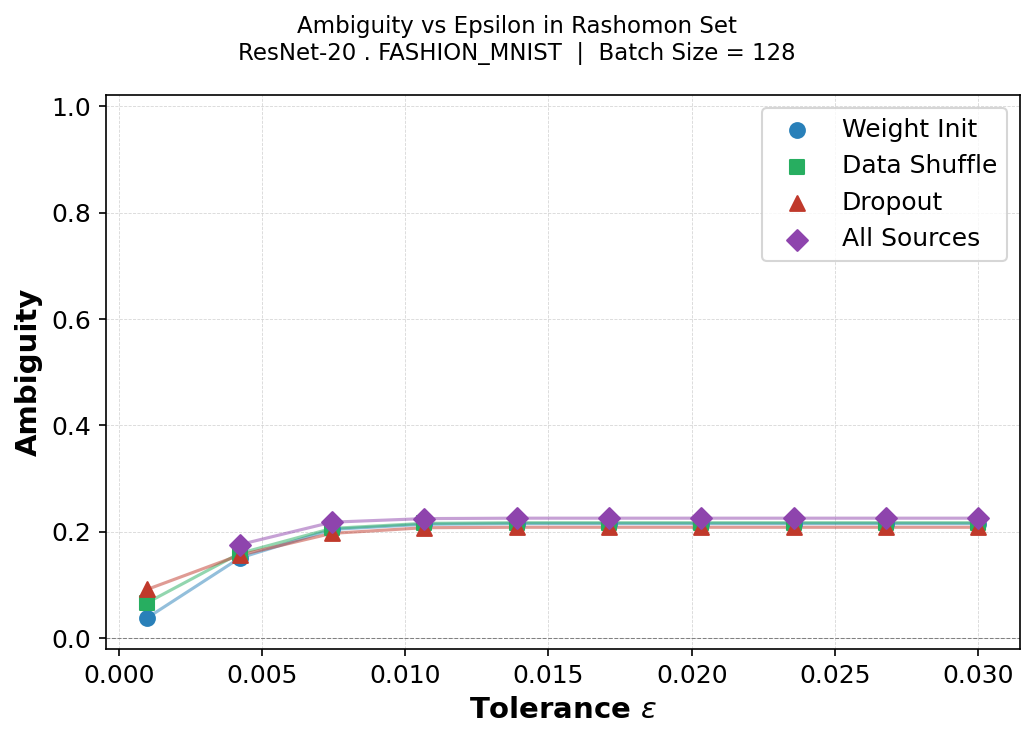}
&
\ambcell{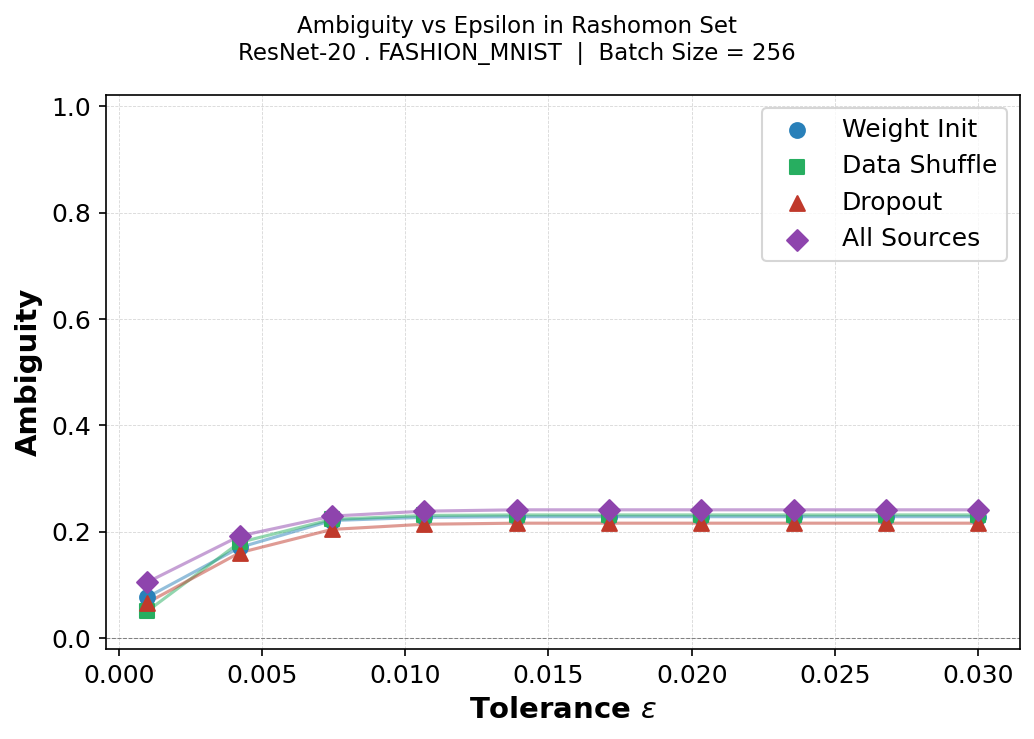}
\\
\end{tabular}
}
\caption{Ambiguity as a function of $\epsilon$ for image datasets (ResNet-20), broken down by stochasticity source. Rows correspond to datasets, columns to batch sizes.}
\label{fig:ambiguity-image}
\end{figure*}

\begin{figure*}[!t]
\centering
\setlength{\tabcolsep}{2pt}
\renewcommand{\arraystretch}{1.1}

\resizebox{\textwidth}{!}{%
\begin{tabular}{rcccc}
&
\textbf{BS=32} &
\textbf{BS=64} &
\textbf{BS=128} &
\textbf{BS=256}
\\

\rotatebox[origin=c]{90}{\textbf{ISOLET}}
&
\figcellxai{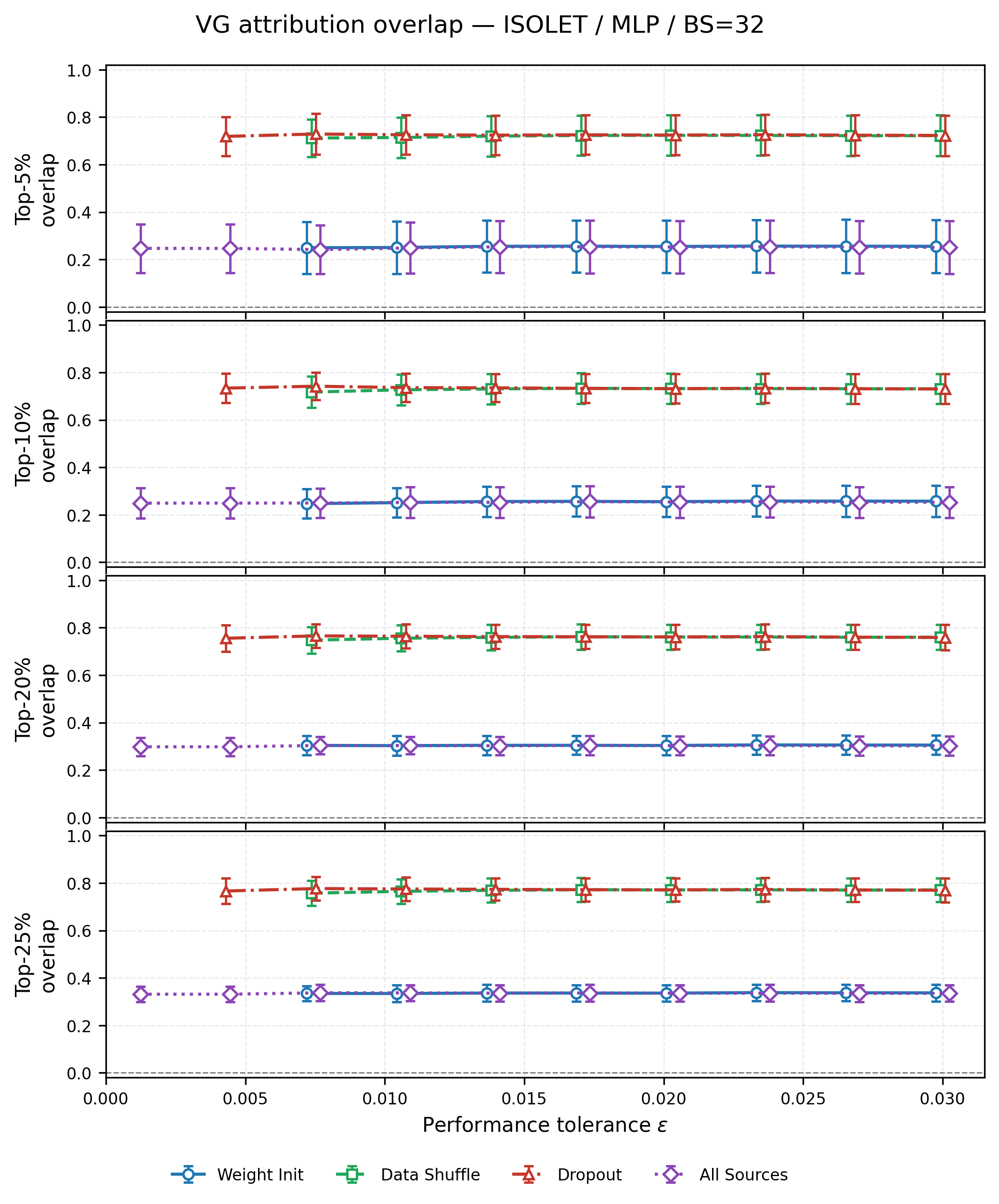}
&
\figcellxai{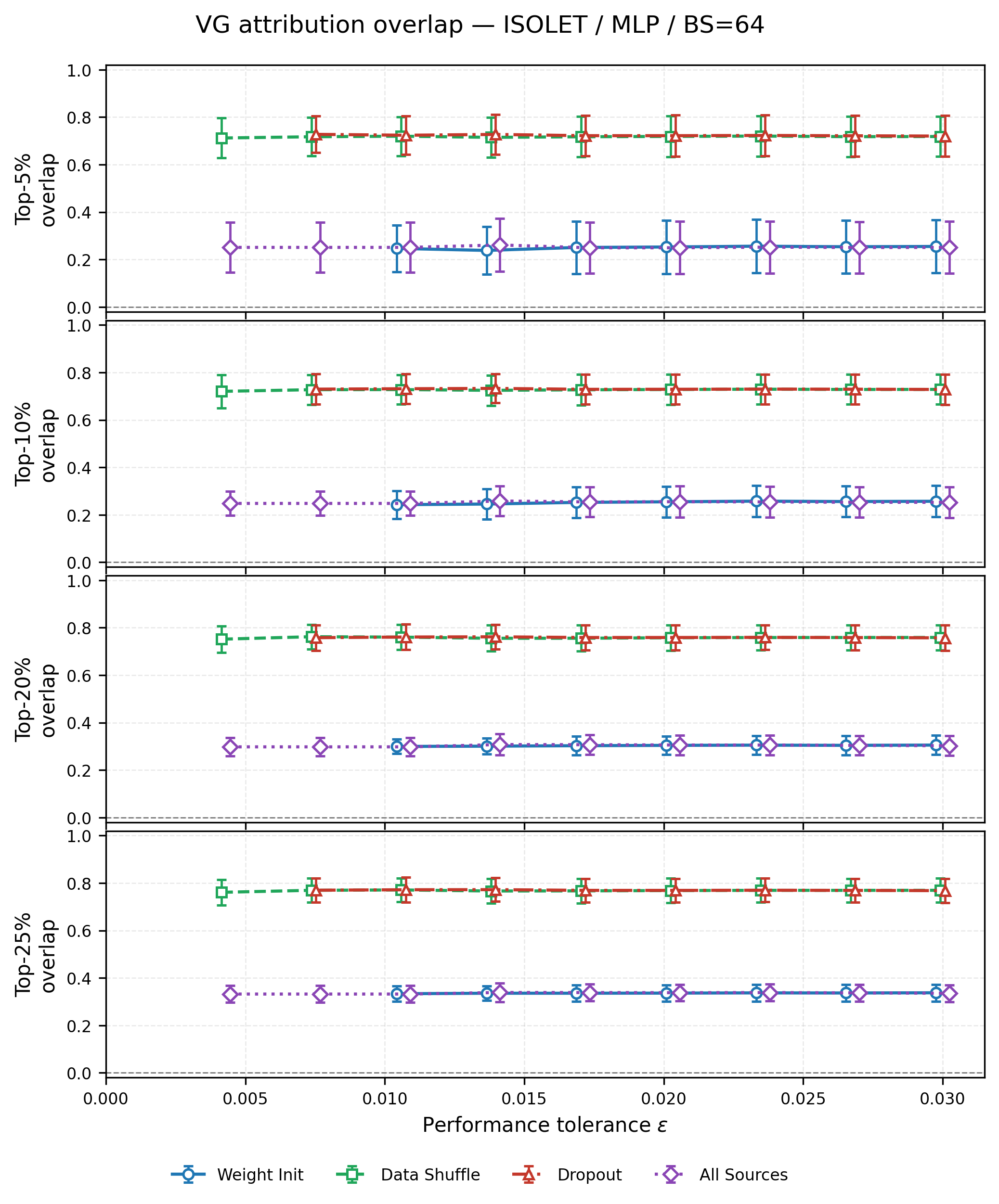}
&
\figcellxai{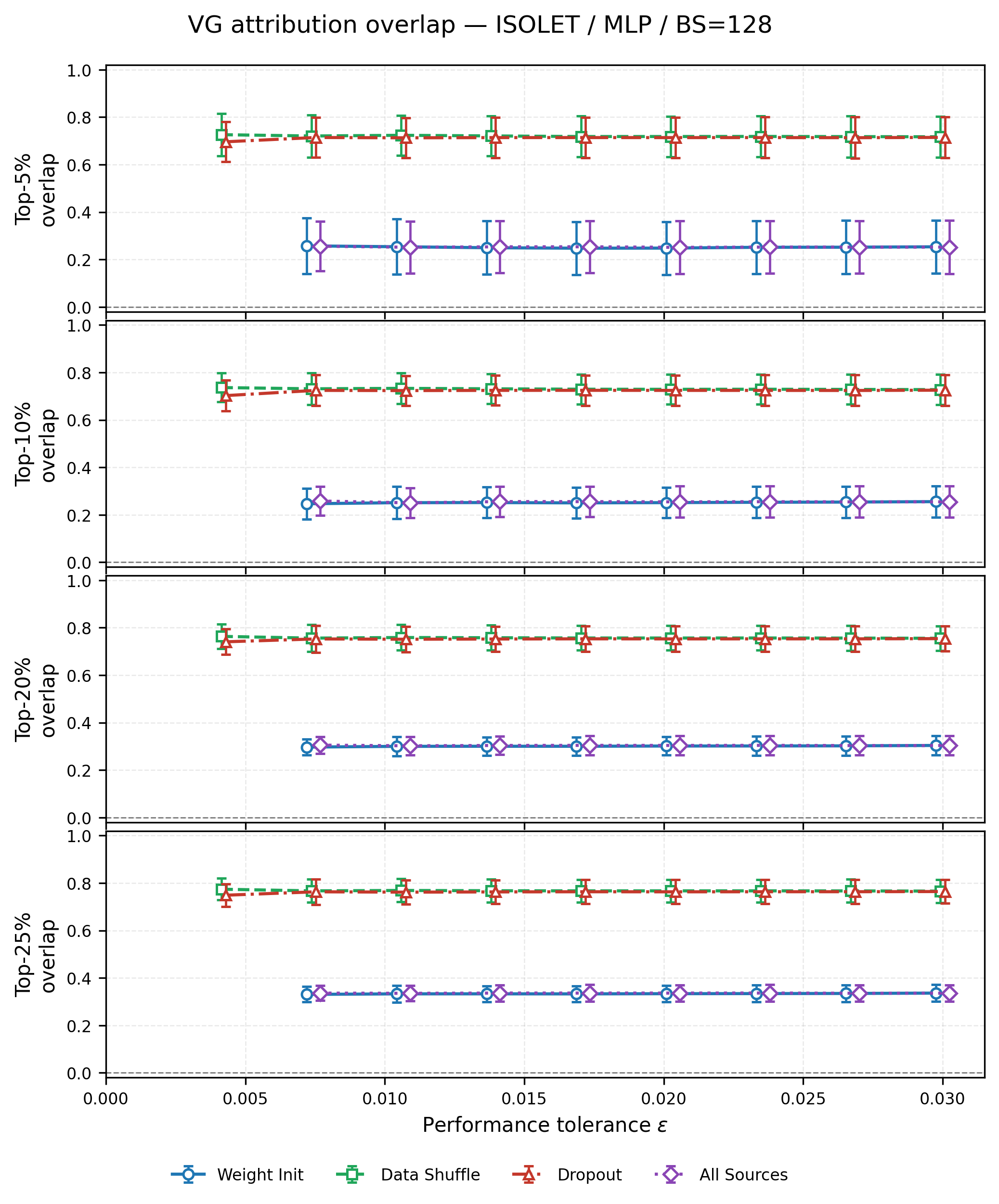}
&
\figcellxai{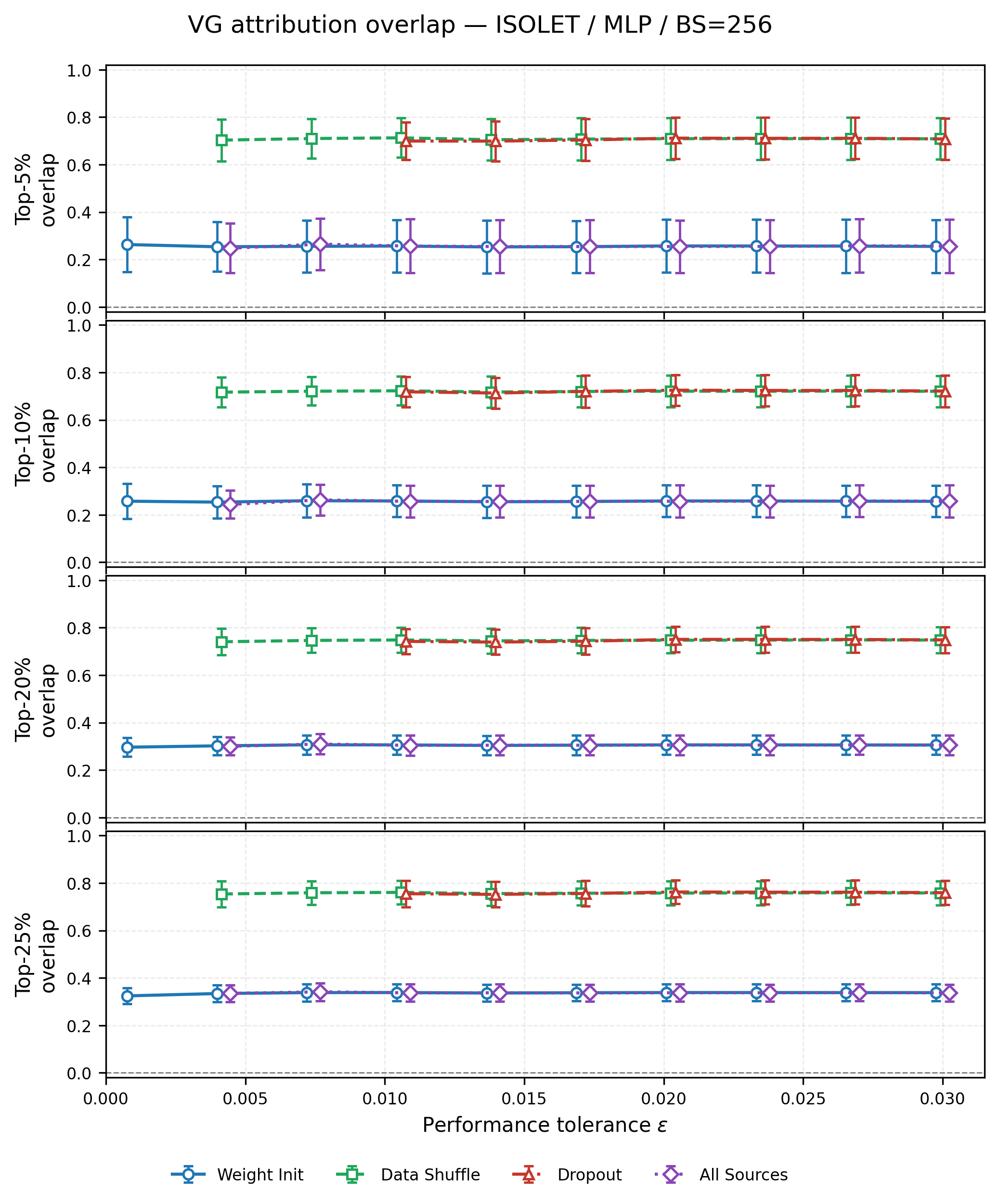}
\\

\rotatebox[origin=c]{90}{\textbf{Optical}}
&
\figcellxai{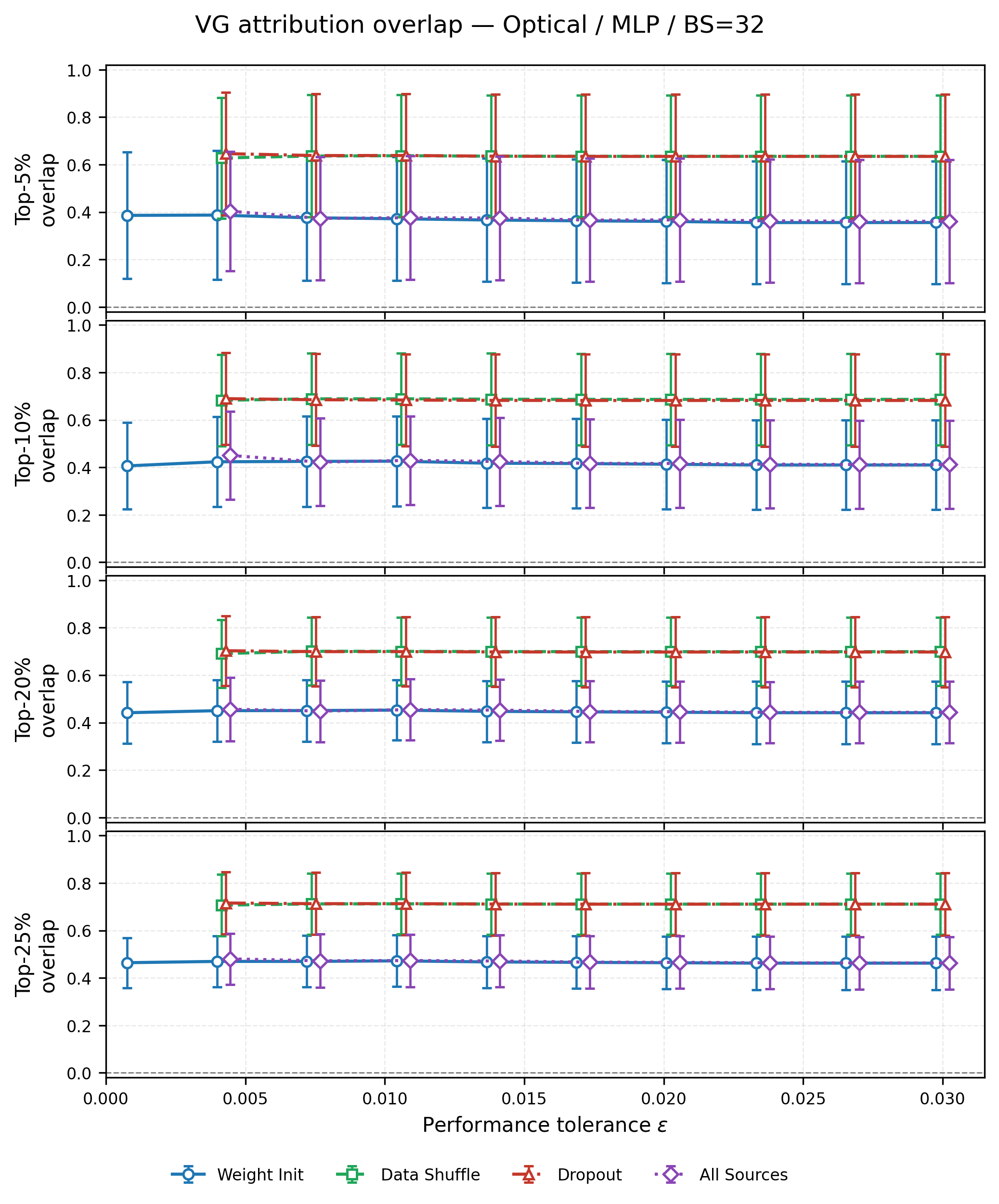}
&
\figcellxai{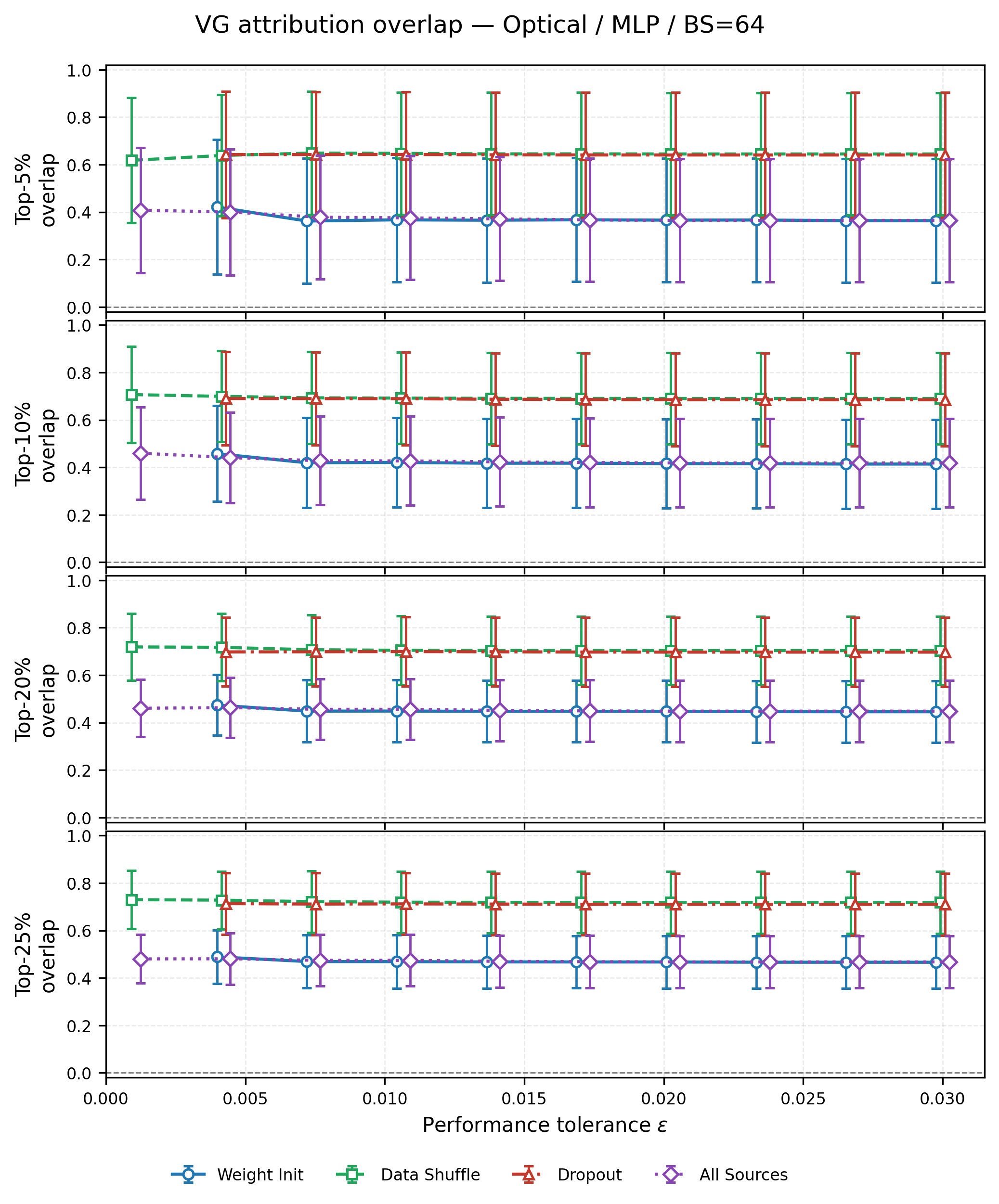}
&
\figcellxai{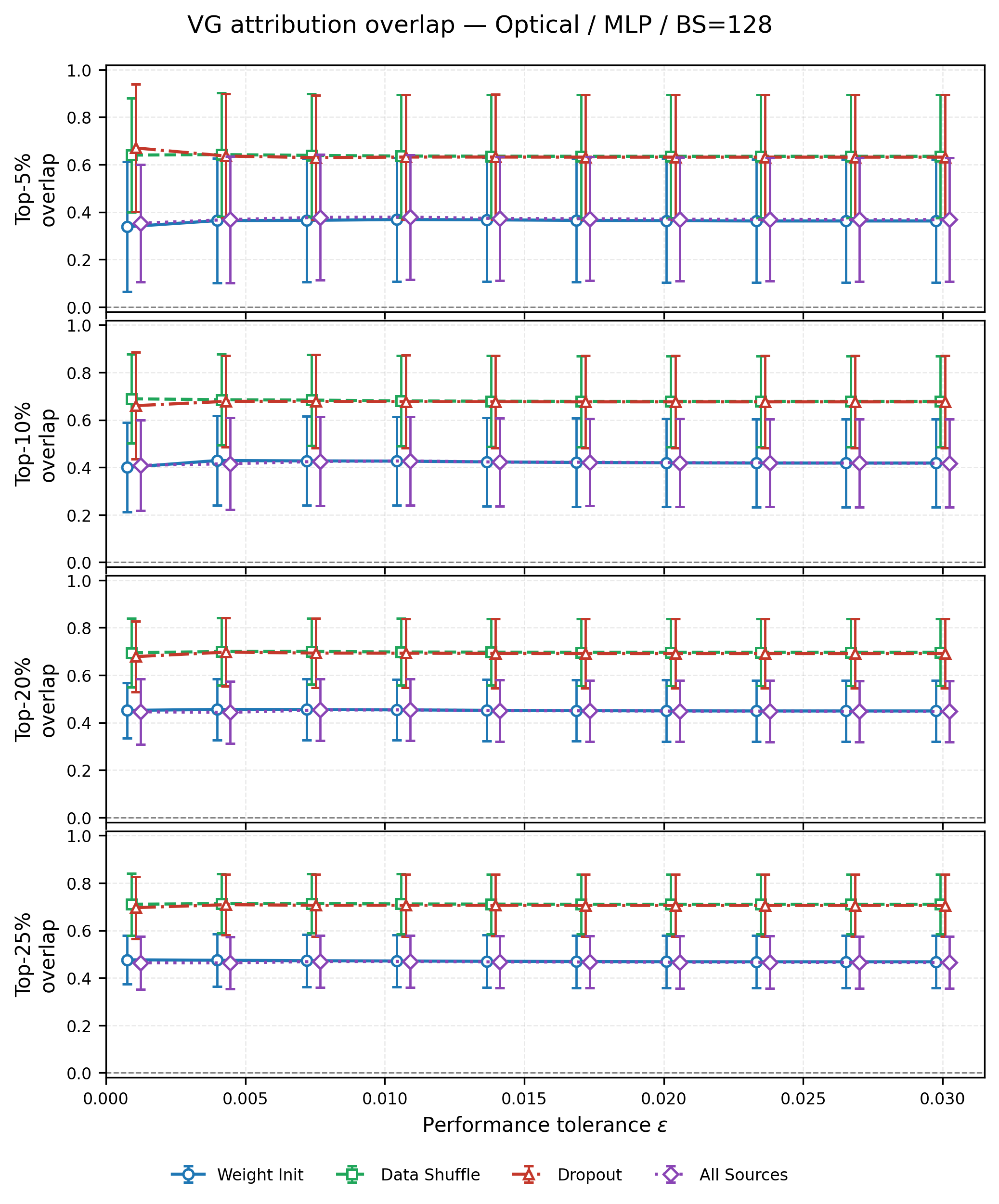}
&
\figcellxai{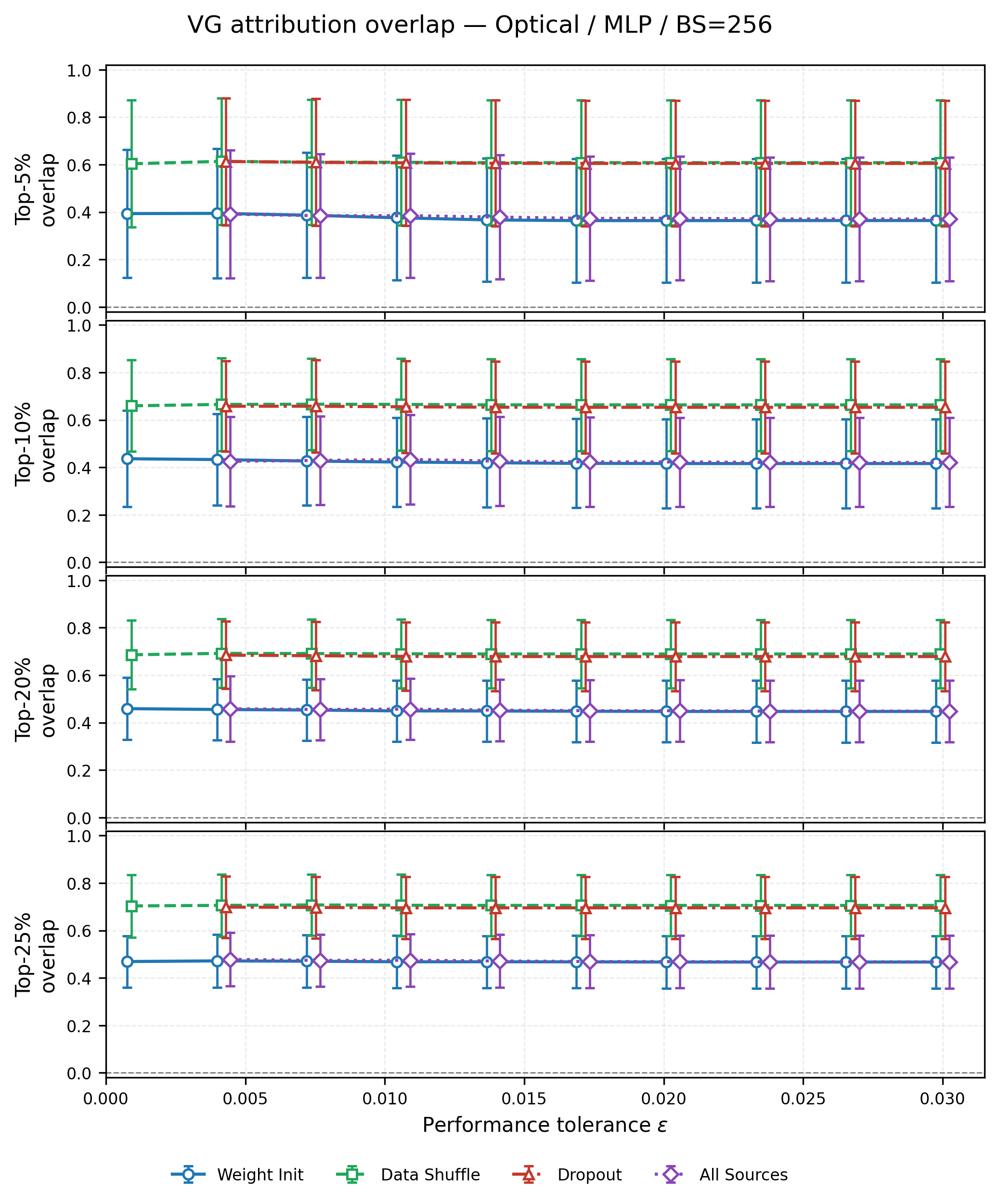}
\\

\rotatebox[origin=c]{90}{\textbf{Waveform}}
&
\figcellxai{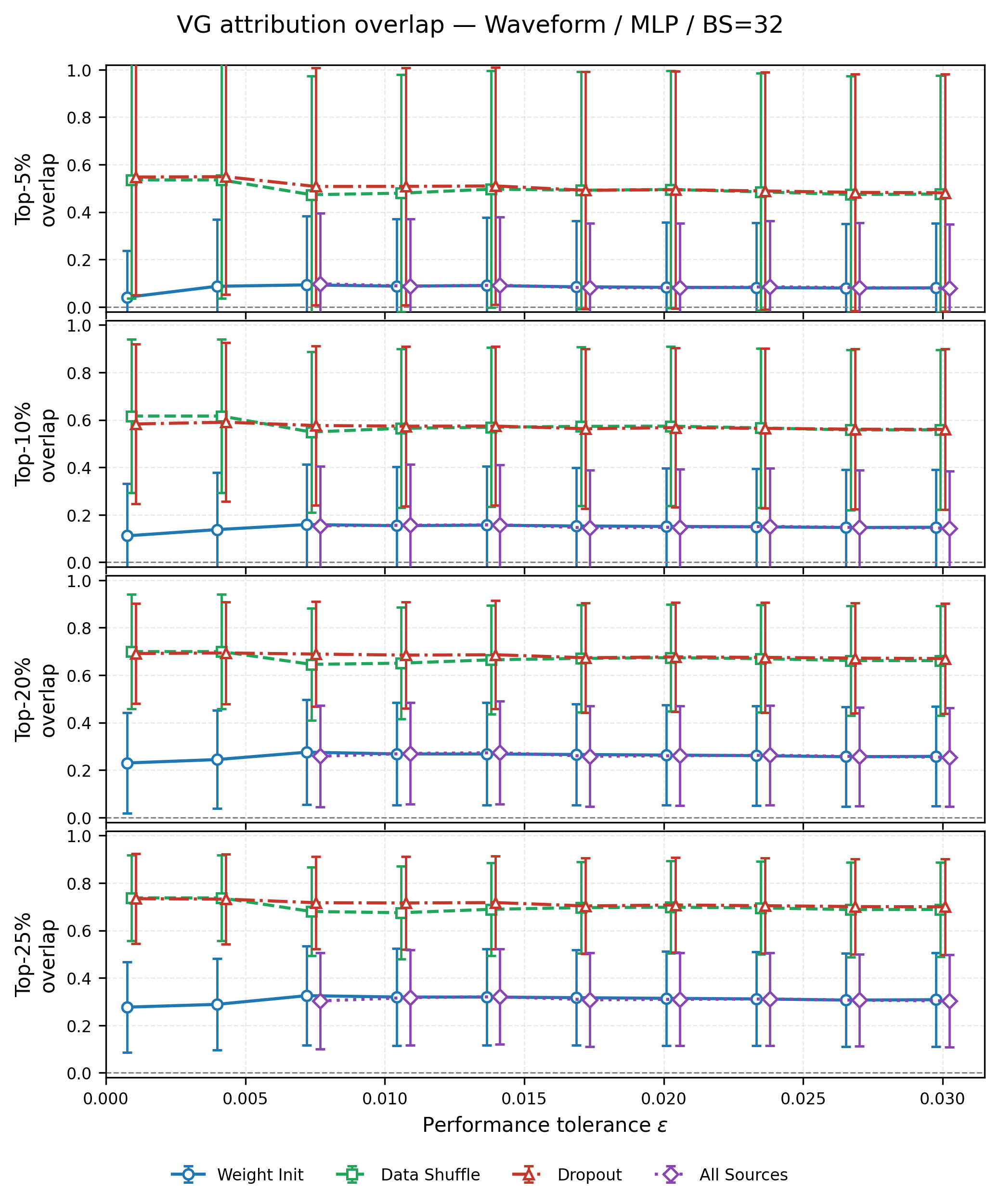}
&
\figcellxai{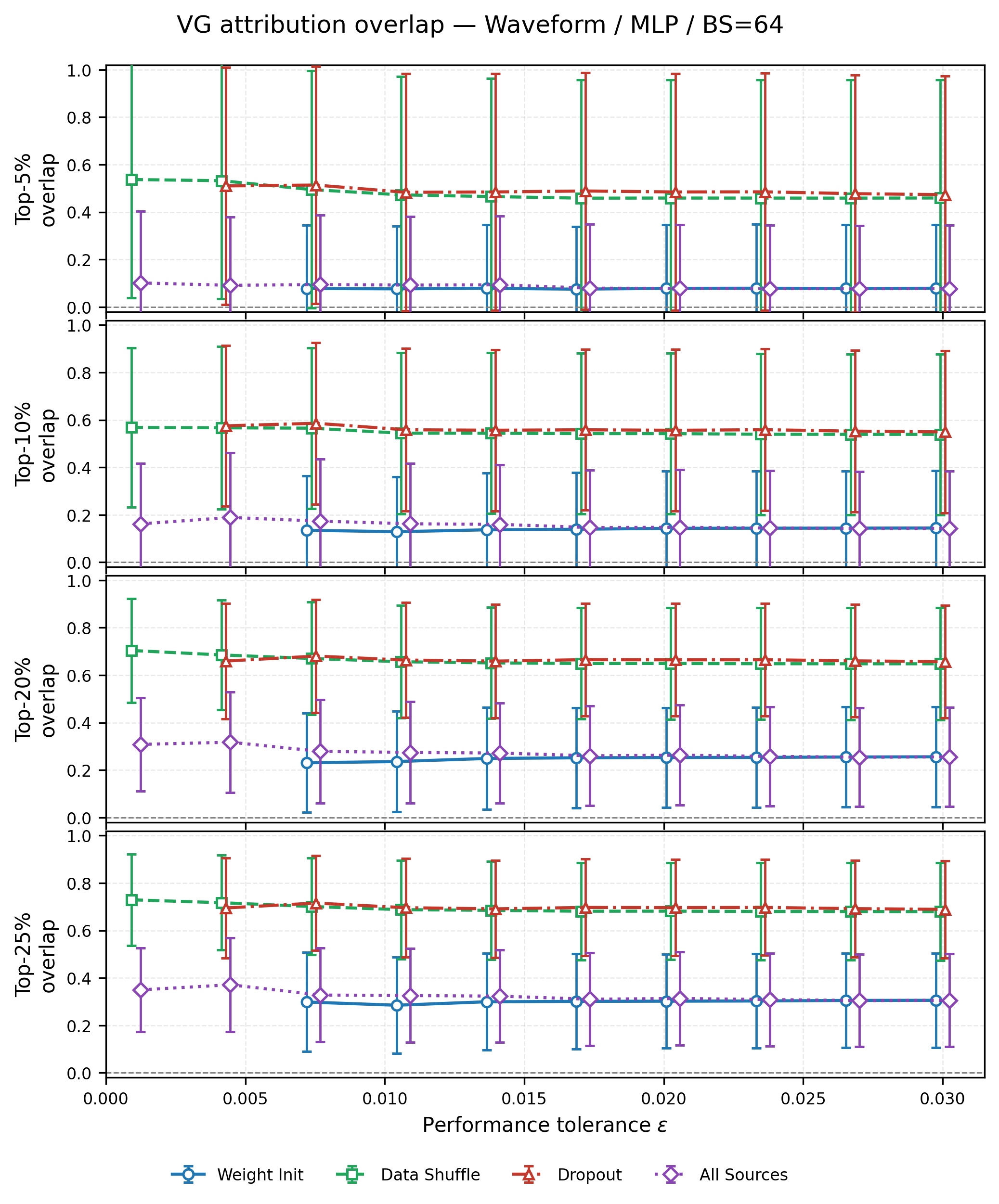}
&
\figcellxai{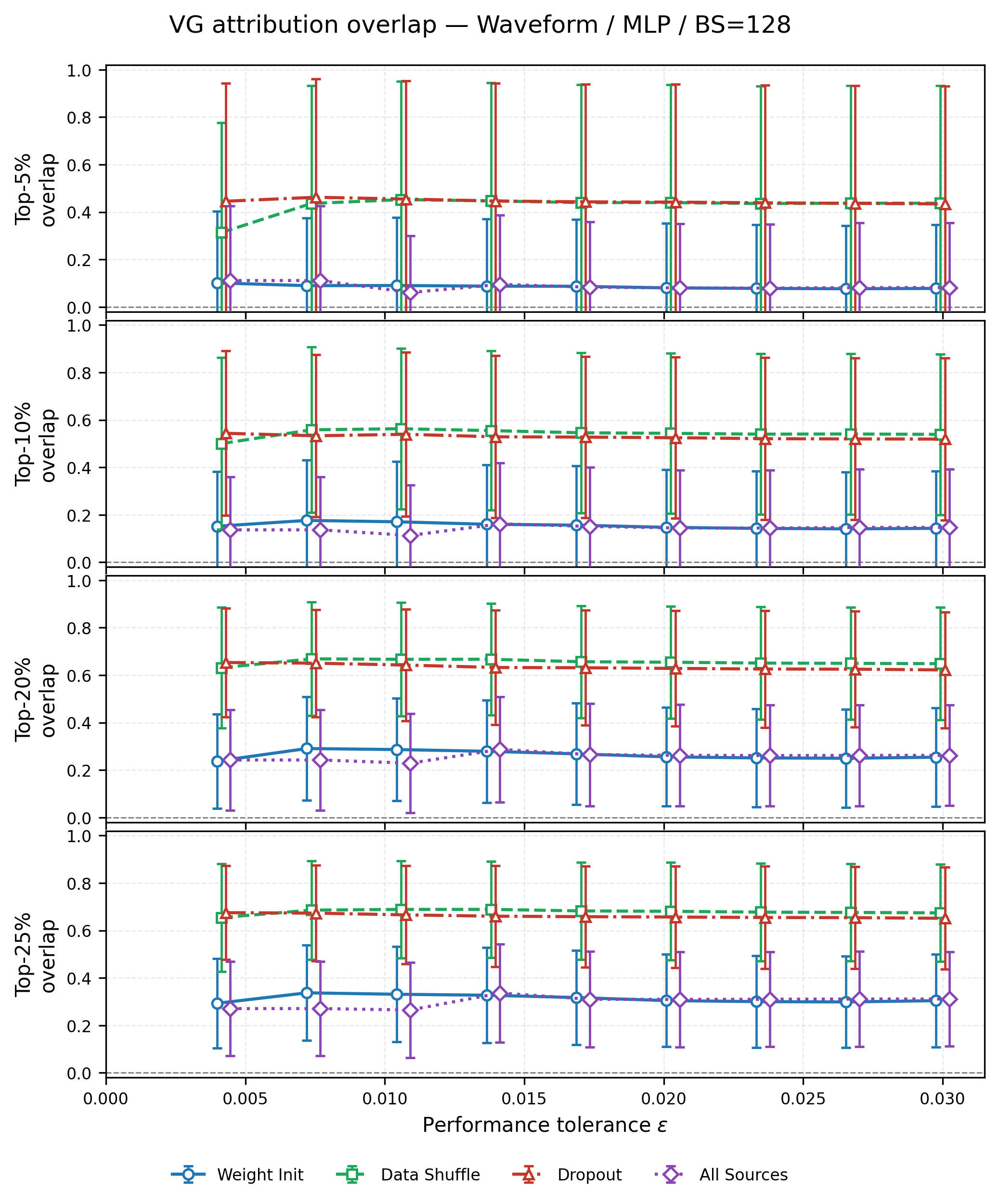}
&
\figcellxai{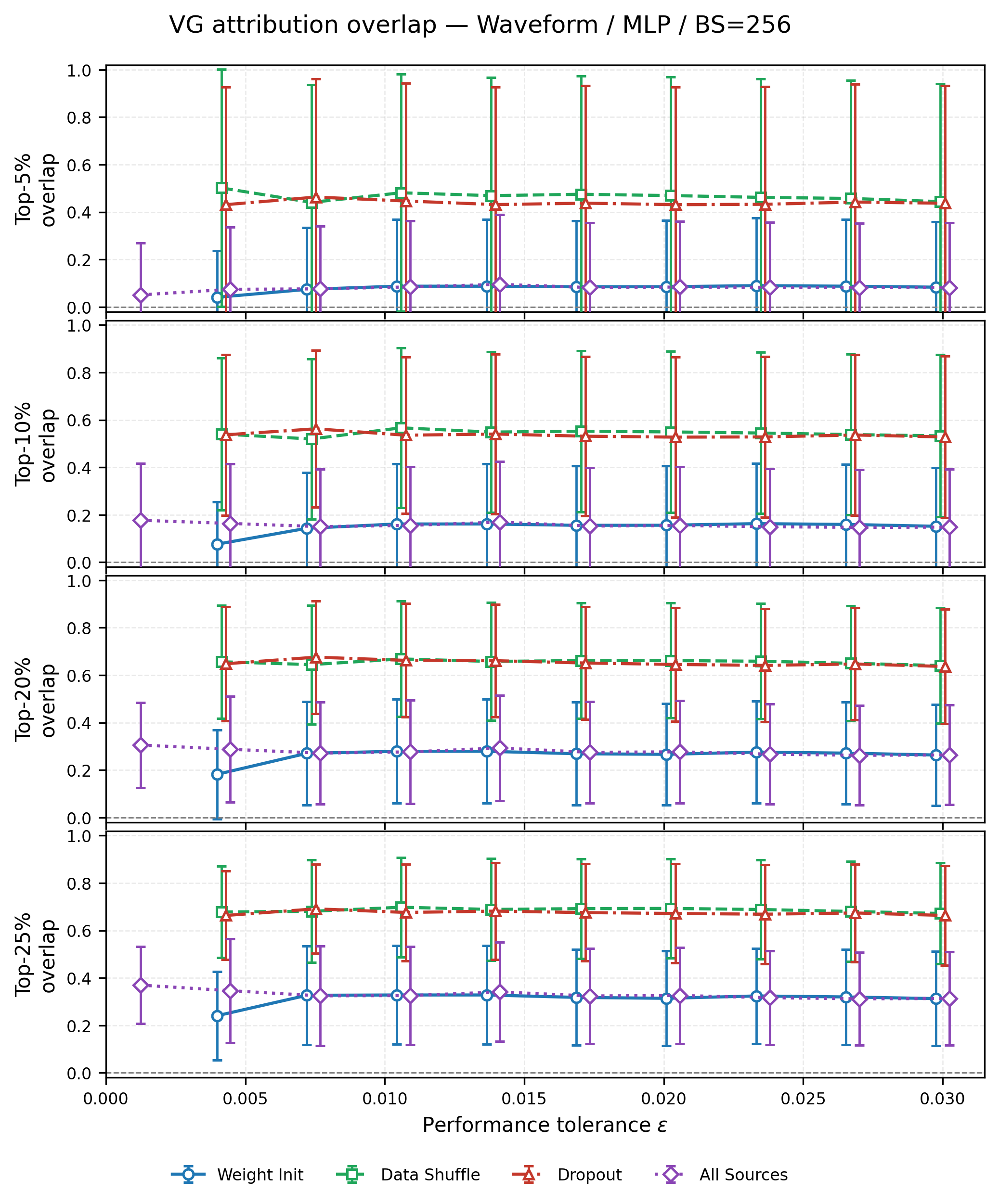}
\\
\end{tabular}
}

\caption{Top-$k$ attribution overlap obtained using Vanilla Gradients
(VG) as a function of the performance tolerance $\epsilon$ for the
tabular datasets and MLP architecture. Rows correspond to datasets and
columns to batch sizes. Each plot contains separate panels for Top-5\%,
Top-10\%, Top-20\%, and Top-25\% overlap. Lower overlap indicates
stronger attribution-based evidence of decision-basis multiplicity.
Error bars report the standard deviation across model-pair--instance
comparisons. Points are omitted when the corresponding empirical
Rashomon set contains fewer than two models, since pairwise attribution
agreement is then undefined. Small horizontal offsets are used only to
make overlapping markers and error bars visible.}

\label{fig:xai-topk-vg-tabular}
\end{figure*}

\begin{figure*}[!t]
\centering
\setlength{\tabcolsep}{2pt}
\renewcommand{\arraystretch}{1.1}

\resizebox{\textwidth}{!}{%
\begin{tabular}{rcccc}
&
\textbf{BS=32} &
\textbf{BS=64} &
\textbf{BS=128} &
\textbf{BS=256}
\\

\rotatebox[origin=c]{90}{\textbf{CIFAR-10}}
&
\figcellxai{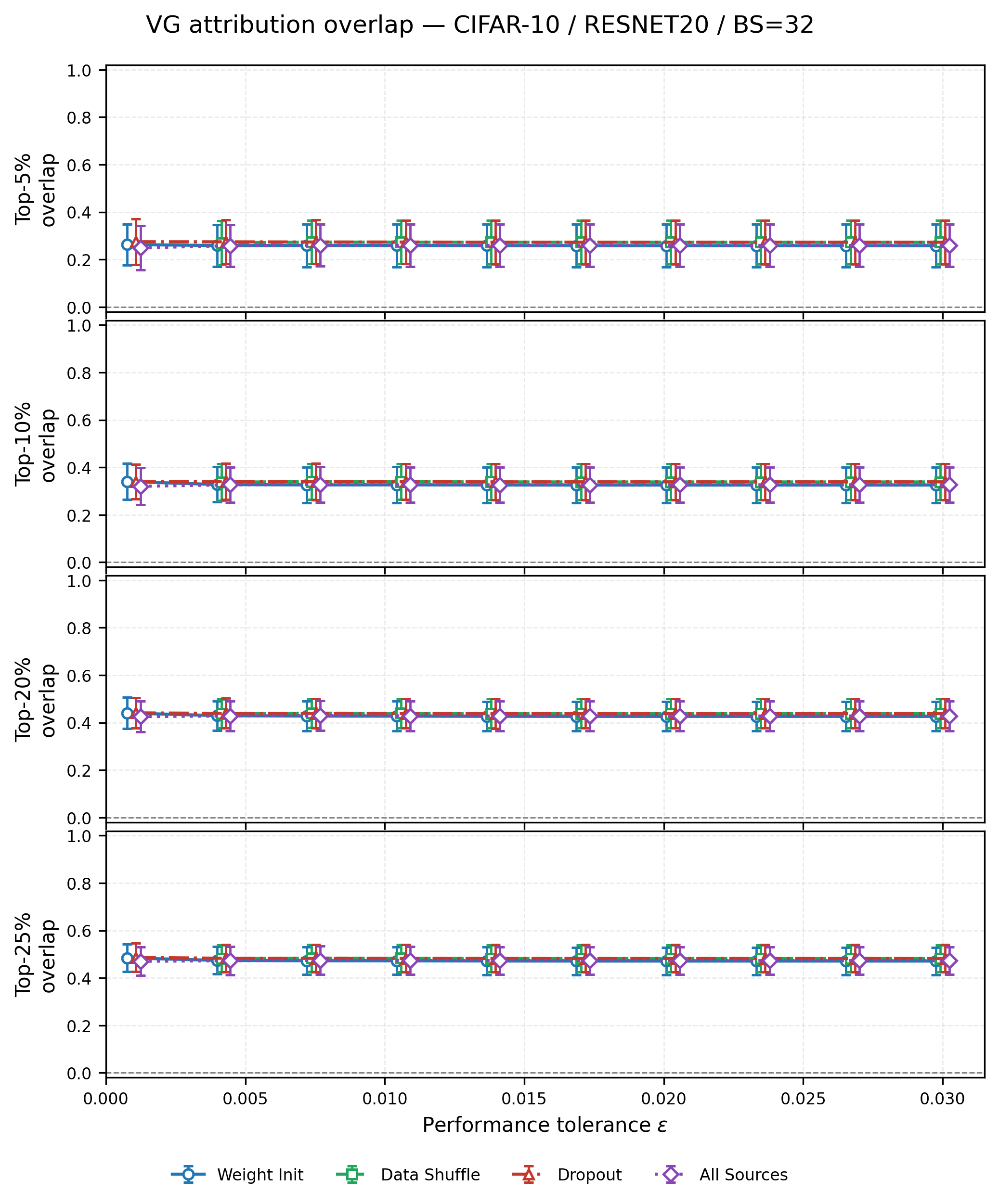}
&
\figcellxai{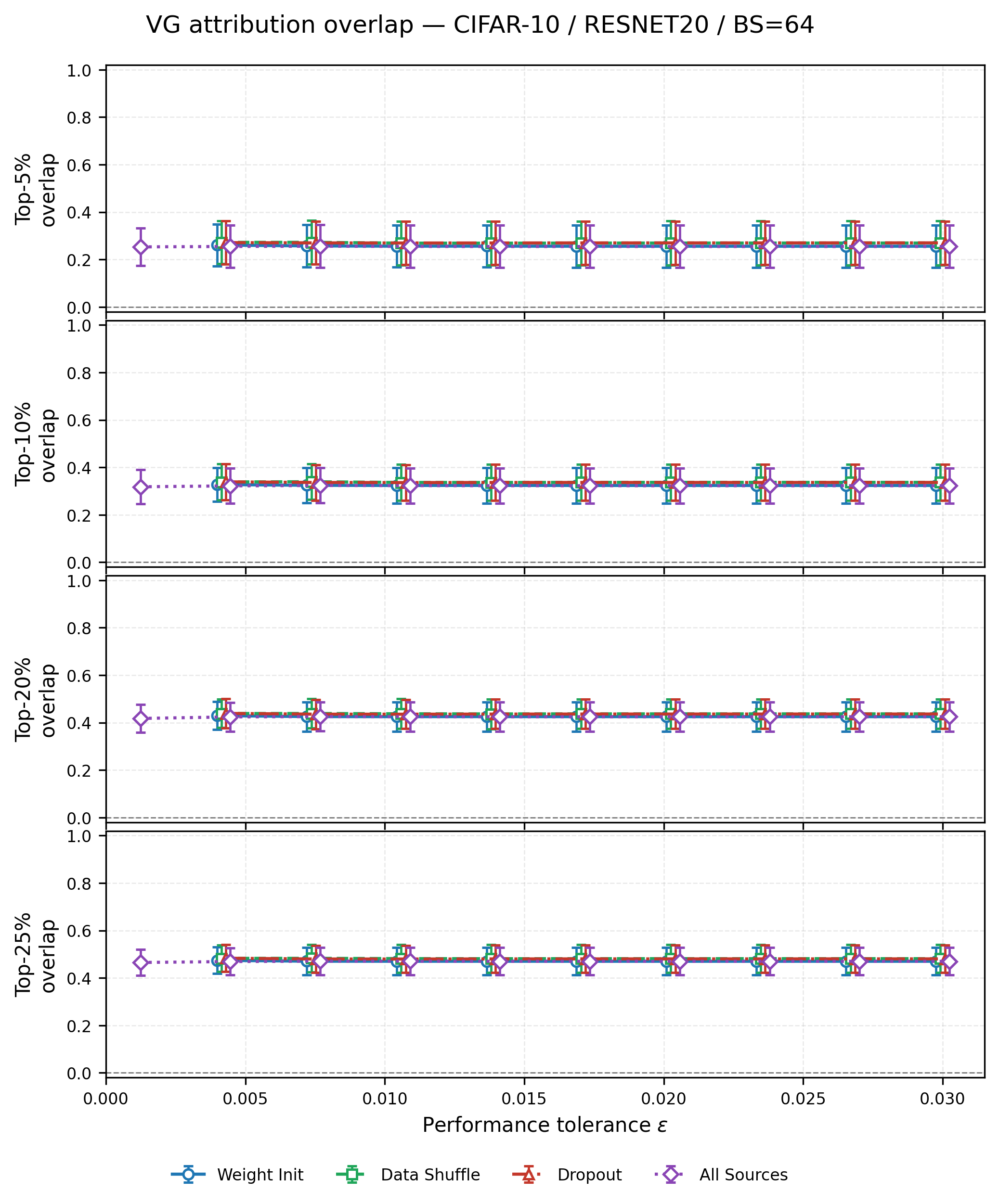}
&
\figcellxai{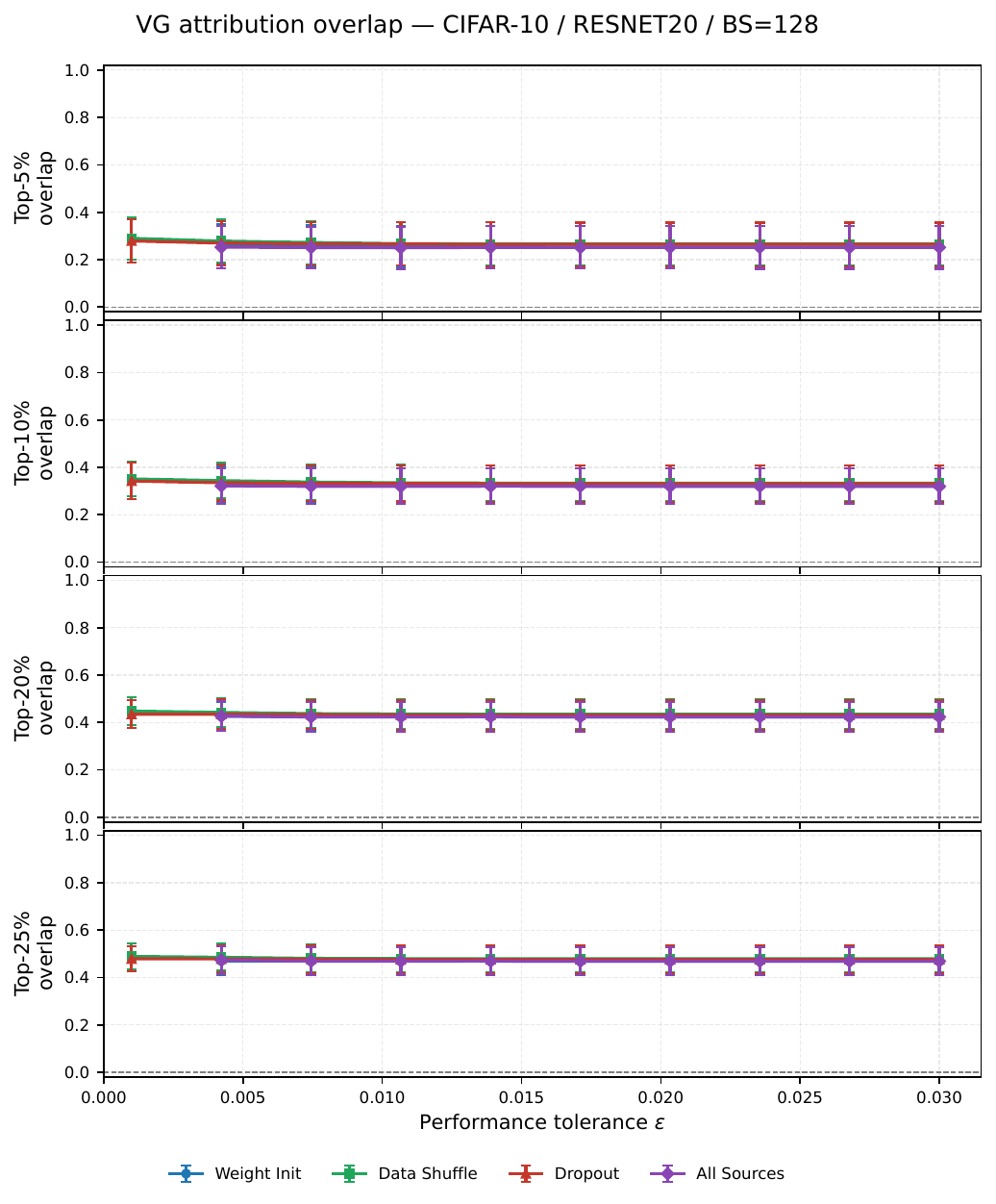}
&
\figcellxai{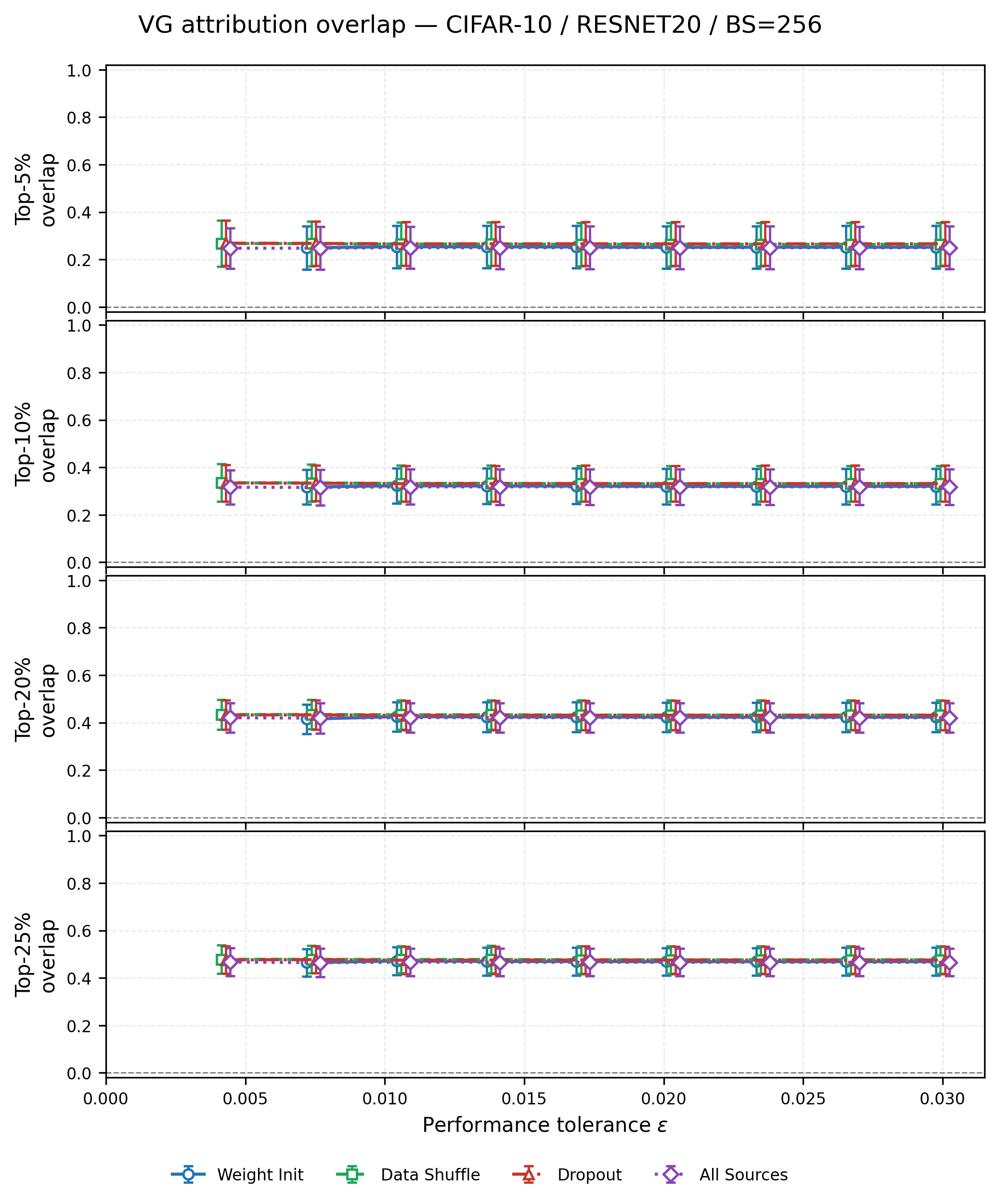}
\\

\rotatebox[origin=c]{90}{\textbf{Fashion-MNIST}}
&
\figcellxai{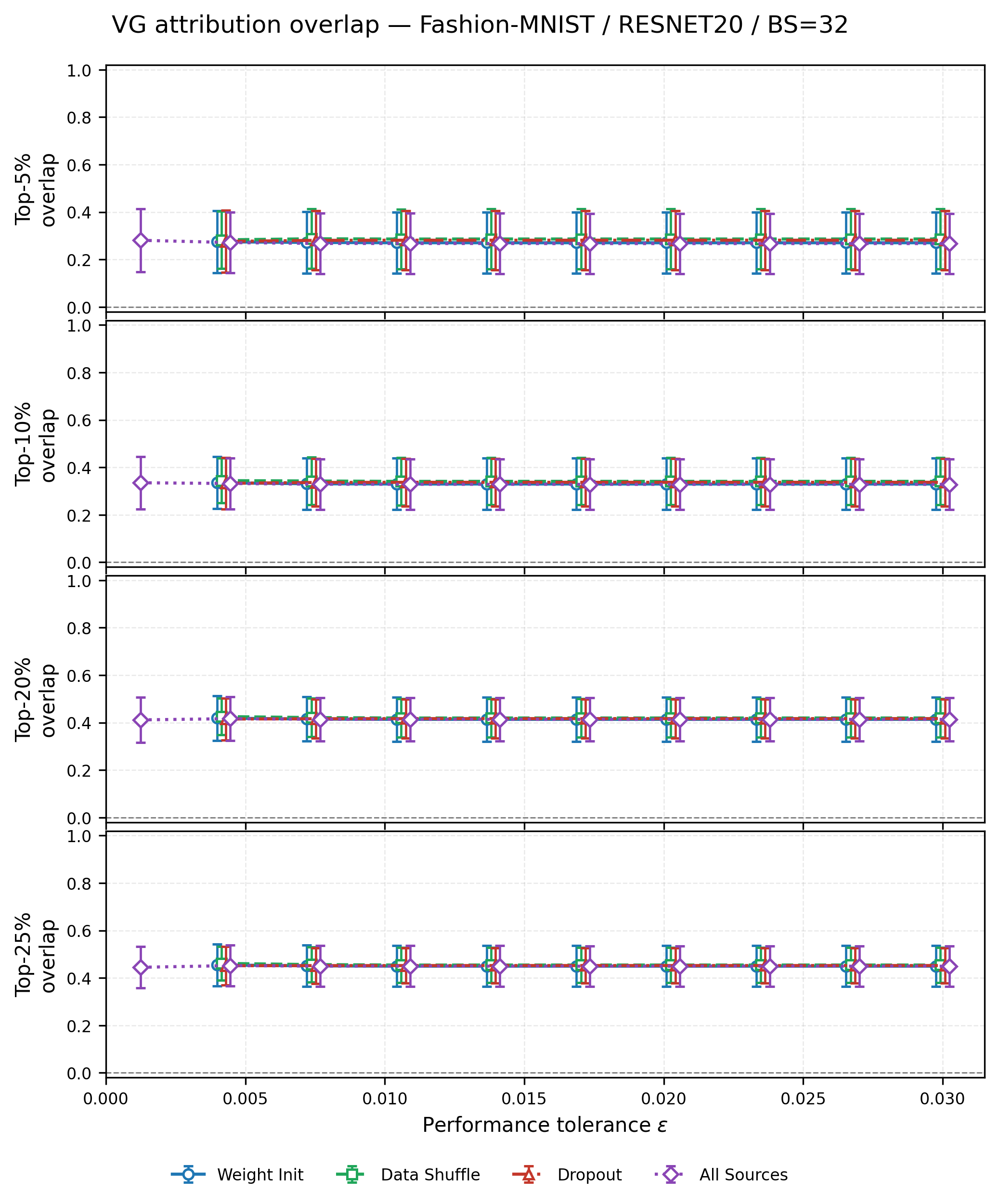}
&
\figcellxai{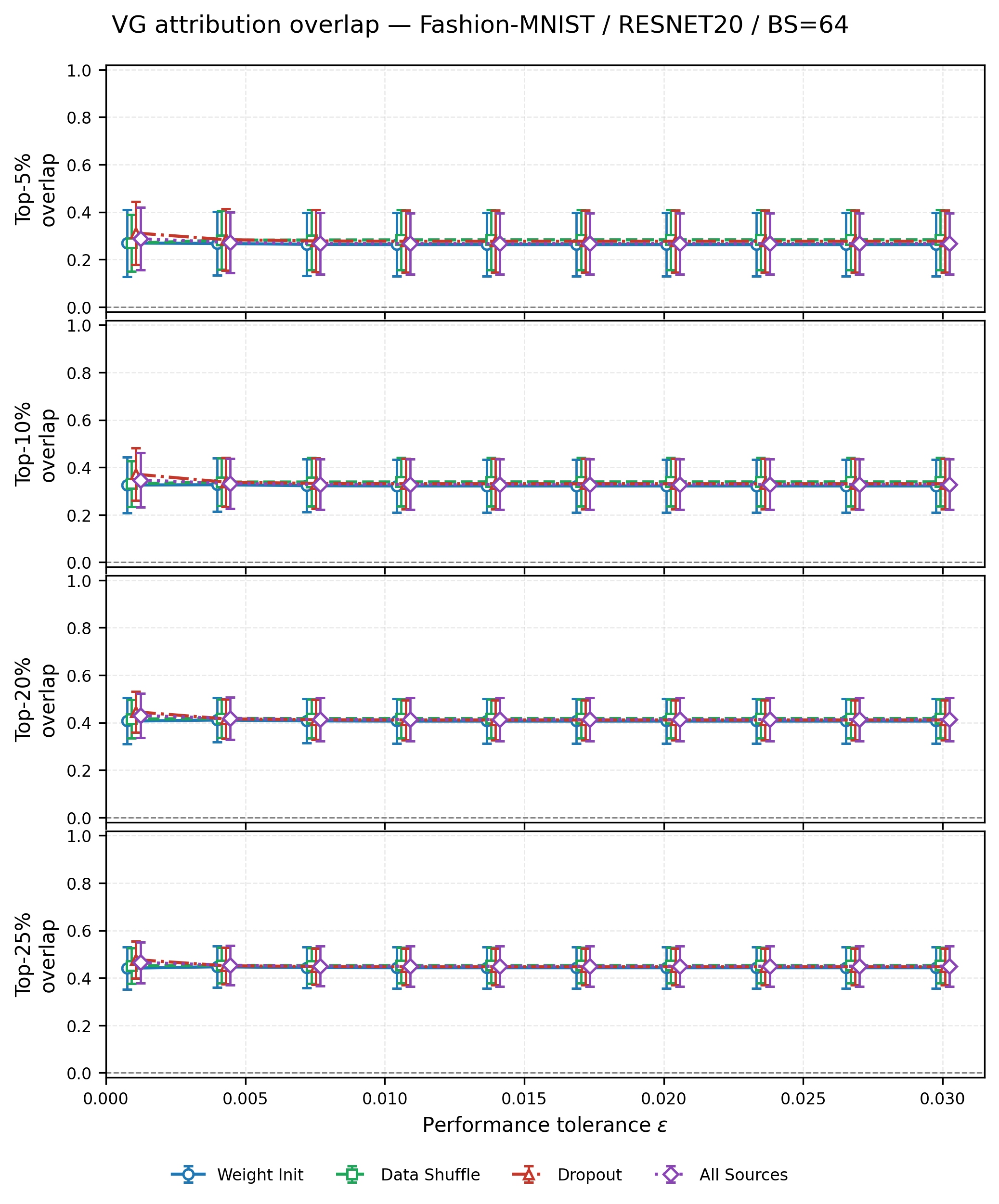}
&
\figcellxai{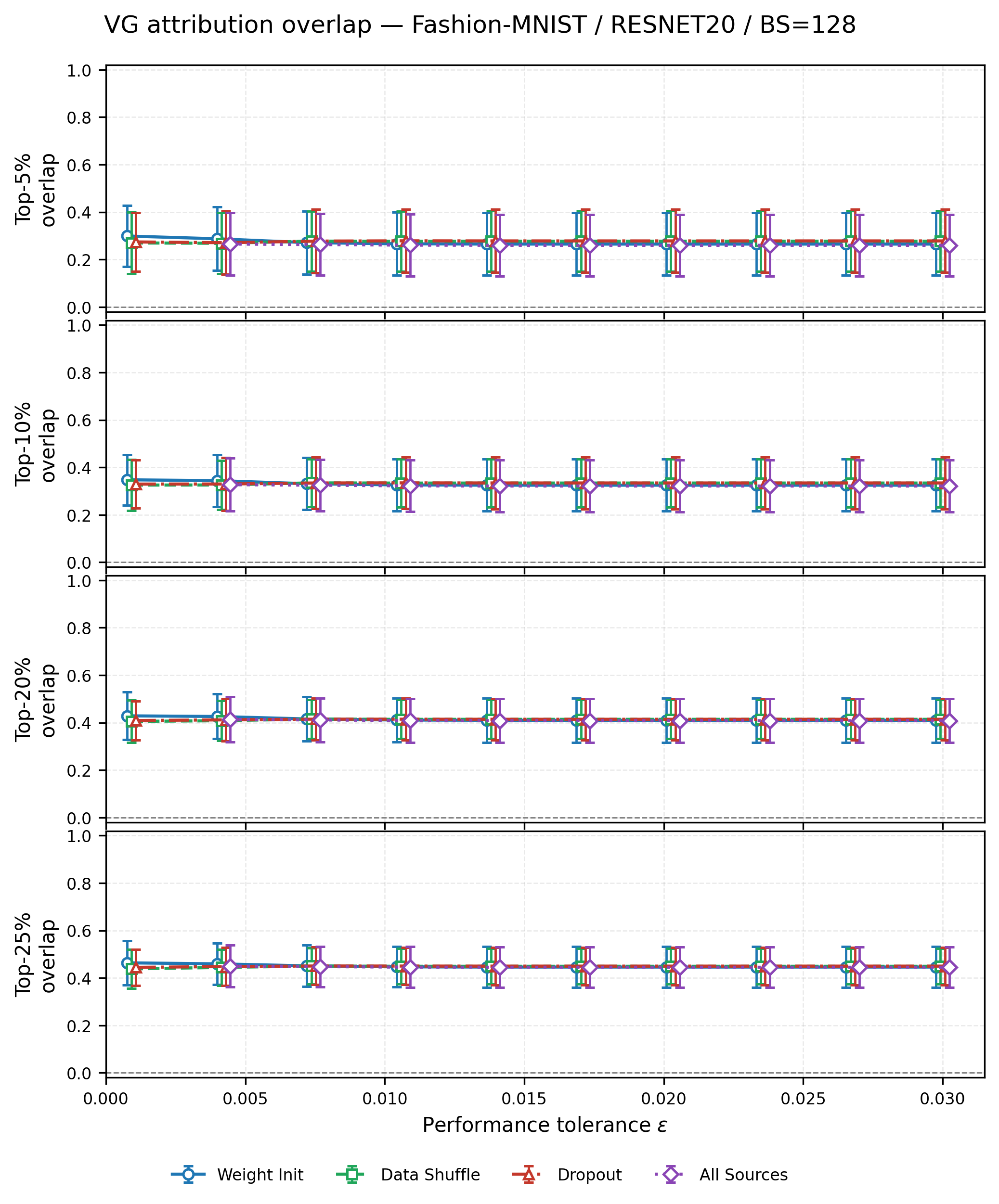}
&
\figcellxai{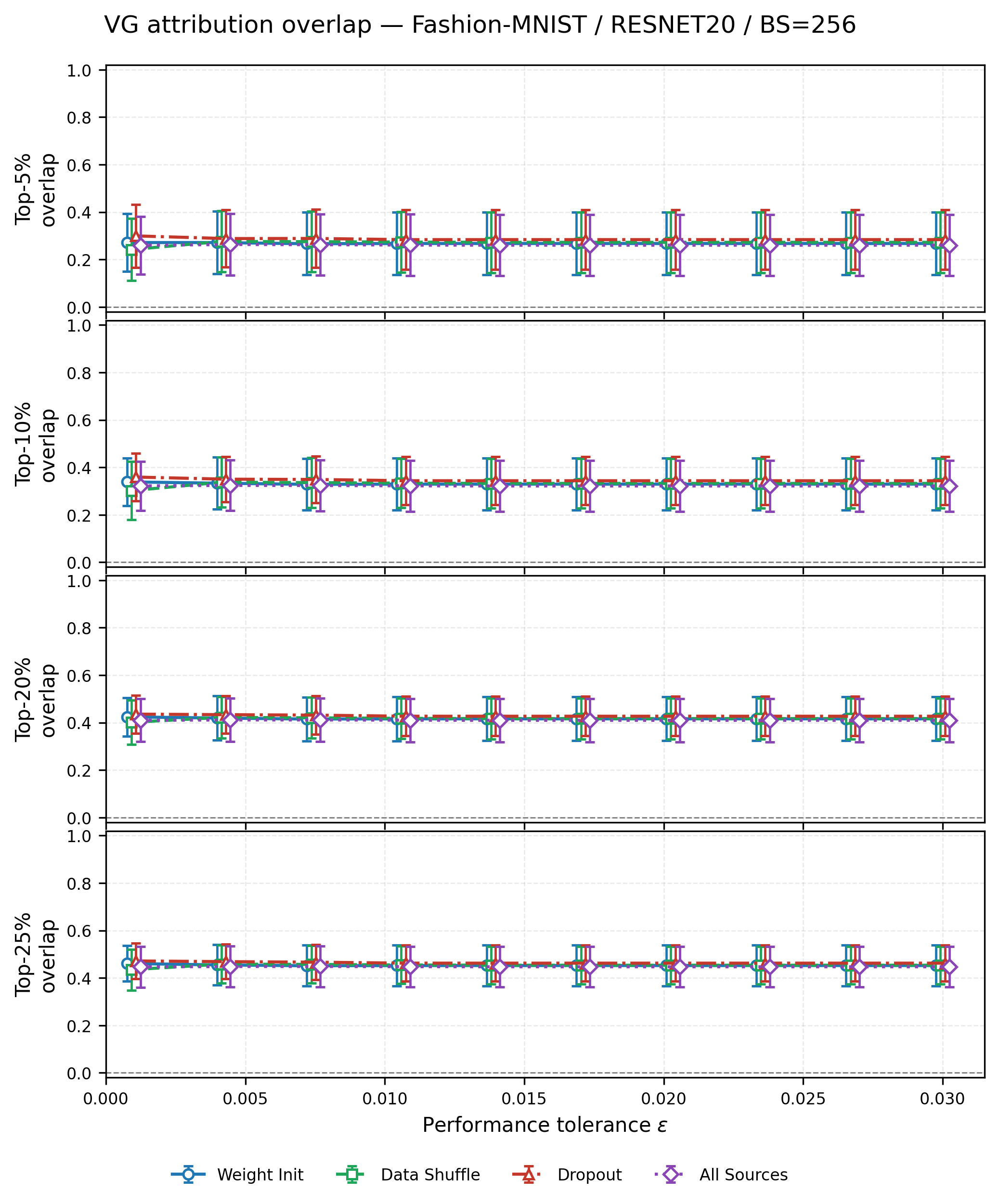}
\\
\end{tabular}
}

\caption{Top-$k$ attribution overlap obtained using Vanilla Gradients
(VG) as a function of the performance tolerance $\epsilon$ for the image datasets and ResNet-20 architecture. Rows correspond to datasets and columns to batch sizes. Each plot contains separate panels for
Top-5\%, Top-10\%, Top-20\%, and Top-25\% overlap. Lower overlap
indicates stronger attribution-based evidence of decision-basis
multiplicity. Error bars report the standard deviation across
model-pair--instance comparisons. Points are omitted when the
corresponding empirical Rashomon set contains fewer than two models, since pairwise attribution agreement is then undefined. Small horizontal offsets are used only to make overlapping markers and error bars visible.}

\label{fig:xai-topk-vg-image}
\end{figure*}

\FloatBarrier

Figs. \ref{fig:ambiguity-tabular} and \ref{fig:ambiguity-image} report ambiguity against the best accuracy in the Rashomon set for each dataset and configuration.

\begin{insertedtext}
\begin{figure*}[!t]
\centering

\noindent
\begin{minipage}[c]{0.495\textwidth}
    \begin{minipage}[c]{0.07\linewidth}
        \centering
        \rotatebox{90}{\textbf{BS=32}}
    \end{minipage}%
    \begin{minipage}[c]{0.92\linewidth}
        \centering
        \includegraphics[width=\linewidth]
        {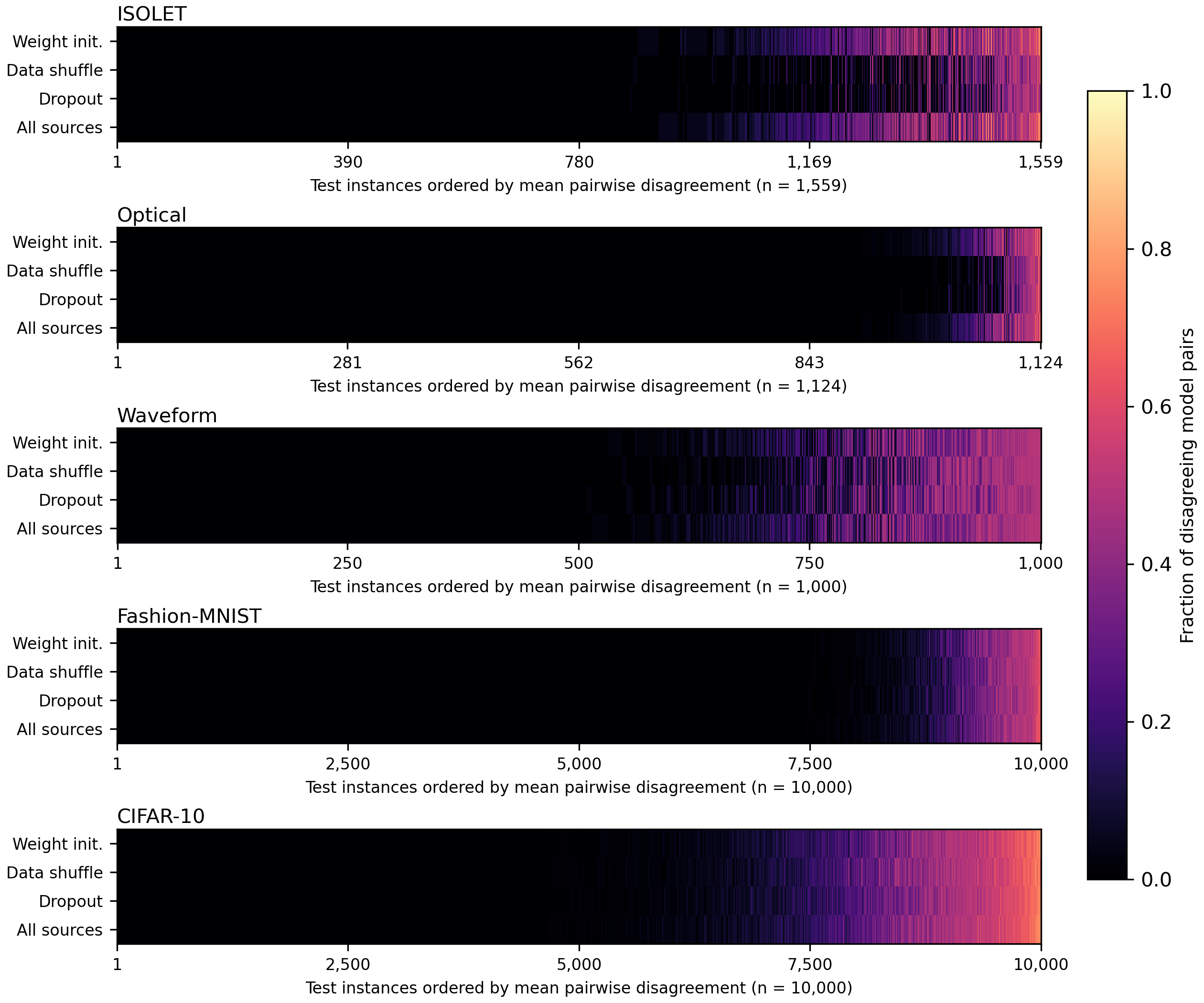}
    \end{minipage}
\end{minipage}%
\hfill
\begin{minipage}[c]{0.495\textwidth}
    \begin{minipage}[c]{0.07\linewidth}
        \centering
        \rotatebox{90}{\textbf{BS=64}}
    \end{minipage}%
    \begin{minipage}[c]{0.92\linewidth}
        \centering
        \includegraphics[width=\linewidth]
        {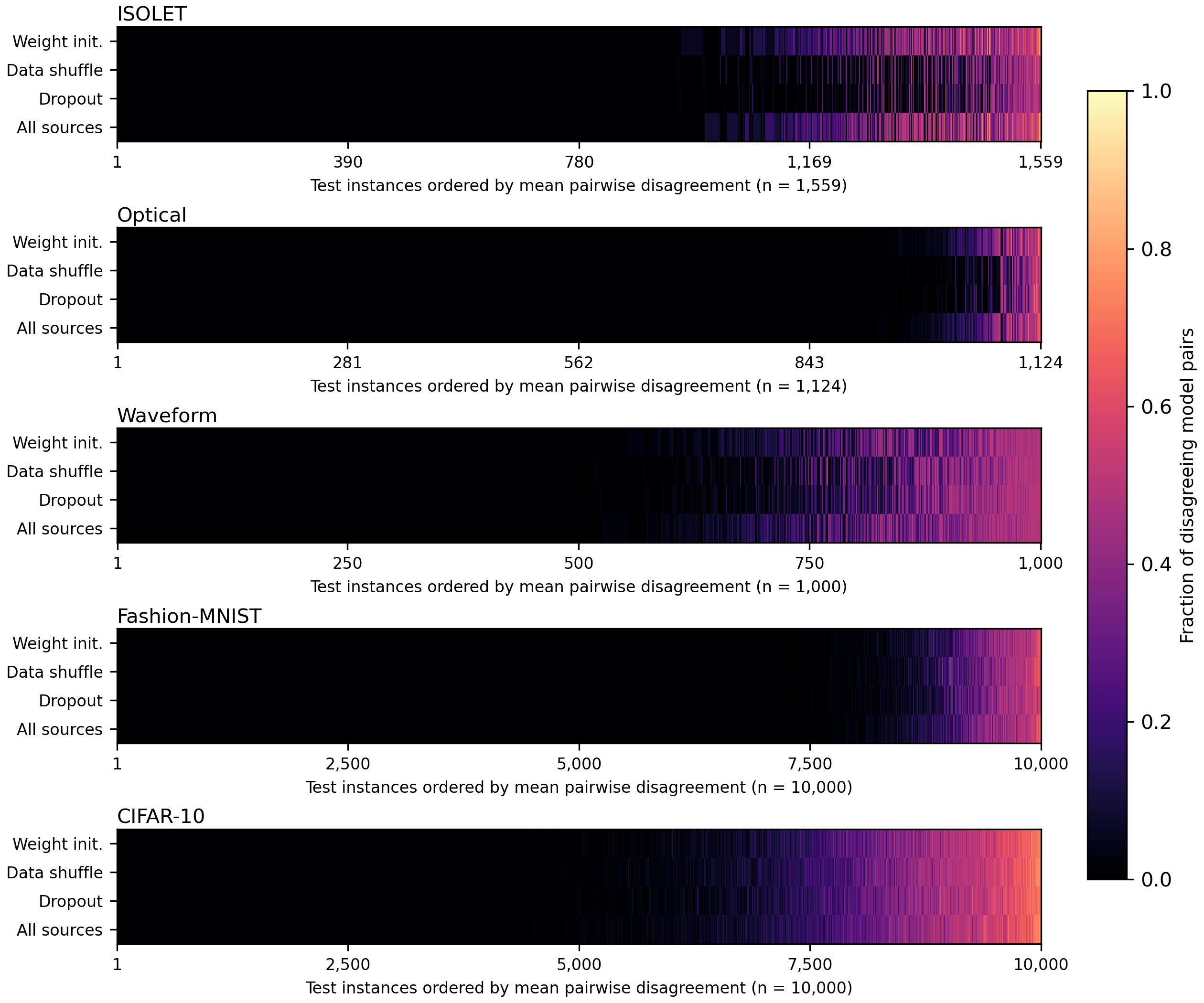}
    \end{minipage}
\end{minipage}

\vspace{0.5em}

\begin{minipage}[c]{0.495\textwidth}
    \begin{minipage}[c]{0.07\linewidth}
        \centering
        \rotatebox{90}{\textbf{BS=128}}
    \end{minipage}%
    \begin{minipage}[c]{0.92\linewidth}
        \centering
        \includegraphics[width=\linewidth]
        {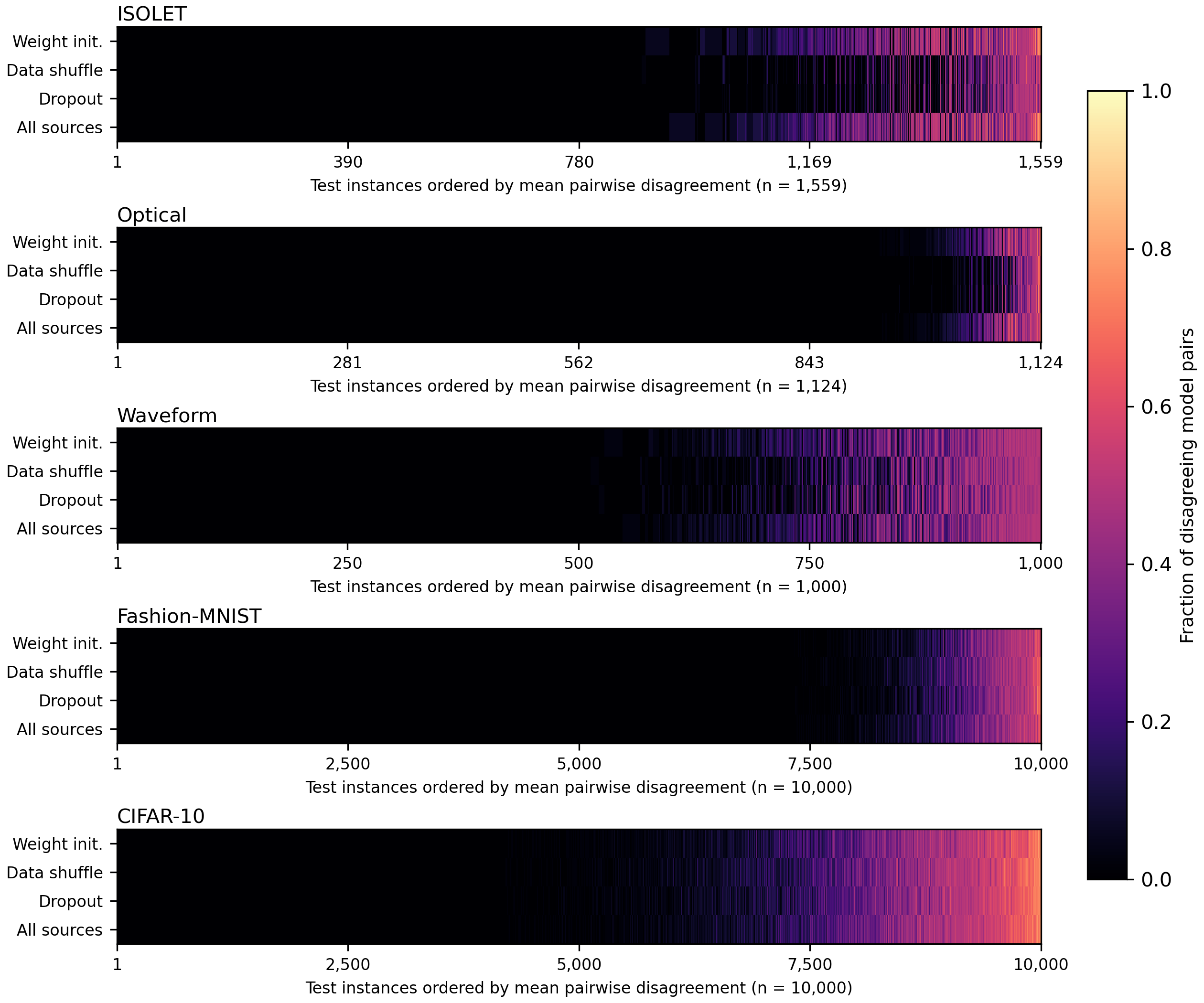}
    \end{minipage}
\end{minipage}%
\hfill
\begin{minipage}[c]{0.495\textwidth}
    \begin{minipage}[c]{0.07\linewidth}
        \centering
        \rotatebox{90}{\textbf{BS=256}}
    \end{minipage}%
    \begin{minipage}[c]{0.92\linewidth}
        \centering
        \includegraphics[width=\linewidth]
        {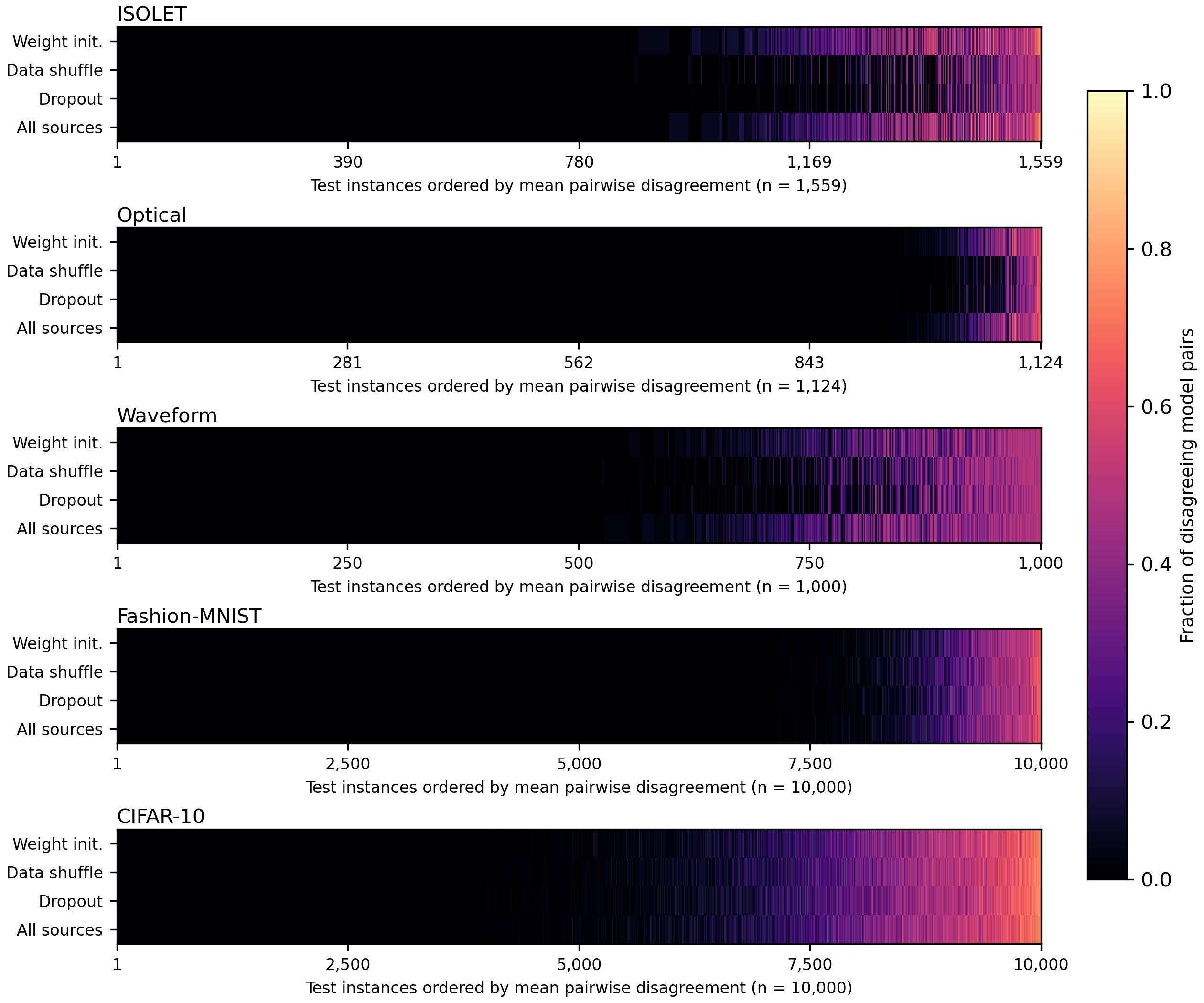}
    \end{minipage}
\end{minipage}

\caption{Per-instance predictive disagreement at $\epsilon=0.03$ across batch sizes. For each stochasticity condition $s$, the colour of the cell associated with test instance $x_i$ represents
$d_i^{(s)}=\frac{1}{\binom{K_s}{2}}\sum_{m<n}\mathbf{1}\!\left[\hat{y}_m(x_i)\neq\hat{y}_n(x_i)\right],$
namely, the fraction of pairs among the $K_s$ models in the
corresponding empirical Rashomon set that predict different classes for that instance. Black indicates complete or near-complete agreement, while lighter colours indicate greater pairwise disagreement. Within each dataset and batch-size configuration, instances are ordered from left to right according to their mean $d_i^{(s)}$ across the four stochasticity conditions. The same instance therefore occupies the same column in all four rows of a given panel, with the instances exhibiting the greatest average disagreement appearing on the right.
The horizontal axis reports the actual rank in this ordering, from 1 to the total number of test instances.}

\label{fig:per_instance_predictive_ambiguity}
\end{figure*}

\FloatBarrier
Furthermore, to provide a more fine-grained characterization of predictive multiplicity, in Fig. \ref{fig:per_instance_predictive_ambiguity}, we report a per-instance ambiguity heat map for each stochasticity condition. Given a test instance $x_i$, we define its pairwise predictive disagreement as the fraction of all model pairs in the empirical Rashomon set that assign different labels to instance $x_i$. A value of zero indicates unanimous predictions, whereas larger values indicate greater predictive disagreement. The heat maps are reported at $\epsilon=0.03$, the largest performance tolerance considered in our experiments. This setting yields the broadest empirical Rashomon sets and therefore provides an informative view of the heterogeneity that may be hidden by aggregate measures.    
\end{insertedtext}

A first observation is that a larger Rashomon set does not necessarily imply higher ambiguity: \texttt{shuffle} and \texttt{dropout} produce the largest Rashomon sets in tabular data (Fig. \ref{fig:ambiguity-tabular}) yet consistently occupy the lower-ambiguity region of the plot, while \texttt{init}, which yields the smallest sets in several cases, reaches the highest ambiguity values. This shows that the size of the Rashomon set and the predictive multiplicity of its members are driven by different aspects of the training dynamics.

The \texttt{all} configuration closely tracks \texttt{init}, confirming that weight initialisation is the dominant source of predictive disagreement when all stochastic factors vary simultaneously.
The \texttt{shuffle} and \texttt{dropout} clusters are shifted rightward (higher mean accuracy) and downward (lower ambiguity), indicating that models
generated by these configurations are both more accurate on average and more
consistent in their individual predictions.
For the Waveform dataset the separation between \texttt{shuffle} and \texttt{dropout} is less consistent
than on the other datasets, likely reflecting the lower overall accuracy and
higher task difficulty.

For CIFAR10 (fig. \ref{fig:ambiguity-image}), ambiguity reaches relatively high values (up to $\sim 50\%$), indicating substantial predictive multiplicity even among models with similar accuracy. 

Interestingly, ambiguity tends to saturate for moderate values of $\epsilon$, indicating that beyond a certain threshold, adding more models to the Rashomon set does not significantly increase disagreement. 
This suggests that the diversity of predictions is already captured by a relatively small subset of near-optimal models.

In contrast, Fashion-MNIST exhibits significantly lower ambiguity levels (typically below $20\%$), indicating that models with similar performance tend to produce more consistent predictions. 
Furthermore, the ambiguity curves across different stochastic configurations are largely overlapping, reflecting  limited differences of the impact of stochastic training factors on predictive behavior.
Across both datasets, stochastic factors affect predictive multiplicity in a non-uniform manner.    

\begin{insertedtext}
Finally, per-instance ambiguity heat maps show that predictive disagreement is not uniformly distributed across the test set. For several datasets, most instances remain stable, while ambiguity is concentrated in a smaller subset of observations. In contrast, more challenging datasets (such as CIFAR-10 and fashion-MNIST) exhibit disagreement across a broader portion of the test set. These additions complement the aggregate ambiguity scores and provide a more detailed account of individual model behaviour.
\end{insertedtext}

\subsection{Decision-basis multiplicity}

We assess explanation stability by comparing pairwise attribution map similarity across models within~$\mathcal{R}_\epsilon$, using different XAI methods: VG, IG, Occlusion and LRP. Similarity is quantified through Top-$k$
\begin{insertedtext}
    with $k \in \{5\%,10\%,20\%,25\%\}$
\end{insertedtext}
feature overlap, which measures the fraction of the $k$ most relevant features shared between two attribution maps. The analysis is conducted on pairs of models that predict the same class for the same input, thereby focusing on attribution disagreement when the predicted output is unchanged.

Figs. \ref{fig:xai-topk-vg-tabular} and \ref{fig:xai-topk-vg-image} summarise  the results using VG. Results for other XAI methods exhibit consistent trends and are therefore given in Appendix for completeness.
\begin{insertedtext}
The absolute attribution-overlap values vary with the Top-$k$ threshold, as larger values of $k$ increase the number of features that two explanations can share. Nevertheless, the comparative patterns across stochasticity sources remain broadly stable over the evaluated thresholds. The limited ranking changes mainly occur between conditions with similar overlap values and do not modify the overall interpretation of the results.
\end{insertedtext}

\begin{insertedtext}
Across datasets, batch sizes, and Top-$k$ thresholds, data shuffling and dropout generally yield higher attribution overlap, whereas weight initialization and the simultaneous variation of all sources tend to yield lower overlap. Occasional exchanges between closely ranked conditions do not alter this overall pattern.
\end{insertedtext}

This finding suggests that predictive similarity and attribution-overlap are not necessarily correlated. Although \texttt{Init} generates a more restricted Rashomon set in terms of near-optimal models, the resulting solutions often exhibit markedly different feature-attribution profiles, resulting in lower explanatory agreement. Conversely, models obtained through variations in mini-batch ordering (\texttt{Shuffle}) or dropout (\texttt{Dropout}) tend to exhibit greater consistency in the features identified as relevant.


\begin{insertedtext}
On image datasets, this separation is weaker (Fig. \ref{fig:xai-topk-vg-image}). This difference may be associated with a combination of data modality, input dimensionality, and architectural inductive biases. For example, convolutional models may learn partially shared hierarchical feature representations across stochastic realisations. 
\end{insertedtext}
Furthermore, for CIFAR-10, Top-10\% feature overlap  operates on approximately 307 components, making incidental agreement between any two maps more likely than on tabular datasets where the same threshold covers only a handful of features.
\begin{insertedtext}
However, because the tabular and image experiments differ in both datasets and model architectures, our experimental design cannot determine whether the observed differences are caused by data modality, input dimensionality, architectural properties, or a combination of these factors. These explanations should therefore be regarded as hypotheses rather than causal conclusions.
\end{insertedtext}

Finally, for the image datasets, attribution similarity is largely insensitive to the value of~$\epsilon$ across all settings: the overlap profiles are nearly flat across the range explored, indicating that the choice of tolerance threshold $\epsilon$ has limited practical impact on the XAI conclusions drawn from the Rashomon set. Considering these results in relation to the Rashomon set size, increasing the number of near-optimal models by relaxing the accuracy threshold does not substantially alter the degree of explanatory agreement among them, particularly on the image datasets. Despite potentially large variations in Rashomon set size, models within the set tend to exhibit comparable levels of explanation diversity. This suggests that explanatory agreement is only weakly related to the number of admissible near-optimal solutions and is instead more strongly influenced by the characteristics of the data and the learning process itself.

\section{Discussion}
\label{sec:discussion}
The results presented in this work provide empirical evidence that the Rashomon phenomenon manifests at multiple levels of model behaviour, and that the stochastic components of the training procedure play a distinct and measurable role at each level. We discuss the main findings across the three dimensions analysed.

\paragraph{Rashomon set size and accuracy distributions.}
The empirical size of the Rashomon set is primarily determined by the spread of the accuracy distribution within each pool, which in turn depends on the source of stochasticity being varied.
Weight initialisation produces in several cases the widest accuracy distributions, resulting in smaller Rashomon sets at fixed $\epsilon$: because individual models differ more substantially in their convergence quality, only a limited fraction falls within a given tolerance of the best model.
Data shuffling and dropout, by contrast, produce more concentrated distributions, yielding larger Rashomon sets.
The \texttt{all} configuration, in which all sources of randomness vary simultaneously, closely tracks \texttt{init} in most settings, suggesting that weight initialisation is a dominant factor in shaping the variability of the
solution space.

The effect is strongly dataset-dependent.
On simpler datasets such as Optical Recognition of Handwritten Digits and
Fashion-MNIST, the Rashomon set saturates to the full pool even for small
$\epsilon$, indicating that the optimisation landscape admits a large number of near-equivalent solutions.
Conversely, on more complex datasets  the growth of the Rashomon set is more gradual and more sensitive to the stochastic configuration.

\paragraph{Predictive multiplicity.}
The analysis of ambiguity reveals a dissociation between Rashomon set size and predictive multiplicity.
Configurations such as \texttt{shuffle} and \texttt{dropout}, which typically
produce larger Rashomon sets, exhibit relatively low ambiguity: models that differ only in mini-batch ordering or dropout mask realisations tend to agree on individual predictions despite their parameter differences. Conversely, \texttt{init} produces smaller Rashomon sets but higher ambiguity, particularly on more complex or tabular datasets.

This dissociation has direct practical implications: the number of near-optimal
models is not a reliable proxy for the degree of predictive disagreement among
them. A Rashomon set can be large yet homogeneous in its predictions, or small
yet highly diverse. The source of stochasticity that generates the model pool plays a more critical role than its size.

\paragraph{Explanation stability}
The analysis of explanation similarity further highlights the multi-layered nature of model multiplicity. Across configurations, models belonging to the same Rashomon set can exhibit non-negligible variability in their attribution maps, indicating that similar predictive performance does not necessarily imply similar decision bases.

In line with the results on predictive multiplicity, weight initialisation tends to induce the largest variability in explanations, suggesting that different initialisations  lead to models producing less consistent attribution maps.
Conversely, data shuffling and dropout generally produce more stable explanations, indicating that these sources of stochasticity affect model parameters without substantially altering the underlying input--output relationships.

An additional layer of variability arises from the choice of the explanation method itself.
Different XAI techniques may produce divergent attribution patterns even for the same model and input, highlighting that explanation instability is not only a consequence of model multiplicity but also of methodological differences in explanation generation.
This further complicates the interpretation of explanation similarity as a proxy for model equivalence.

\paragraph{Observations}
Two observations cut across all three dimensions of the analysis.

First, weight initialisation emerges as the most influential stochastic factor on tabular datasets. It consistently produces the widest accuracy distributions, the highest predictive multiplicity, and the lowest explanation stability. 
\begin{insertedtext}
On image datasets, however, the differences among stochasticity sources are substantially reduced. This contrast may reflect several inseparable factors, including data modality, input dimensionality, and architectural inductive biases. For instance, 
\end{insertedtext}
convolutional architectures learn a hierarchy of low- and mid-level visual features that remains relatively stable across different stochastic realizations. Consequently, variations in training stochasticity mainly affect the final classifier rather than the feature extraction process, leading to smaller differences in both predictions and explanations.
\begin{insertedtext}
However, since MLP and ResNet-20 are not evaluated on the same datasets, our experiments cannot attribute the contrast specifically to architecture or network depth. A controlled cross-architecture study on common datasets would be required to test this hypothesis.
\end{insertedtext}
 
Second, the relationship between Rashomon set size and the other dimensions is non-monotonic and configuration-dependent.
A larger Rashomon set does not necessarily imply higher predictive multiplicity or lower explanation stability. This effect is particularly evident for predictive multiplicity across all datasets and for explanation stability on the image datasets, where explanatory agreement remains remarkably stable despite substantial variations in Rashomon set size. This finding highlights that solution-space multiplicity, predictive multiplicity, and decision-basis multiplicity capture complementary aspects of variability, and should be analysed jointly rather than in isolation.

Taken together, the results suggest that solution-space multiplicity, predictive multiplicity, and decision-basis multiplicity represent distinct manifestations of the Rashomon phenomenon. Characterizing only one of these dimensions provides an incomplete picture of model variability, as models that appear similar from one perspective may differ substantially from another.

\paragraph{Limitations}
Several limitations should be acknowledged.
First, the empirical Rashomon set considered in this work captures only the subset of near-optimal models reachable through the specific stochastic training procedures adopted, and does not approximate the full theoretical Rashomon set over the hypothesis space. Second, the analysis is restricted to a limited set of architectures and datasets; extending the study to larger-scale models and more complex tasks would be necessary to assess the generality of the observed trends.
\begin{insertedtext}
Moreover, architecture and data modality are coupled in our benchmark: MLPs are used for tabular datasets, whereas ResNet-20 is used for image datasets. The reported comparison is therefore descriptive and does not identify a causal effect of architecture or network depth independently of dataset characteristics. A controlled evaluation of multiple architectures on common datasets is required to test architecture-dependent effects.
Third, the evaluation of explanation similarity depends on the choice of XAI methods and similarity metrics, which may introduce additional sources of bias in the analysis.
\end{insertedtext}

\begin{insertedtext}
Finally, the observed reduction in sampling variability supports the adequacy of the selected pool size for the configurations and metric considered here, but it does not establish that 100 models are universally sufficient for other datasets, architectures, stochasticity sources, or multiplicity measures. Future studies could define a minimum effect of interest and conduct an a priori power analysis tailored to a prespecified statistical comparison.
\end{insertedtext}

\paragraph{Future directions}
The findings open several directions for future research. First, extending the analysis to additional sources of stochasticity, such as optimizer dynamics or data augmentation strategies, could provide a more complete characterisation of solution space exploration.
Second, investigating alternative measures of predictive and explanatory similarity may help better capture the functional diversity of models within the Rashomon set.
Finally, a promising direction is to study how these different forms of multiplicity interact in real-world deployment settings, where consistency in both predictions and explanations is critical for trust and reliability.

\begin{insertedtext}
\section{Guidelines for Mitigating Stochastic Multiplicity}
\label{sec:guidelines}
Practitioners seeking to reproduce a particular training run should separately control and record the random states governing weight initialization, data shuffling, and stochastic regularisation, together with the relevant software, hardware, and deterministic-computation settings. Fixing a single generic seed without documenting the individual sources may be insufficient to reconstruct the experimental conditions.
For robustness assessment, we recommend training multiple models under controlled changes to each stochastic source and examining both aggregate performance and model-level disagreement. Models within a predefined performance tolerance should not be regarded as interchangeable solely because they achieve similar accuracy: their predictive outputs and feature attributions should also be compared.
When multiple high-performing models are available, their predictions may be combined through majority voting or probability averaging. Their feature attributions may also be aggregated. However, an aggregated attribution should be accompanied by an uncertainty measure, such as its standard deviation across models or the frequency with which each feature appears among the Top-$k$ features. This prevents the aggregated explanation from hiding substantial disagreement between individual models. Finally, instances exhibiting substantial disagreement across high-performing models should be considered as uncertainty cases rather than presented with a single unqualified prediction or explanation.
These measures do not eliminate all sources of nondeterminism, particularly across different software and hardware environments. However, they provide an actionable protocol for either controlling stochastic variation, quantifying its effects, or explicitly incorporating it into model selection and reporting.
\end{insertedtext}

\section{Conclusion}
\label{sec:conclusion}
In this work, we investigated the Rashomon phenomenon in deep learning through three complementary dimensions:
(i) solution-space multiplicity, quantified through the empirical size of the Rashomon set;
(ii) predictive multiplicity, analysed through ambiguity among model predictions; and
(iii) decision-basis multiplicity, investigated through attribution agreement using gradient-based, relevance-propagation, and perturbation-based methods.

The results show that stochastic components of the training process play a central role in shaping all three dimensions. Among the considered factors, weight initialisation emerged as the most influential source of variability, consistently producing wider accuracy distributions, higher predictive disagreement, and lower explanation stability.
Conversely, data shuffling and dropout generally produced larger Rashomon sets while maintaining relatively consistent predictive and explanatory behaviour in the investigated XAI approaches.

Importantly, the experiments revealed that the different dimensions of multiplicity are only partially related: a large Rashomon set does not necessarily imply high predictive multiplicity, nor does predictive agreement guarantee explanation consistency.
In several cases, models exhibiting nearly identical predictive performance still relied on different confidence distributions or different subsets of input features. These findings highlight that model equivalence depends strongly on the perspective from which it is analysed.

From a practical standpoint, the results suggest that models belonging to the same Rashomon set cannot always be considered interchangeable, particularly in applications where consistency of predictions and explanations is critical.
In particular, the strong impact of weight initialisation in several cases indicates that controlling the initial optimisation conditions may be important not only for reproducibility, but also for ensuring stable predictive and explanatory behaviour when models are deployed in real-world settings.

Overall, this work highlights the need to analyse the Rashomon phenomenon beyond aggregate performance metrics alone.
The proposed multi-dimensional perspective provides a more comprehensive understanding of model multiplicity in deep learning and opens new directions for studying the relationship between optimisation dynamics, predictive behaviour, and interpretability.

\section*{Acknowledgements}
This work was partially funded by the PNRR MUR project PE0000013-FAIR (CUP: E63C25000630006). We thank Valentina Piscopo for her contribution to the preliminary experimental investigation of this topic.

\appendix
\section{Top-$k$ attribution-overlap sensitivity analysis}
\label{app:topk-sensitivity}

This appendix reports the complete sensitivity analysis of attribution agreement with respect to the Top-$k$ threshold. Results are presented for Top-5\%, Top-10\%, Top-20\%, and Top-25\%, considering all datasets, batch sizes, stochasticity conditions, and XAI methods. Lower overlap indicates stronger attribution-based evidence of decision-basis multiplicity.

\newcommand{\xaitabulargrid}[2]{%
\begin{figure*}[!t]
\centering
\setlength{\tabcolsep}{2pt}
\renewcommand{\arraystretch}{1.1}

\resizebox{\textwidth}{!}{%
\begin{tabular}{rcccc}
&
\textbf{BS=32} &
\textbf{BS=64} &
\textbf{BS=128} &
\textbf{BS=256}
\\

\rotatebox[origin=c]{90}{\textbf{ISOLET}}
&
\figcellxai{xai_topk_#1_mlp_isolet_bs32.png}
&
\figcellxai{xai_topk_#1_mlp_isolet_bs64.png}
&
\figcellxai{xai_topk_#1_mlp_isolet_bs128.png}
&
\figcellxai{xai_topk_#1_mlp_isolet_bs256.png}
\\

\rotatebox[origin=c]{90}{\textbf{Optical}}
&
\figcellxai{xai_topk_#1_mlp_optical_recognition_of_handwritten_digits_bs32.png}
&
\figcellxai{xai_topk_#1_mlp_optical_recognition_of_handwritten_digits_bs64.png}
&
\figcellxai{xai_topk_#1_mlp_optical_recognition_of_handwritten_digits_bs128.png}
&
\figcellxai{xai_topk_#1_mlp_optical_recognition_of_handwritten_digits_bs256.png}
\\

\rotatebox[origin=c]{90}{\textbf{Waveform}}
&
\figcellxai{xai_topk_#1_mlp_waveform_database_generator_version_1_bs32.png}
&
\figcellxai{xai_topk_#1_mlp_waveform_database_generator_version_1_bs64.png}
&
\figcellxai{xai_topk_#1_mlp_waveform_database_generator_version_1_bs128.png}
&
\figcellxai{xai_topk_#1_mlp_waveform_database_generator_version_1_bs256.png}
\\
\end{tabular}
}

\caption{Top-$k$ attribution overlap obtained using #2 as a function
of the performance tolerance $\epsilon$ for the tabular datasets and
MLP architecture. Rows correspond to datasets and columns to batch
sizes. Each plot contains the results for Top-5\%, Top-10\%, Top-20\%,
and Top-25\%. Lower overlap indicates stronger attribution-based
evidence of decision-basis multiplicity. Error bars report the standard
deviation across model-pair--instance comparisons. Points are omitted
when the empirical Rashomon set contains fewer than two models, since
pairwise agreement is then undefined. Small horizontal offsets are
used only to make overlapping markers and error bars visible.}
\label{fig:xai-topk-#1-tabular}
\end{figure*}
}

\newcommand{\xaiimagegrid}[2]{%
\begin{figure*}[!t]
\centering
\setlength{\tabcolsep}{2pt}
\renewcommand{\arraystretch}{1.1}

\resizebox{\textwidth}{!}{%
\begin{tabular}{rcccc}
&
\textbf{BS=32} &
\textbf{BS=64} &
\textbf{BS=128} &
\textbf{BS=256}
\\

\rotatebox[origin=c]{90}{\textbf{CIFAR-10}}
&
\figcellxai{xai_topk_#1_resnet20_cifar10_bs32.png}
&
\figcellxai{xai_topk_#1_resnet20_cifar10_bs64.png}
&
\figcellxai{xai_topk_#1_resnet20_cifar10_bs128.png}
&
\figcellxai{xai_topk_#1_resnet20_cifar10_bs256.png}
\\

\rotatebox[origin=c]{90}{\textbf{Fashion-MNIST}}
&
\figcellxai{xai_topk_#1_resnet20_fashion_mnist_bs32.png}
&
\figcellxai{xai_topk_#1_resnet20_fashion_mnist_bs64.png}
&
\figcellxai{xai_topk_#1_resnet20_fashion_mnist_bs128.png}
&
\figcellxai{xai_topk_#1_resnet20_fashion_mnist_bs256.png}
\\
\end{tabular}
}

\caption{Top-$k$ attribution overlap obtained using #2 as a function
of the performance tolerance $\epsilon$ for the image datasets and
ResNet-20 architecture. Rows correspond to datasets and columns to
batch sizes. Each plot contains the results for Top-5\%, Top-10\%,
Top-20\%, and Top-25\%. Lower overlap indicates stronger
attribution-based evidence of decision-basis multiplicity. Error bars
report the standard deviation across model-pair--instance comparisons.
Points are omitted when the empirical Rashomon set contains fewer than
two models, since pairwise agreement is then undefined. Small
horizontal offsets are used only to make overlapping markers and error
bars visible.}
\label{fig:xai-topk-#1-image}
\end{figure*}
}
\FloatBarrier

\subsection{Integrated Gradients}

\xaitabulargrid{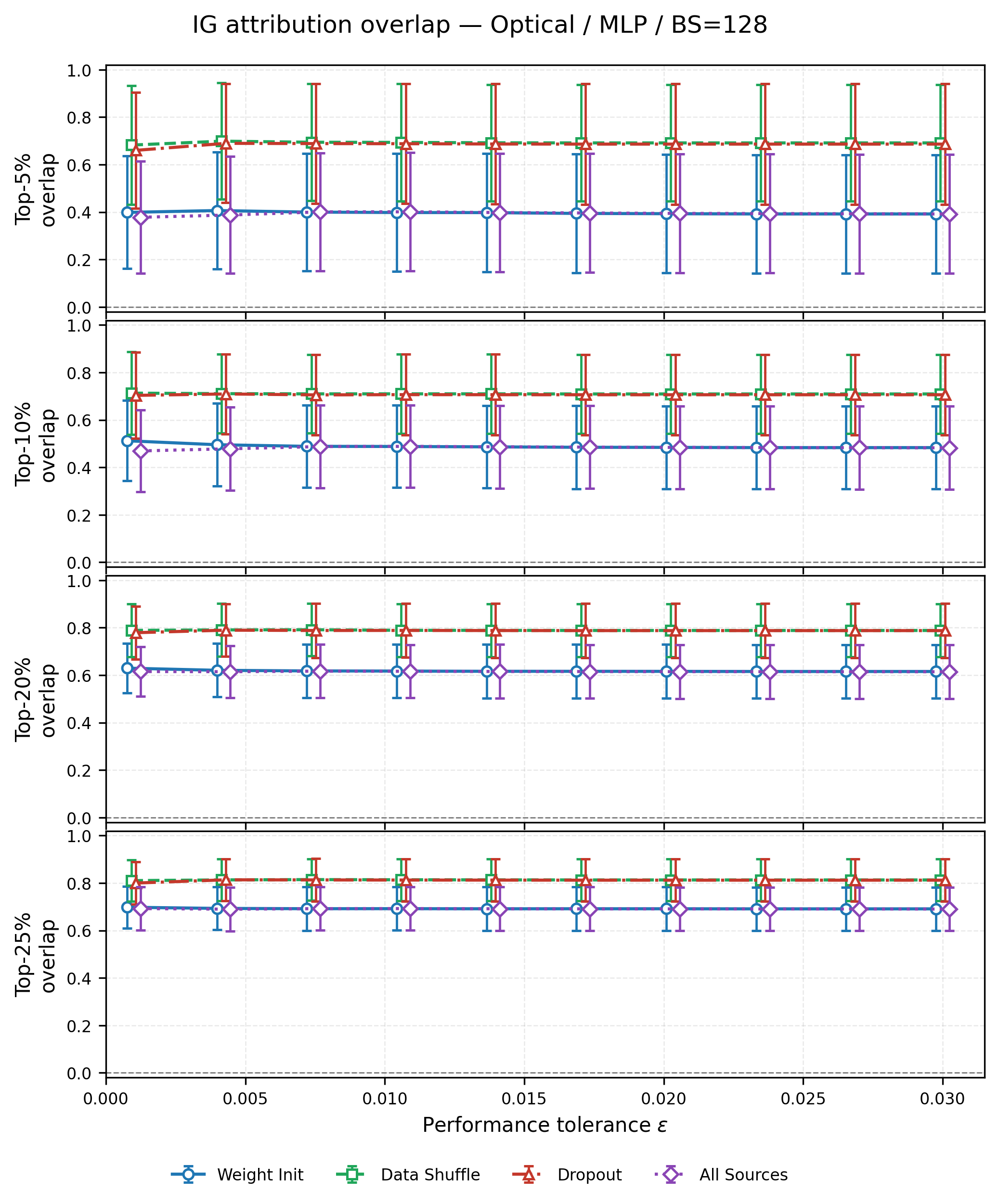}{Integrated Gradients (IG)}
\xaiimagegrid{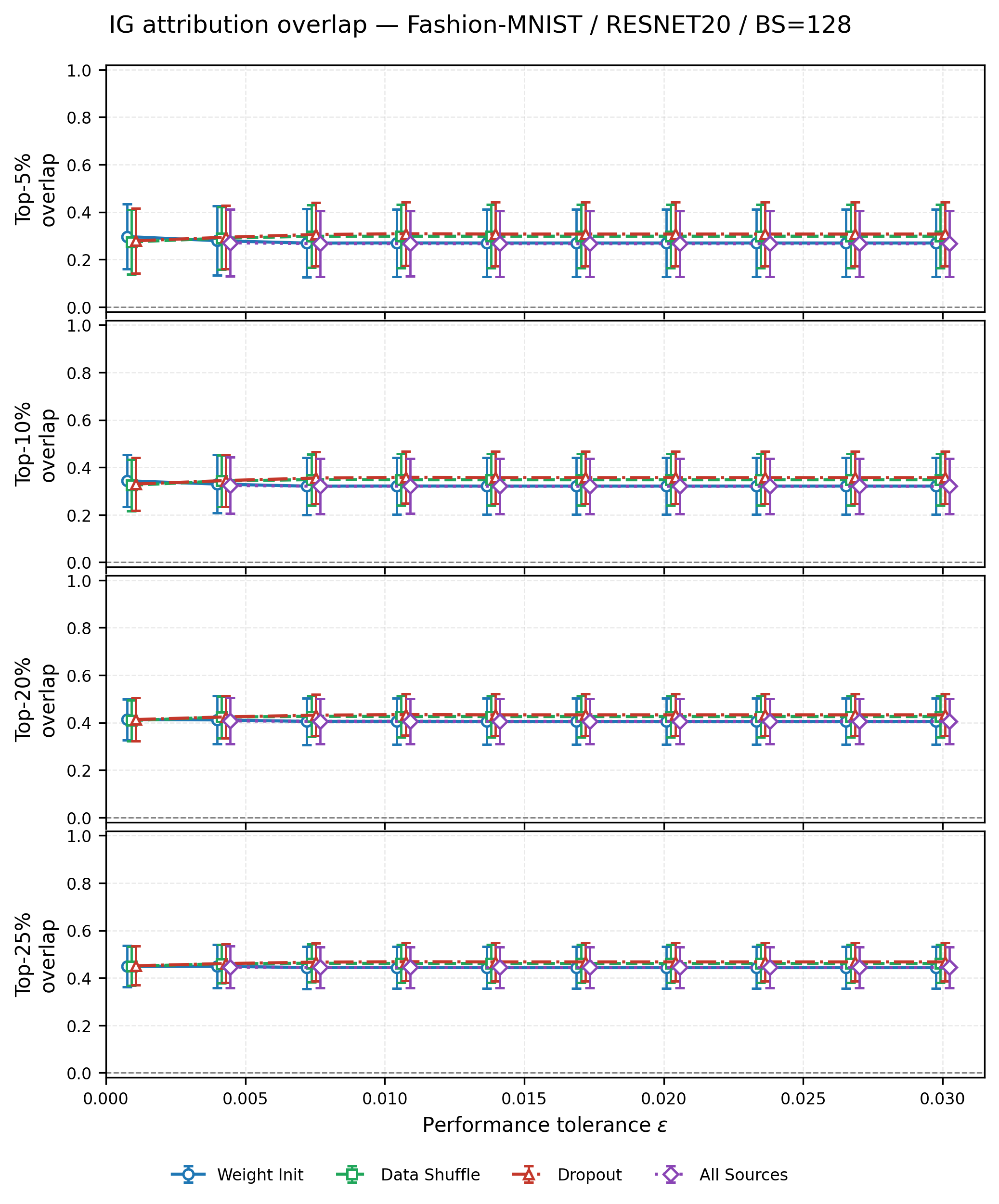}{Integrated Gradients (IG)}
\FloatBarrier

\subsection{Layer-wise Relevance Propagation}

\xaitabulargrid{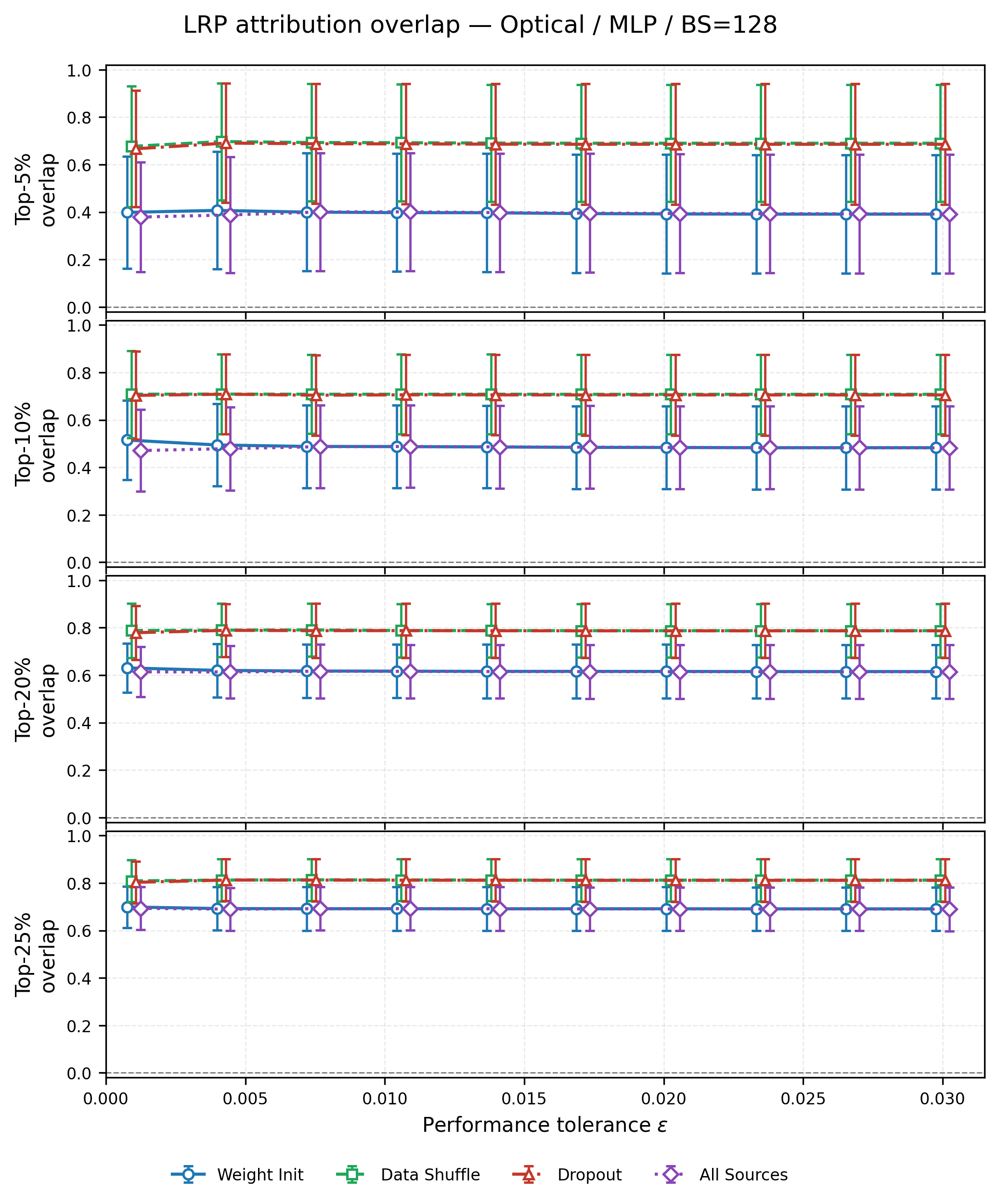}{Layer-wise Relevance Propagation (LRP)}
\xaiimagegrid{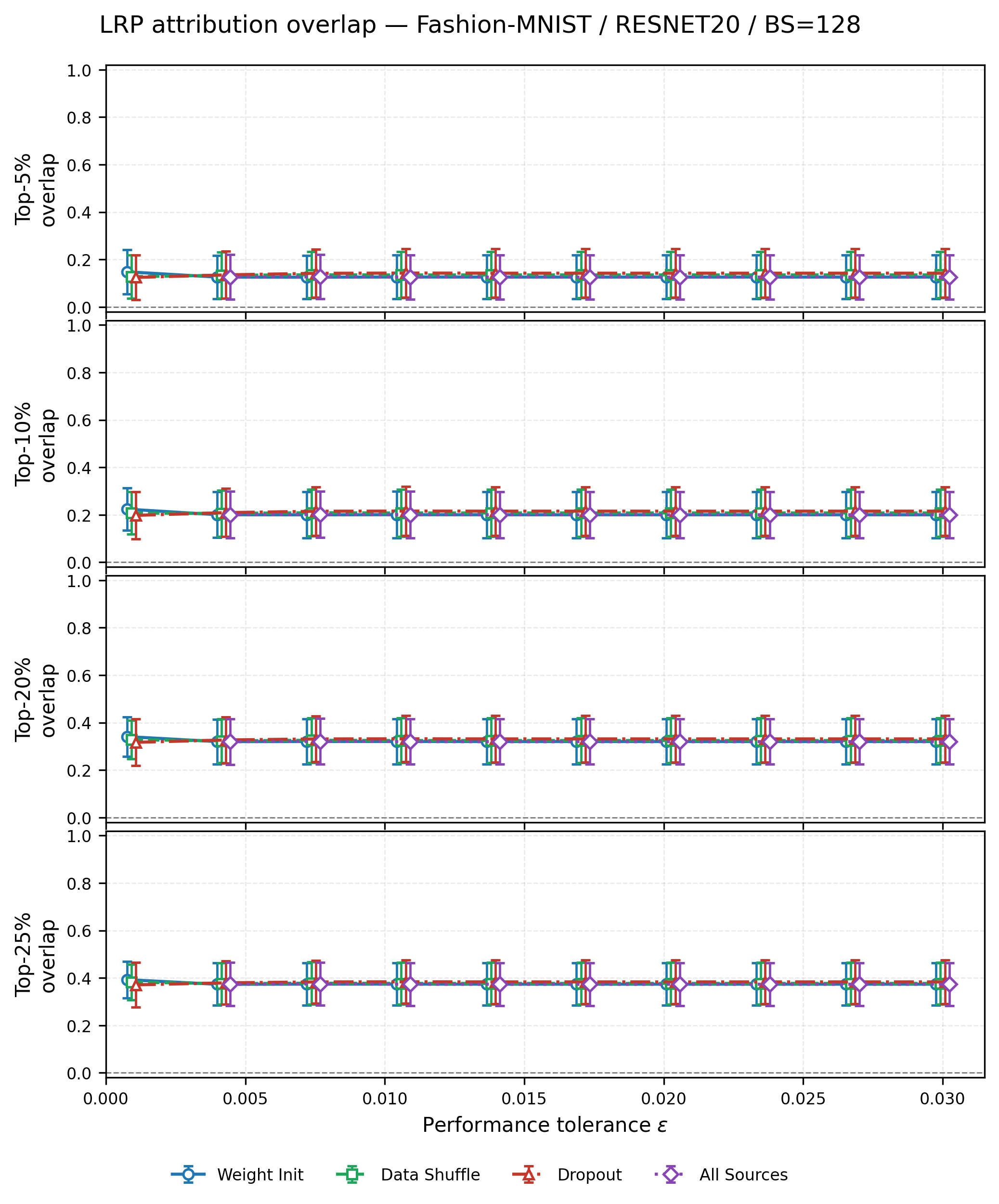}{Layer-wise Relevance Propagation (LRP)}
\FloatBarrier

\subsection{Occlusion}

\xaitabulargrid{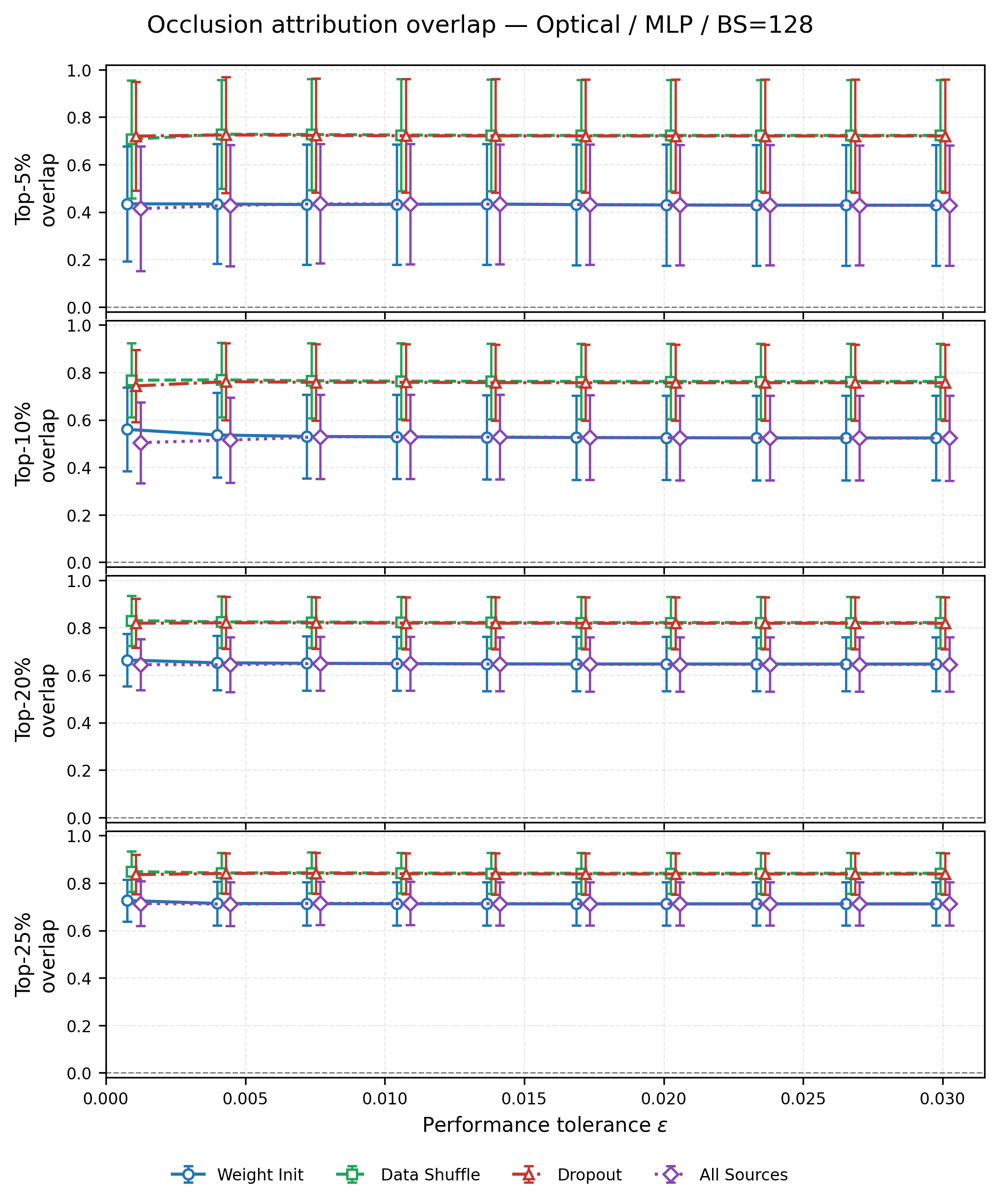}{Occlusion}
\xaiimagegrid{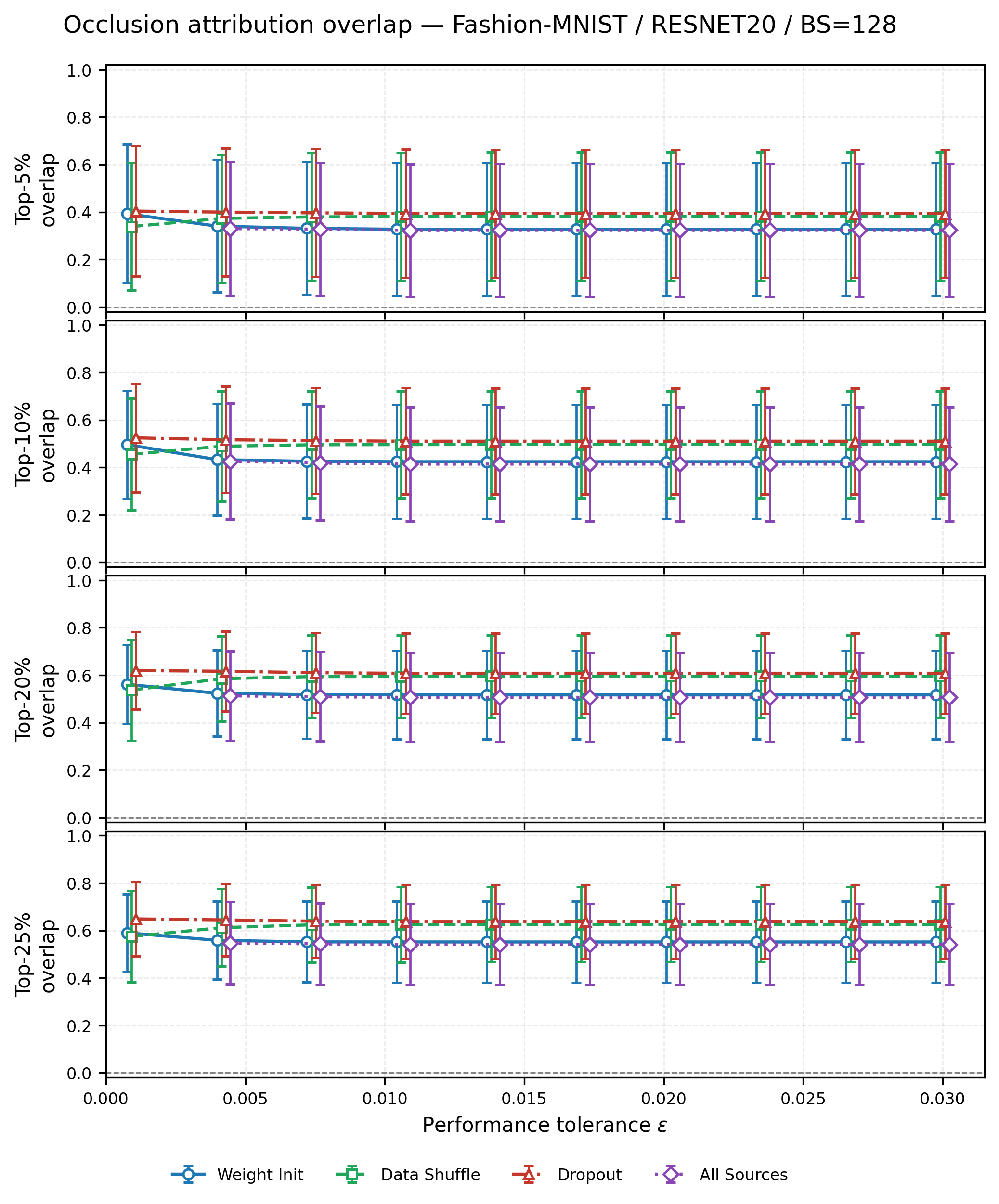}{Occlusion}
\FloatBarrier

\bibliographystyle{elsarticle-harv} 
\bibliography{references}
\newpage

\end{document}